\documentclass[11pt,twoside]{article}

\usepackage{fancyhdr}

\usepackage{amssymb}
\usepackage{graphicx}
\usepackage{subfigure}
\usepackage{srcltx}
\usepackage{latexsym}
\usepackage{amsmath}
\usepackage{eufrak}
\usepackage{amscd}
\usepackage[font=footnotesize]{subfig}

\begin{document}

\title{Fast multi-scale edge-detection in medical ultrasound signals.}

\author{Preben Gr\aa berg Nes \\ \small Department of Mathematics,
  Norwegian University of Science of Technology\\ \small N-7491
  Trondheim, Norway\\ \small e-mail:preben.nes@math.ntnu.no}
\date{}
\maketitle
\thispagestyle{fancy}

\markboth{\footnotesize \rm \hfill P. NES \hfill} {\footnotesize \rm
  \hfill  \hfill}

\begin{abstract}
  \indent In this article we suggest a fast multi-scale edge-detection
scheme for medical ultrasound signals. The edge-detector is based on
well-known properties of the continuous wavelet transform. To achieve
both good localization of edges \textit{and} detect only significant
edges, we study the maxima-lines of the wavelet transform. One can
obtain the maxima-lines between two scales by computing the
wavelet transform at several intermediate scales. To reduce
computational effort and time we suggest a time-scale filtering
procedure which uses only few scales to connect modulus-maxima across
time-scale plane. The design of this procedure is based on a study of
maxima-lines corresponding to edges typical for medical ultrasound
signals. This study allows us to construct an algorithm for medical
ultrasound signals which meets the demand for speed, but not on
expense of reliability.\\ \indent The edge-detection algorithm has
been applied to a large class of medical ultrasound signals including
tumour-, liver- and artery-images. Our results show that the proposed
algorithm effectively detects major features in such signals,
including edges with low contrast. 
\end{abstract}


\section{Introduction}
\indent Recent advances within medical research calls for new
techniques to improve the quality of medical ultrasound
signals. Ultrasound is versatile and non-invasive compared to many
other medical visualization-systems. Unfortunately ultrasound signals
are often heavily contaminated with speckle-noise. In addition, edges
in such signals often have low contrast.  This makes medical
ultrasound signals difficult to interpret and analyze. Improving the
quality of such signals is therefore desirable.\\ \indent In this
article we study a fast wavelet-based edge-detector scheme for medical
ultrasound signals. This scheme is based on properties of the
continuous wavelet transform. We refer to \cite{iD92} and
\cite[Ch.~6]{sM98} for an introduction to the wavelet transform as
well as its application in edge-detection. The continuous wavelet
transform can be used to study local structures - such as edges - at
different scales in a signal. In particular, if the wavelet is given
by the 1st derivative (1-D) or gradient (2-D) of the
Gaussian-function, the local extrema (modulus-maxima) of the wavelet
transform at a scale correspond to edge-points in the signal.\\
\indent In order to explain the idea behind our investigation we will
give a brief survey of the now classic Canny edge-detector
\cite{jC86}. The Canny edge-detector computes the wavelet transform of
a signal at a single scale. If the modulus of the wavelet transform at
a modulus-maximum is larger than a threshold-value its spatial
location represents an edge. At a fine scale the Canny edge-detector
provides fine details, but is sensitive to noise. At a coarse scale
one can detect low-contrast edges, but the positional accuracy of a
modulus-maximum is poor compared to the location of the edge.\\
\indent At each scale the wavelet transform yields a different
description of the edges in the signal. It is clear that for a noisy
signal - such as a medical ultrasound signal - there is no single
scale which provides a categorically correct representation of the
edges. To deal with this ambiguity we - as many others e.g.
\cite{fB87,aW84,jL92,bS05,yY06,lZ02} - suggest to use the wavelet
transform at several scales. The idea is to study modulus-maxima at
coarse scales to detect the edges, and use a fine scale to localize them.\\
\indent The main purpose of this article is to address two issues
which frequently occur when using multi-scale wavelet transform to
detect edges. These are;
\begin{enumerate}
  \item Which modulus-maxima at different scales correspond to the
    same edge in the underlying signal, and
  \item how can one effectively decide which modulus-maxima correspond
    to the same edge.
\end{enumerate}
\indent As the scale decreases each modulus-maximum belongs to a curve
(maxima-line) in the time-scale plane. As $s{\rightarrow} 0$ each
maxima-line propagates towards an edge in the signal. The maxima-lines
provide a natural identification of which modulus-maxima corresponds
to the same edge in the signal.\\ \indent To decide which
modulus-maxima belong to the same maxima-line one can compute the
wavelet transform at several close scales and gradually trace the
modulus-maxima towards fine scales \cite{fB87,aW84}. However, the
wavelet transform computed at two close scales carries almost the same
information, and do not contribute with new information with regards
to the edge-detector scheme studied in this article.  Computing the
wavelet transform at several scales will therefore yield redundancy,
as well as making the procedure computationally expensive.\\ \indent
Our aim is to decide which modulus-maxima belong to the same
maxima-line by computing the wavelet transform at only a few
'different' scales. To achieve this it is essential to understand the
behavior of the maxima-lines in the time-scale plane. This depends on
several factors, such as the intensity and mutual position of the
edges. In general the ambiguity introduced by using multiple scales is
inescapable. The goal is not to eliminate the ambiguity, but rather
manage it and reduce it where possible.\\ \indent For this purpose we
study typical maxima-lines corresponding to medical ultrasound
signals. It happens that local intensity-changes in medical ultrasound
signals often can be approximated by a set of model-edges. In this
article we refer to these as patterns. In particular, the maxima-lines
of each pattern have their special behavior in the time-scale plane.\\
\indent To the best of our knowledge the idea of using model-edges to
study properties of the wavelet transform in the time-scale plane is
due to \cite{mS86} who created an 'elementary' set of
model-edges. Since then this idea has been used by a number of
researchers e.g. \cite{yL89} and more recently \cite{cD04}.\\ \indent
We propose a new time-scale filtering procedure for medical ultrasound
signals. This procedure accommodates the time-scale behavior of both
the maxima-lines and the wavelet transform corresponding to the set of
patterns.  The procedure uses a decision-function which analyzes a
decay- and a distance-criterion to decide which modulus-maxima belong
to the same maxima-line. The procedure is constructed to decide this
by computing the wavelet transform at only a sparse set of scales.\\
\indent The set of patterns serves as test signals for the time-scale
filtering procedure. By analyzing the accuracy of the
decision-function on the maxima-lines corresponding to each pattern we
study the reliability of the procedure. We complement
this with a statistical study on medical ultrasound signals.\\
\indent We also study an edge-detector scheme for medical ultrasound
images which is based on the time-scale filtering procedure. The
edge-detector uses a perceptual criteria similar to \cite{jL92} to
find the significant edges in an image. This method has shown
promising results for detecting and localizing (low-contrast) edges in
medical signals.\\ \indent Several have suggested edge-detectors based
on the multi-scale wavelet transform. We refer to \cite{dZ98} and
references therein for a general exposition of edge-detectors. In
\cite{sM92,sM92a,cD04} edges are characterized by estimating the decay
of the wavelet transform along each maxima-line. In \cite{jL92,lZ02}
they use multi-scale wavelet coefficients to detect and localize
significant edges, and in \cite{tL98,fG10} the authors suggest a
procedure using automatic scale selection. Statistical methods have
been used in \cite{rZ09}, and in \cite{bS05} a method
based on the multi-scale Edge-flow vector field have been proposed.\\
\indent The article is organized as follows.  In the next section we
provide a brief reminder of the wavelet transform and its use to
detect edges. We also discuss general properties of maxima-lines
related to edge-detection. In Sect.\ref{S:III} we propose and study a
time-scale filtering procedure for medical ultrasound signals.  We
conclude Sect.\ref{S:III} with a discussion of a space-scale filtering
procedure for images. This construction is based on our analysis of
the 1-D time-scale filtering procedure. In Sect.\ref{S:IV} we present
the edge-detection algorithm. In the final part of the note we present
experimental results of both time-scale filtering procedure and
edge-detection algorithm. These results are based on a study of both
phantom and actual medical ultrasound images.

\section{Wavelet transform.}\label{S:II}
\subsection{Wavelet transform.}
\indent We consider the following well-known smoothing-function and
wavelet;
\begin{equation*}
  \begin{array}{ll}
  \theta(t) &= \pi^{-1/4}e^{-\frac{t^2}{2}},\\
  \psi(t) &= -\sqrt{2}\,\theta'(t) =
  \sqrt{2}\pi^{-1/4}t\, e^{-\frac{t^2}{2}}.
\end{array}
\end{equation*}
The wavelet transform is defined as e.g. in \cite{sM98};
\begin{displaymath}
  Wf(u,s) =
  \int_{\mathbb{R}}f(t)\frac{1}{\sqrt{s}}\psi\big(\frac{t-u}{s}\big)\;dt.
\end{displaymath}
The candidate edges of $f$ correspond to the points $(u,s)$ in the time-scale
plane where $|Wf(\cdot,s)|$ has a local maximum. These points are referred
to as modulus-maxima, in what follows abbreviated mod-max. The set of mod-max of $Wf(u,s)$ at a
scale $s>0$ will be denoted $\mathcal{M}f(s)$;
\begin{small}
  \begin{displaymath}
    \mathcal{M}f(s) = \big\{u:
    u\,\mathrm{is\,a\,local{-}maximum\,of}\,|Wf(\cdot,s)|\big\}.
  \end{displaymath}
\end{small}
\indent For 2-D signals (images) J. Canny introduced an edge-detector
in \cite{jC86}. This was later formulated in wavelet terminology by
e.g.  \cite{sM92a}. The smoothing function and (directional) wavelets
are given by;
\begin{equation*}
  \begin{array}{ll}
  \theta(x,y) &= \theta(x)\theta(y),\\ \psi^x(x,y) &=
  \psi(x)\theta(y),\quad \psi^y(x,y) =
  \theta(x)\psi(y),
  \end{array} 
\end{equation*}
where $\theta(\cdot)$ and $\psi(\cdot)$ are defined as above. Given a
function $f$ in two variables, let $f_s(x,y) =
s^{-1}f(s^{-1}x{,}s^{-1}y)$. The 2-D wavelet transform is defined as;
\begin{displaymath}
Wf\big((u,s),s\big) = \Bigg( 
\begin{array}{c}
W^xf\big((u,v),s\big)\\
W^yf\big((u,v),s\big)\\
\end{array}
\Bigg) = s\nabla(f\ast \theta_s)(x,y).
\end{displaymath}
The vector $Wf$ is the direction of maximal change of the smoothed
image $f{\ast}\theta_s$. Candidate edges are the points
$\big((u,v),s\big)$ in the space-scale plane where
$|Wf\big((u,v),s\big)|$ has a local maximum in the direction of
maximal change. The set of mod-max at scale $s$ will in what follows
be denoted $\mathcal{M}f(s)$;
\begin{displaymath}
  \mathcal{M}f(s) = \big\{(u,v): (u,v)\,\mathrm{is\,a\,directional}
  \,\mathrm{local\,maximum\,of}\, Wf\big((\cdot,\cdot),s\big)\big\}.
\end{displaymath}
\subsection{Maxima-lines.}\label{SS:mline}
\begin{figure}[!h]
  \centering \subfigure[]{\includegraphics[height=2.4cm,width=6cm]
    {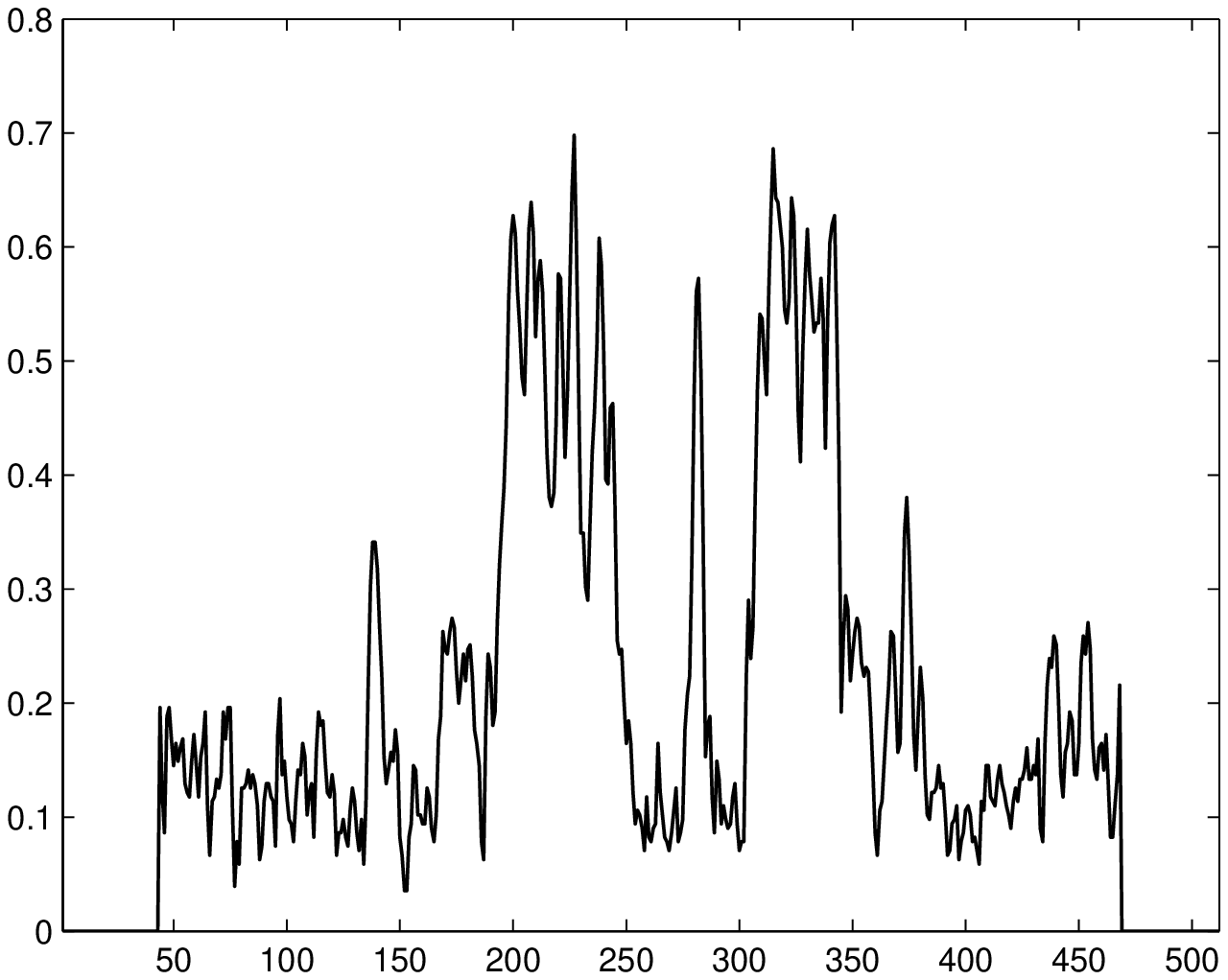}} \subfigure[]{\includegraphics[height=2.4cm,width=6cm]
    {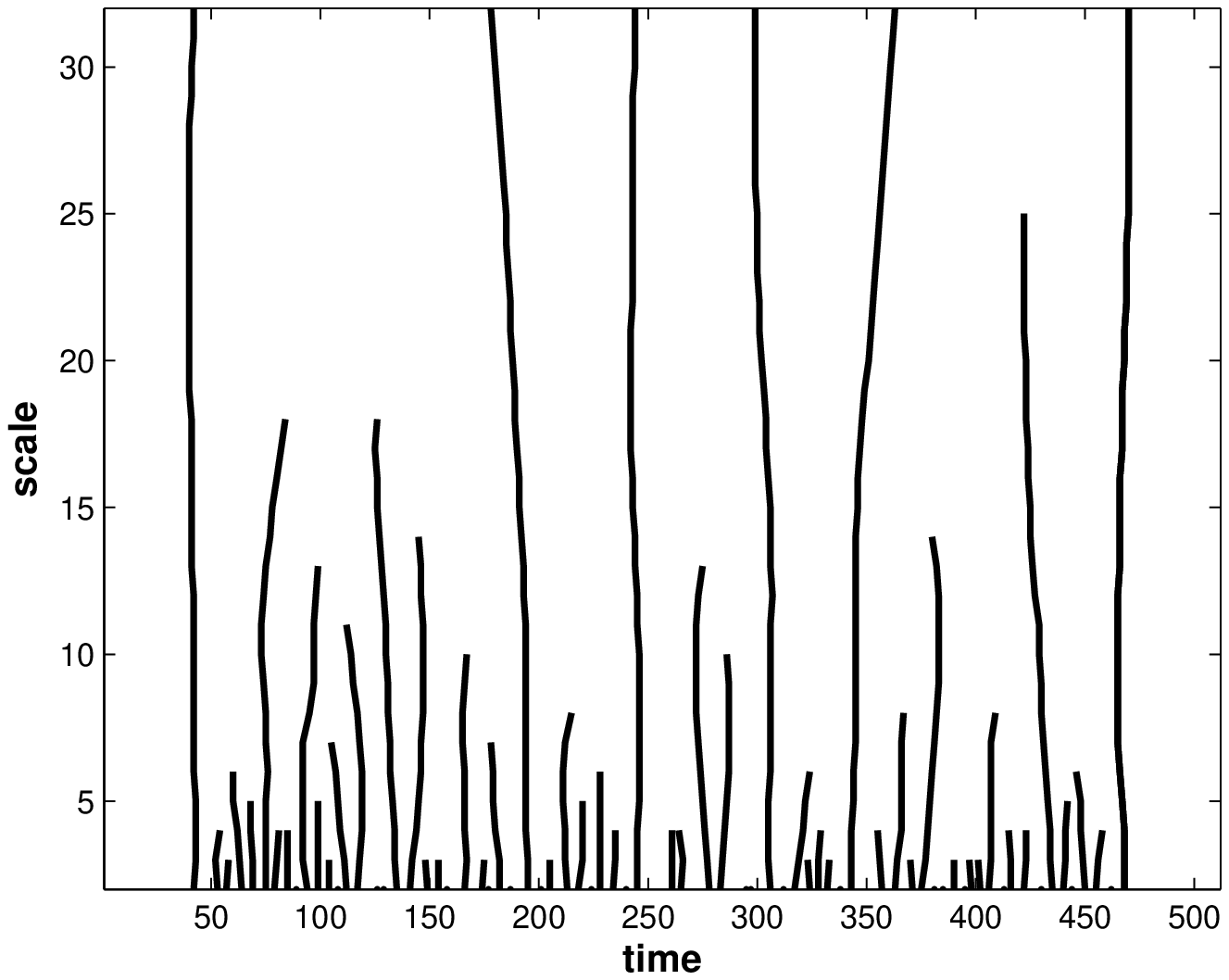}}
  \caption{(a) A 1-D ultrasound signal. (b) The maxima-lines of
    $Wf(u,s)$.}\label{fig:mlines}
\end{figure}

\indent With respect to the wavelet considered in this article it is
well-known that every mod-max of $\mathcal{M}f(s)$ belongs to some
curve
\begin{displaymath}
  \ell = \{(u,s):u = \ell_a(s),\;a\in A\},
\end{displaymath}
where $A$ is some index-set. These curves are called maxima-lines. For
wavelets equal a derivative of the Gaussian-function every maxima-line
$\ell_a(s)$ propagates towards finer scales \cite[page~178]{sM98}, see
fig.\ref{fig:mlines} for an example.
In particular, if no maxima-lines
propagate towards a point in an interval it contains no edges
\cite[Th.~6.5]{sM98}.\\ \indent It will be assumed that a mod-max at a
scale corresponds to only one mod-max at any finer scales. In theory a
maxima-line may bifurcate, but the probability of bifurcating in real
signals is negligible.\\ \indent To illustrate this we consider the
function $f(t) {=} \rho(t) {+} A\rho(t{-}1)$ where $A {>} 0$ and
$\rho(t)$ denotes the Heaviside-function (we remind the definition of
$\rho(t)$ in Sect.\ref{SS:dictionary}). The mod-max of $Wf(u,s)$ are
found by solving
\begin{align*}
  te^{-\frac{t^2}{2s^2}} + A(t-1)e^{-\frac{(t-1)^2}{2s^2}} = 0
  \Leftrightarrow \ln A + \ln \frac{1-t}{t} = \frac{1}{s^2}(\frac{1}{2}
  - t).
\end{align*}
There exists a $s^* > 0$ (see \eqref{eq:s*}) such that this equation
has one and only one solution for every $s>s^*$ and three solutions
(two modulus-maxima and one local minimum) for $s < s^*$. It is clear
that the solution curves (maxima-lines) bifurcate if and only if $\ln
A = 0$ i.e. $A=1$. For all other values of $A$ the maxima-lines do not
bifurcate. Our study of maxima-lines in medical ultrasound signals
indicates that bifurcations are rare in such signals.\\ \indent We
will also assume that two mod-max on the same maxima-line correspond
to the same underlying phenomenon in the signal, and that the accuracy
of the localization of the mod-max increases as $s$ decreases. These
assumptions corresponds to the classical identity- and
localization-assumption of Witkin \cite{aW84}.

\section{Space-scale filtering.}\label{S:III}
\indent The main purpose of this article is to study how one can
connect mod-max between two (perhaps rather distinct) scales. Let
$\ell_a,\,a\in A$ be the maxima-lines of $Wf(u,s)$, and
\begin{displaymath}
  \mathcal{M}f(s_i) = \{\ell_a(s_i) : a\in A\},\qquad i=1,2
\end{displaymath}
be the mod-max at two scales. If $s_2 > s_1$ the number of elements in
$\mathcal{M}f(s_2)$ is less or equal the number of elements in
$\mathcal{M}f(s_1)$ since new maxima-lines may appear between $s_2$
and $s_1$. The question is how one can decide which mod-max of
$\mathcal{M}f(s_1)$ belong to the same maxima-line as a mod-max of
$\mathcal{M}f(s_2)$, and which belong to a maxima-line which appears
between $s_2$ and $s_1$.\\ \indent At fine scales one can decide which
mod-max belong to the same maxima-line by considering their spatial
distance. At coarse scales one can consider the logarithmic-decay of
the wavelet transform.  In addition the sign of the wavelet transform
along a maxima-line can not change.\\ \indent In the next section we
suggest a time-scale filtering procedure for medical ultrasound
signals. This procedure is based on a decision-function
$\mathrm{P}(\cdot,\cdot)$ which analyzes \textit{both} a distance-
\textit{and} a decay-criterion to decide which mod-max belong to the
same maxima-line. We design the function to favor decay at coarse
scales and distance at fine scales. In Sect.\ref{SS:dictionary} we
introduce a set of model-edges, called patterns, which serves as
test-signals for the decision-function. These patterns model local
intensity-changes in medical ultrasound signals. In
Sect. \ref{SS:eval} we study the reliability of the time-scale
filtering procedure with respect to each pattern and their respective
maxima-lines.
\subsection{Decision-function and time-scale filtering
  procedure.}\label{SS:P}
\indent Let $s_2,\,s_1$ be two scales $s_2 \,{>}\, s_1$, and let $(n,s_2)$
and $(m,s_1)$ be two mod-max. We define the decision-function
corresponding to scales $s_2$ and $s_1$ as;
\begin{equation}\label{eq:P}
\mathrm{P}(n,m) = \Delta(n,m)\mathrm{D}(n,m)\mathrm{Sign}(n,m),
\end{equation}
where;
\begin{displaymath}
  \begin{array}{l}
    \Delta(n,m) = \mathrm{exp}\big(-|n-m|s_1^{-\alpha}\big), \\
    \mathrm{D}(n,m) = \mathrm{exp}\Bigg(-\Big|\ln
    \frac{|Wf(n,s_2)|}{|Wf(m,s_1)|}\ln^{-1} \frac{s_2}{s_1} -
    \frac{1}{2}\Big|s_1^{\alpha}\Bigg),\\ \mathrm{Sign}(n,m) =
    \Bigg\{
    \begin{array}{ll} 1 &\quad\textrm{if}\;\mathrm{sign}
      \big(\frac{Wf(n,s_2)}{Wf(m,s_1)}\big) = 1\\ 0
      &\quad\textrm{otherwise.}
    \end{array} 
  \end{array}
\end{displaymath}
The essence of the procedure is to connect mod-max which maximize the
decision-function $\mathrm{P}(\cdot,\cdot)$. For instance, if
$(n,s_2)$ and $(m,s_1)$ \textit{both} belong to the same maxima-line
$\ell_a$ say, then we want $\mathrm{P}(n,m) > \mathrm{P}(n,u)$ for all
$u\in \mathcal{M}f(s_1), u\neq m$. The control-parameter $\alpha$ in
the expressions of $\Delta$ and $\mathrm{D}$ controls the mutual
weighting between the distance- ($\Delta$) and decay-criterion
($\mathrm{D}$). A negative $\alpha$ favors the decay-criterion, while
a positive favors distance. By studying the procedure for various
choices of $\alpha$ we find a weighting between distance and decay
suitable for our type of signals.\\ \indent In the following sections
we will analyze the decision function $\mathrm{P}(\cdot,\cdot)$ and
the corresponding time-scale filtering procedure. This analysis is
based on a set of model-edges which model local intensity
changes in ultrasound signals.
\subsection{Patterns}\label{SS:dictionary}
\indent To investigate whether the suggested procedure successfully
connects mod-max between two scales we have studied a collection of
model-edges, called \textit{patterns}. These patterns - six in total -
model typical transitions between tissues 
\begin{figure}[!h]
  \centering \subfigure{\includegraphics[height=2cm,width=5.2cm]
    {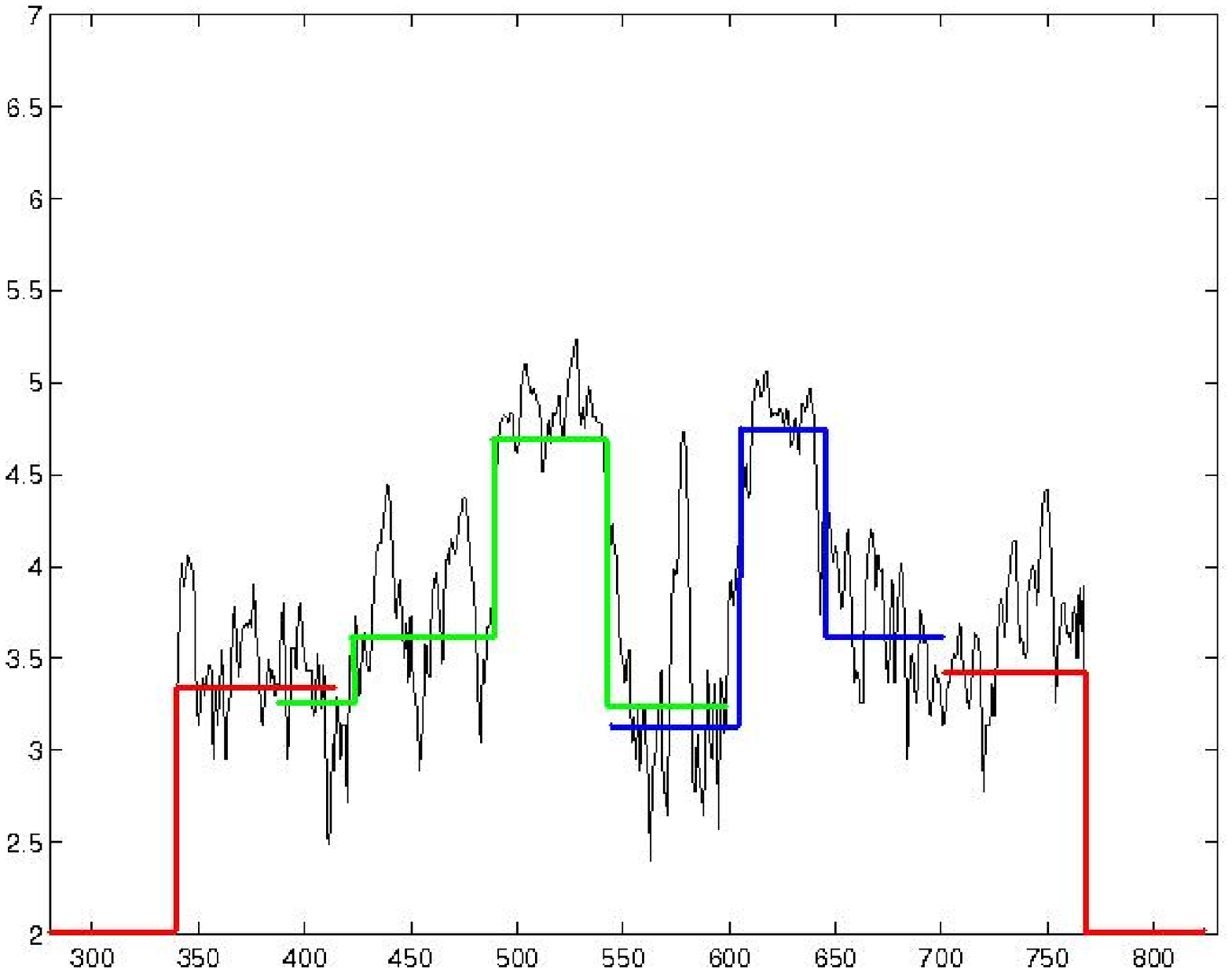}}
  \subfigure{\includegraphics[height=2cm,width=5.2cm]
    {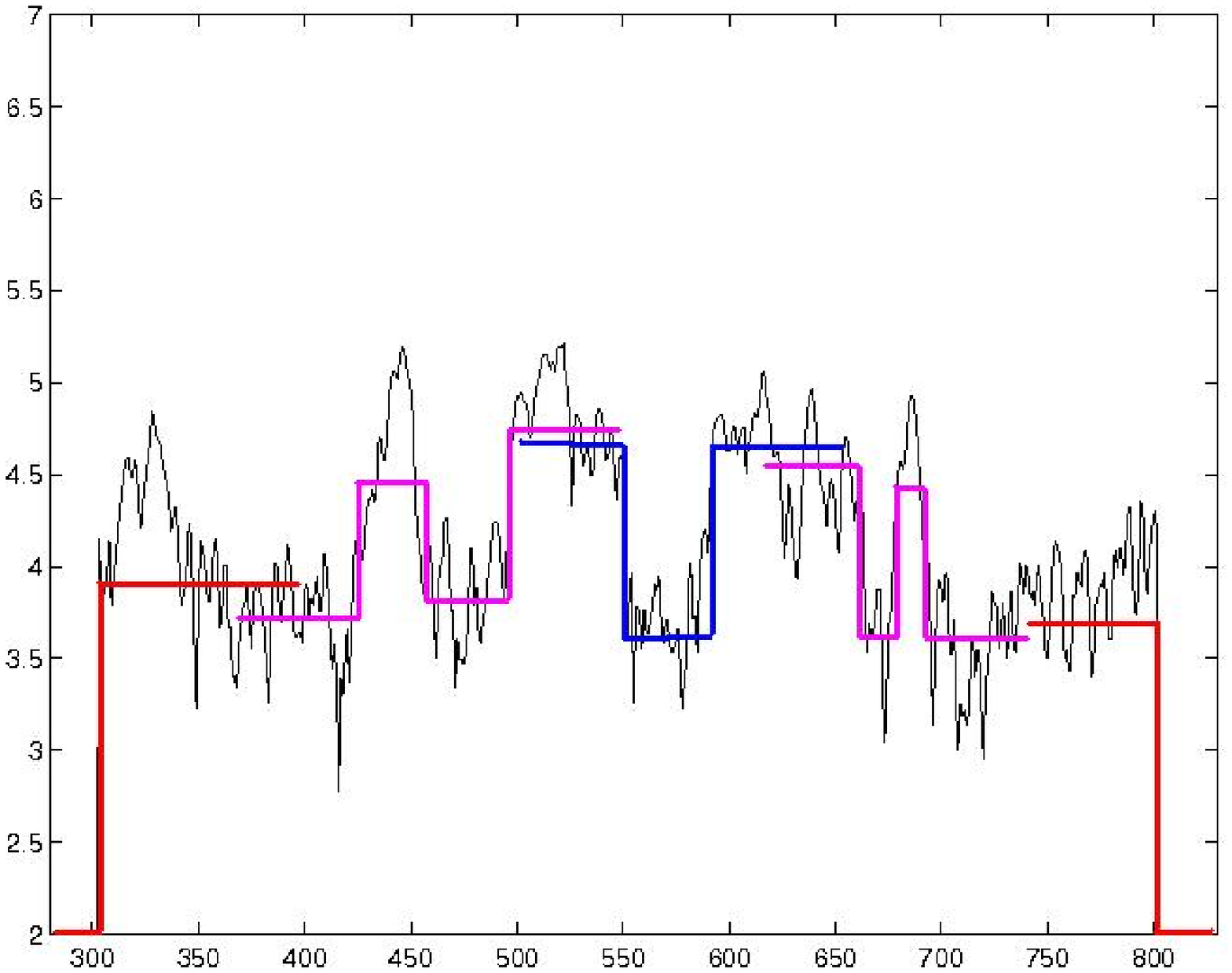}}
  \caption{Illustration of how one can locallly model edges in medical
    ultrasound signals with the patterns from
    Sect.\ref{SS:dictionary}.}\label{fig:atom}
\end{figure}
in medical ultrasound
signals, fig.\ref{fig:atom}.\\ \indent Let $\rho_x$ denote
the translated Heaviside-function;
\begin{displaymath}
  \rho_x(t) \,{=}\, \Big\{\begin{array}{ll}1 & t\geq x\\ 0 & t < x.\end{array}
\end{displaymath}
The patterns can be modeled as;
\begin{small}
\begin{displaymath}
  \begin{array}{llr}
    \mathrm{Pattern\, 1:} & f(t) \,{=}\, \rho_0(t), &\\ 
    \mathrm{Pattern\, 2:} & f(t) \,{=}\, \rho_0(t) {-} 
    B\rho_{\beta}(t), & B,\beta\,{>}\,0,\\ \mathrm{Pattern\, 3:} 
    & f(t) \,{=}\,  \rho_0(t) {+} A\rho_1(t), & A \,{>}\, 1,\\
    \mathrm{Pattern\, 4:} & f(t) \,{=}\, \rho_0(t)
    {+} A\rho_1(t) {-} B\rho_{\beta}(t), & A,\beta \,{>}\, 1,\; 
    B^* \,{>}\, B \,{>}\, 0,\\ \mathrm{Pattern\, 5:} & f(t) 
    \,{=}\,  \rho_0(t) {+} A\rho_1(t) {-} B\rho_{\beta}(t), & 
    \qquad \beta\,{>}\,1, B\,{>}\,0,\; A^* \,{>}\, A \,{>}\, 0,\\ 
    \mathrm{Pattern\, 6:} & f(t) \,{=}\, \rho_0(t) {-}
    B\rho_{\beta}(t) 
    {+} A\rho_1(t), &1{>}\beta{>}0, A,B {>} 0.\\
  \end{array}
\end{displaymath}
\end{small}
\begin{figure}[!h]
  \centering \subfigure[Pattern 1:
  $\rho(t)$]{\includegraphics[height=1.8cm,width=6cm]
    {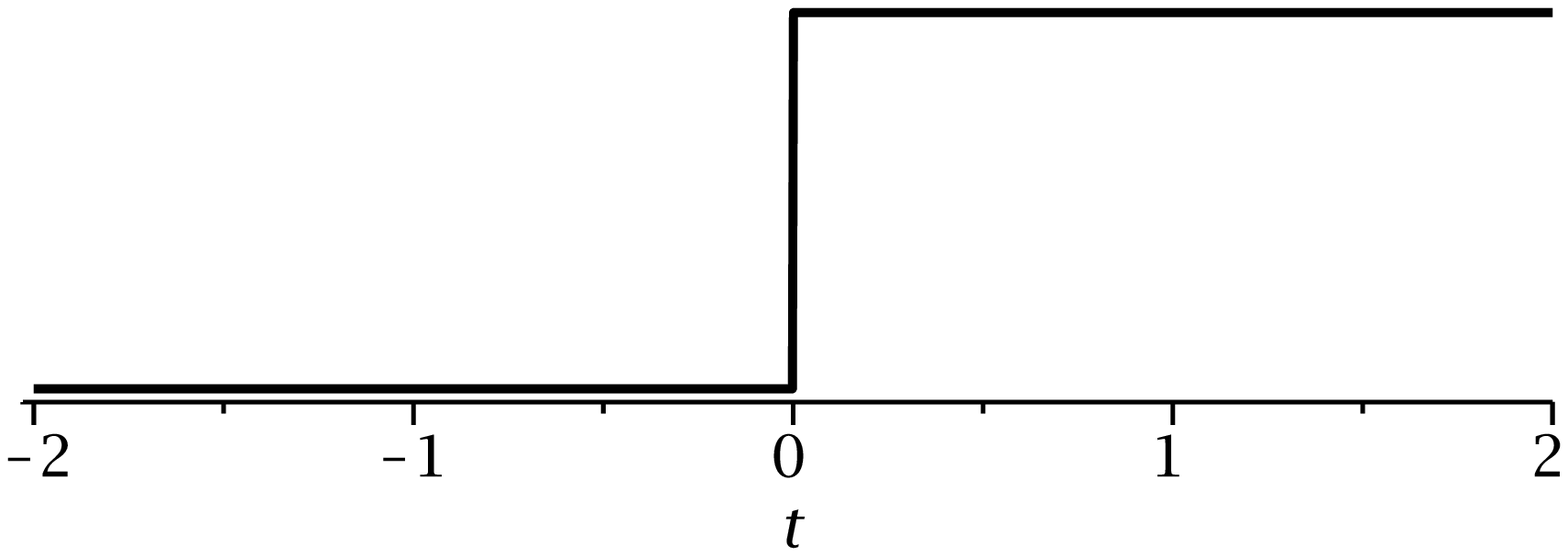}} \subfigure[Maxima-line of pattern
  1]{\includegraphics[height=1.8cm,width=6cm]
    {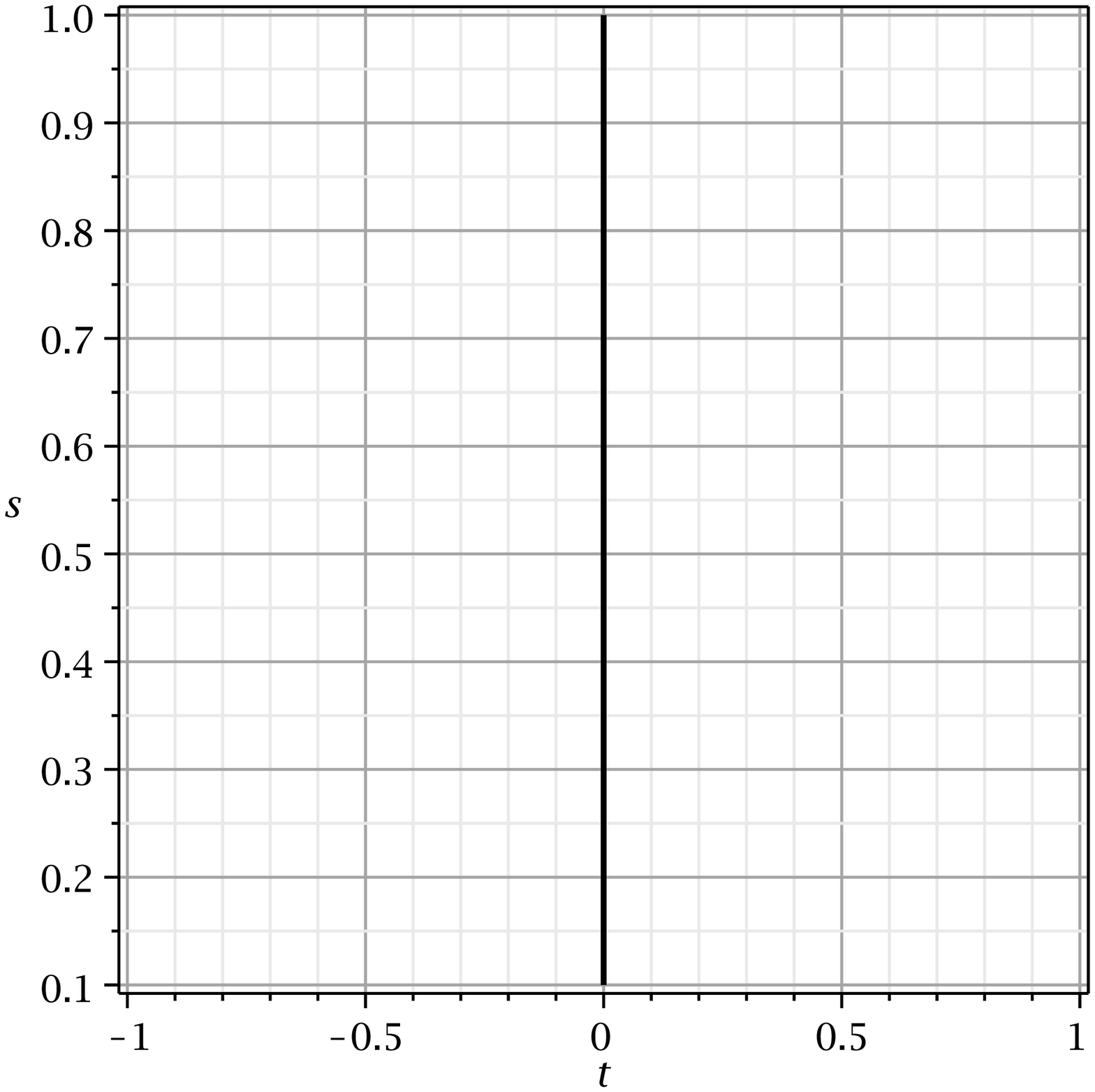}}\\
  \subfigure[Pattern 2: $\rho(t) \,{-}\,
  B\rho_{\beta}(t)$]{\includegraphics[height=1.8cm,width=6cm]
    {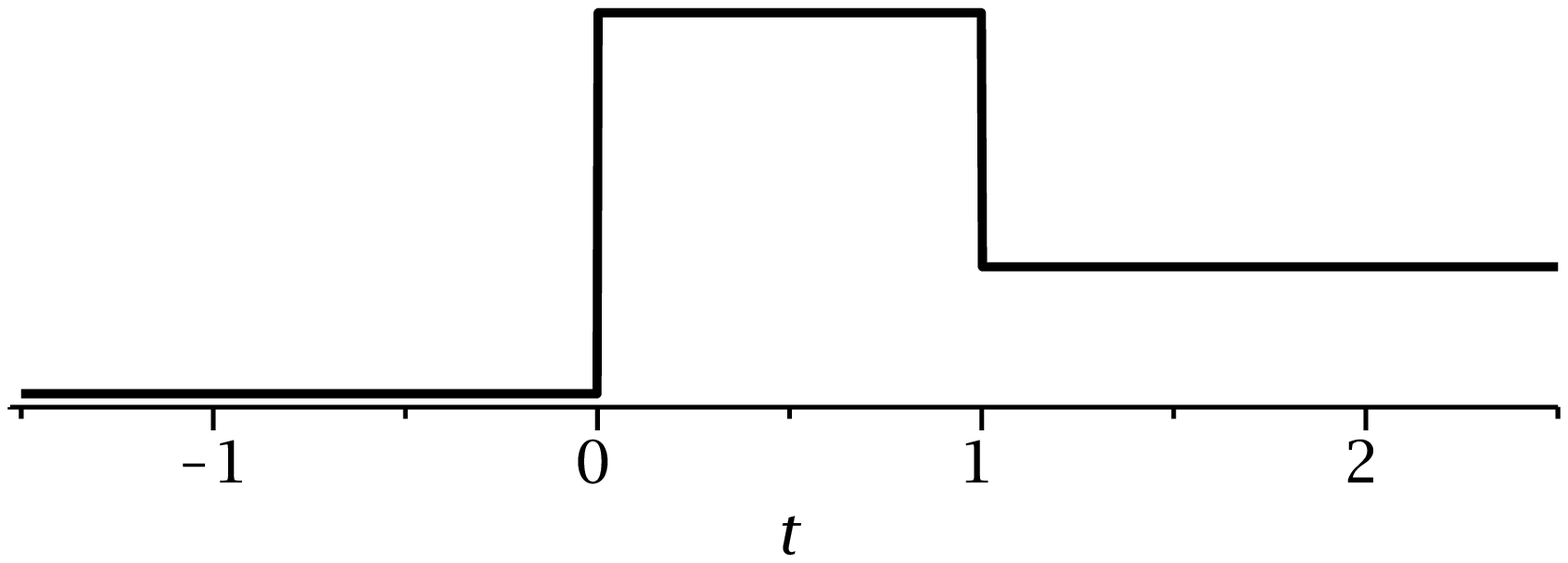}} \subfigure[Maxima-lines of pattern
  2]{\includegraphics[height=1.8cm,width=6cm]
    {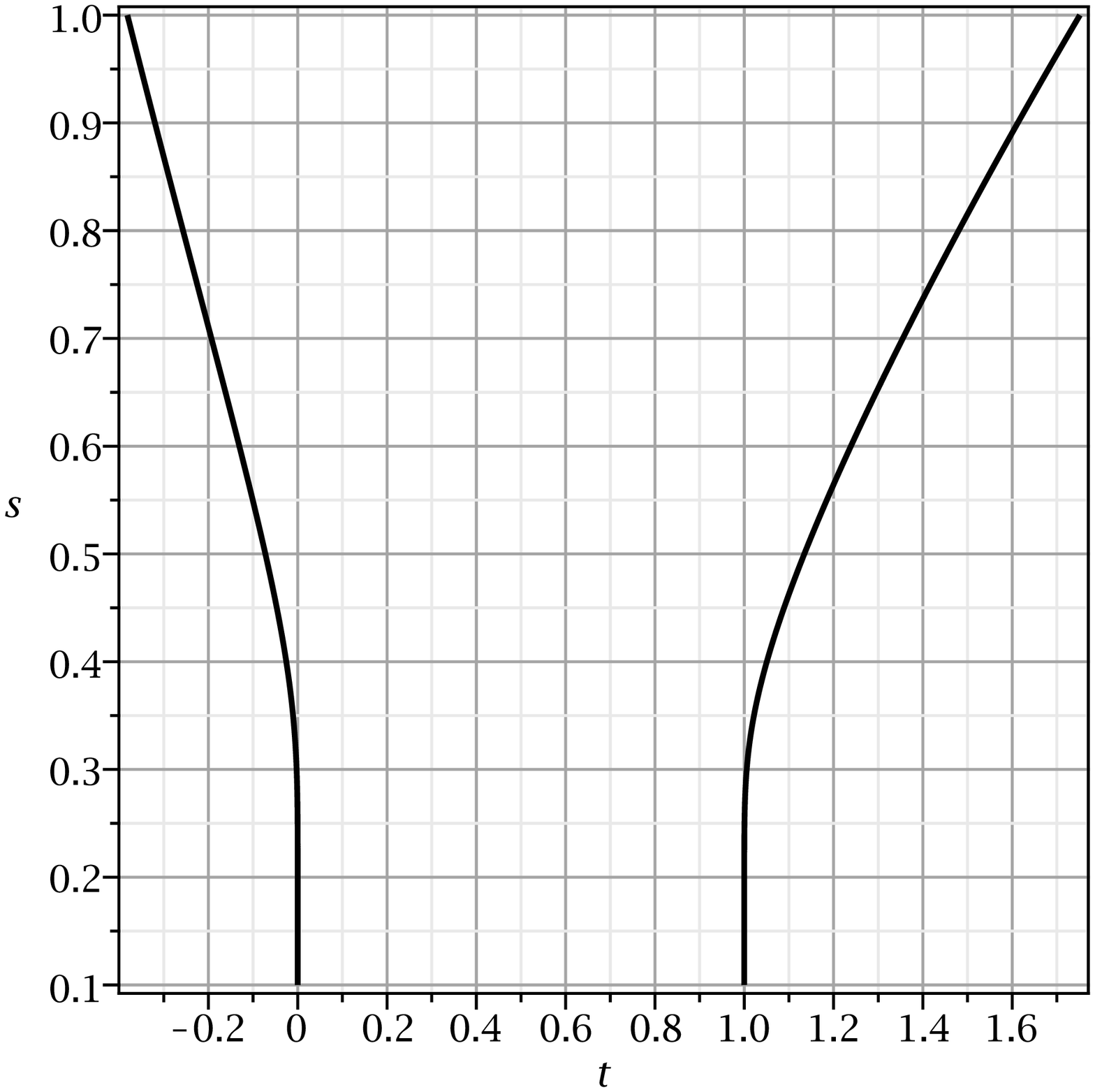}}\\
  \subfigure[Pattern 3: $\rho(t) \,{+}\,
  A\rho_1(t)$]{\includegraphics[height=1.8cm,width=6cm]{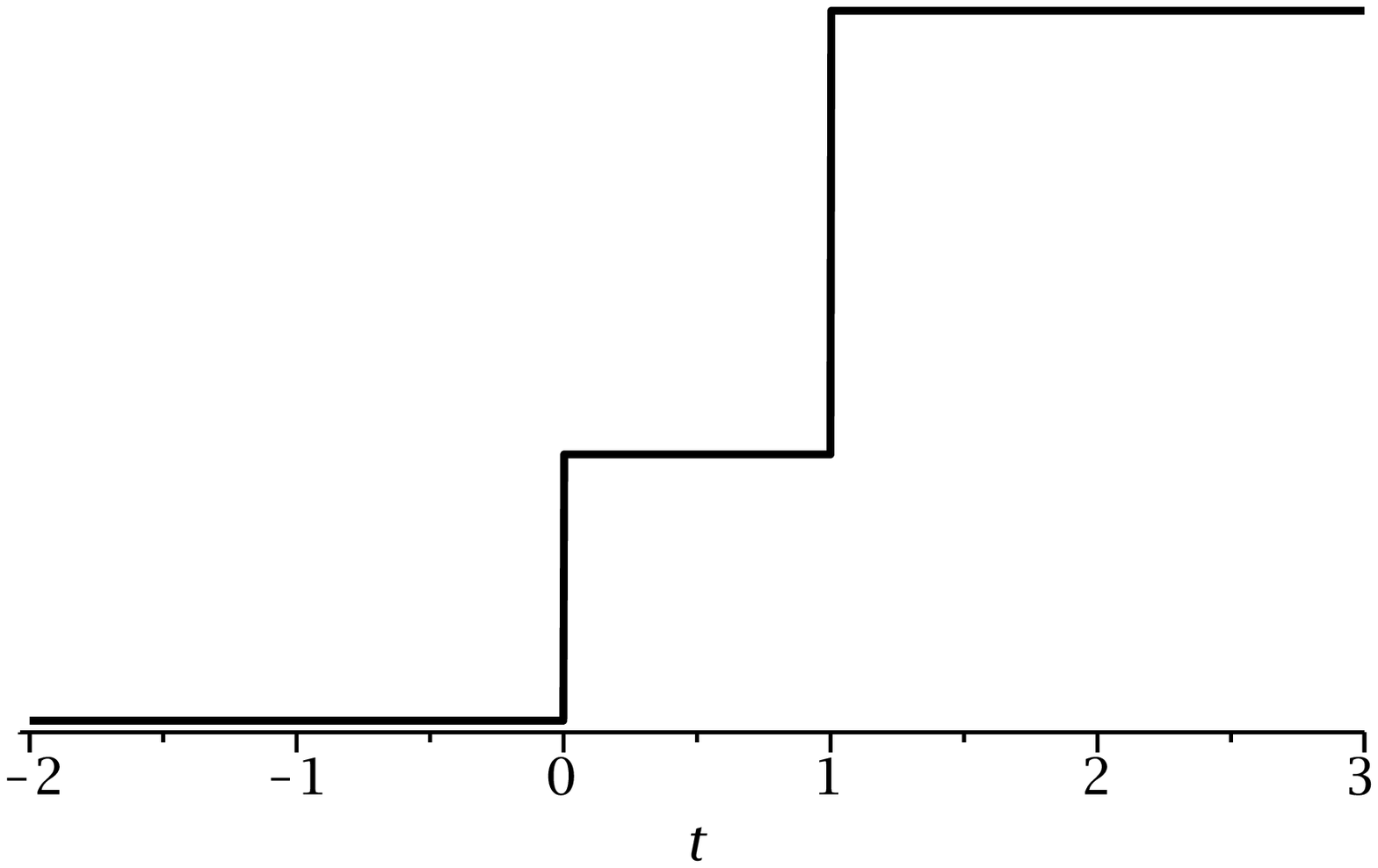}}
  \subfigure[Maxima-lines of pattern
  3]{\includegraphics[height=1.8cm,width=6cm]
    {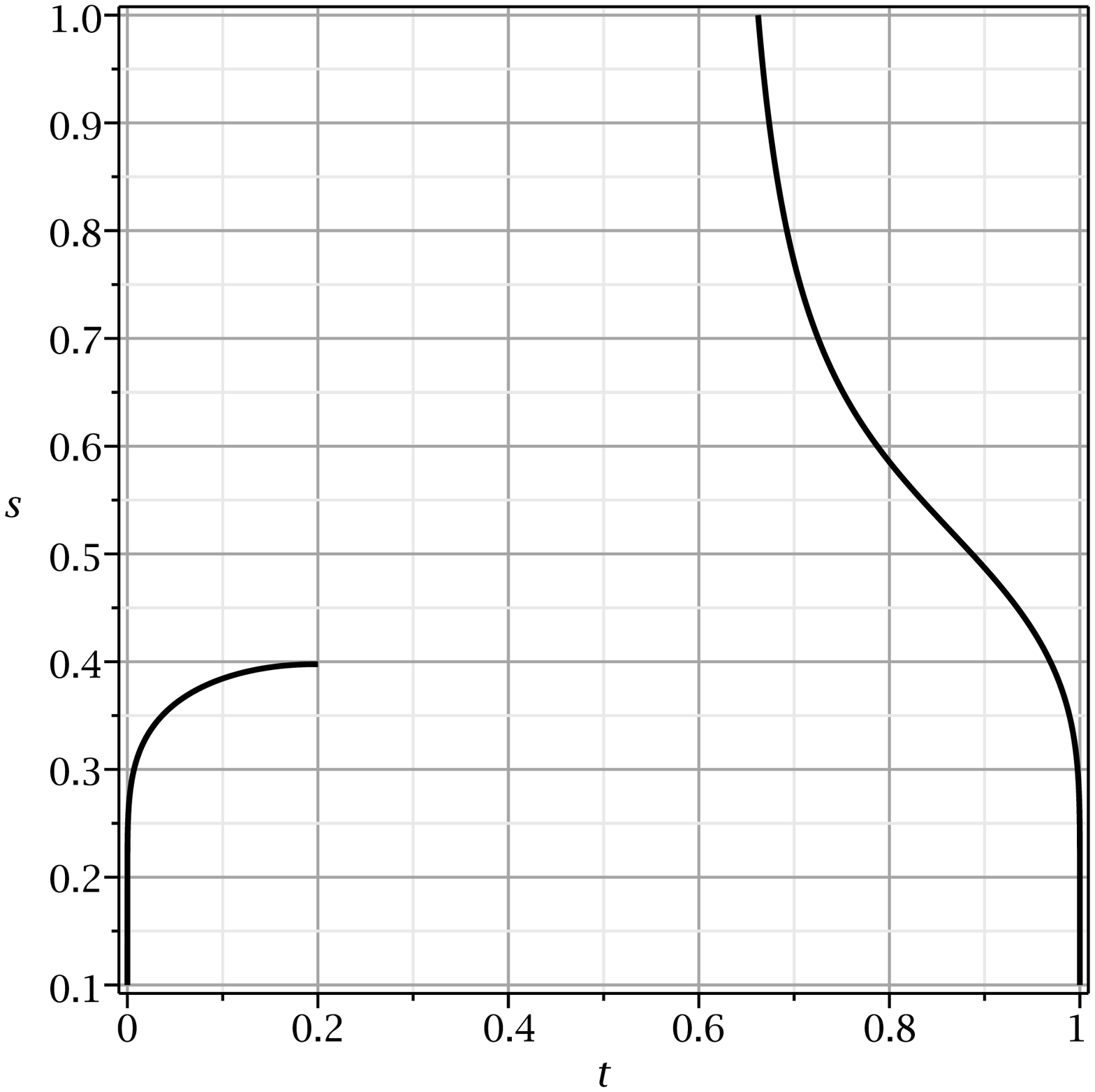}}\\
  \subfigure[Pattern 4: $\rho(t) {+} A\rho_1(t) {-}
  B\rho_{\beta}(t)$]{\includegraphics[height=1.8cm,width=6cm]
    {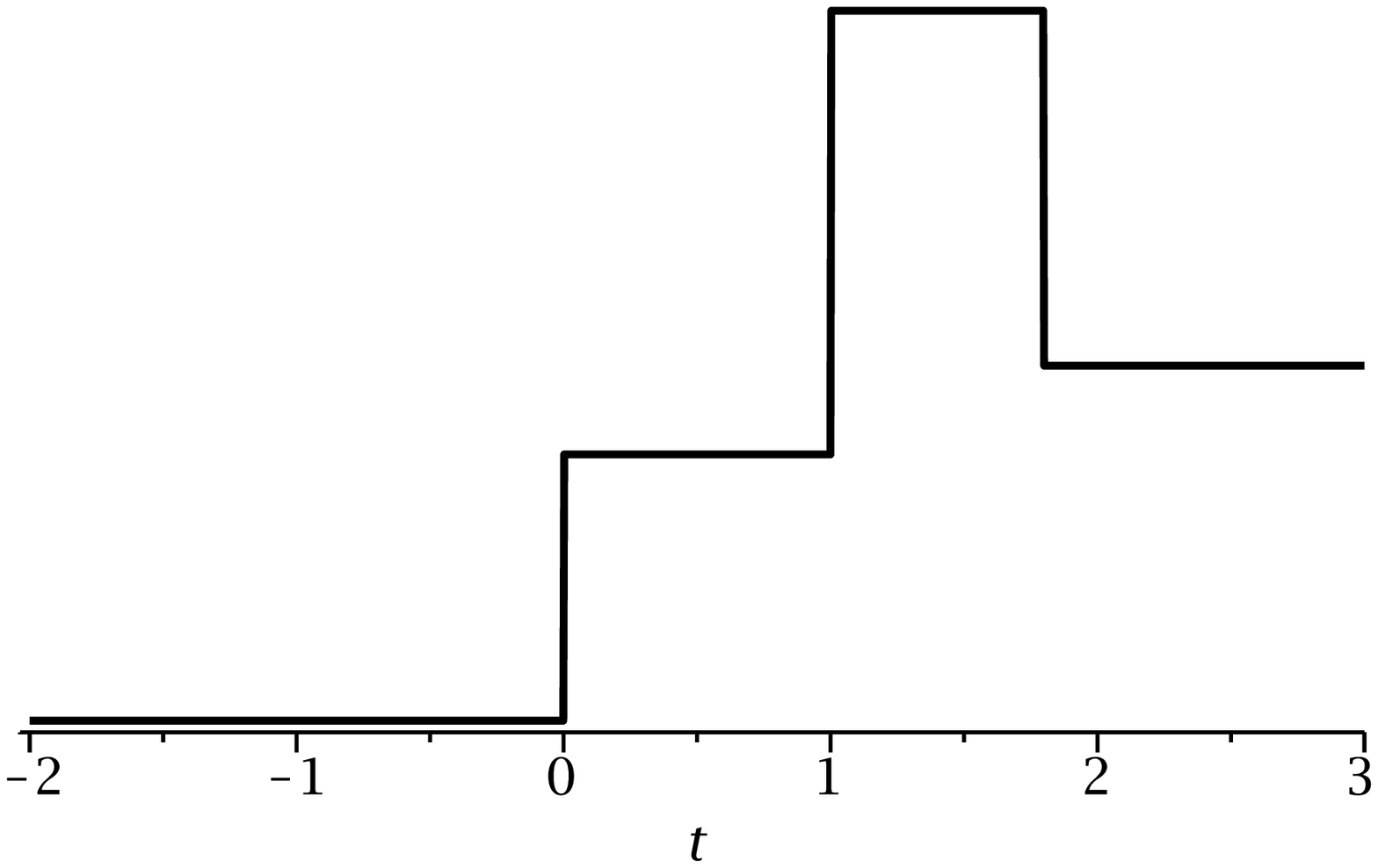}} \subfigure[Maxima-lines of pattern
  4]{\includegraphics[height=1.8cm,width=6cm]{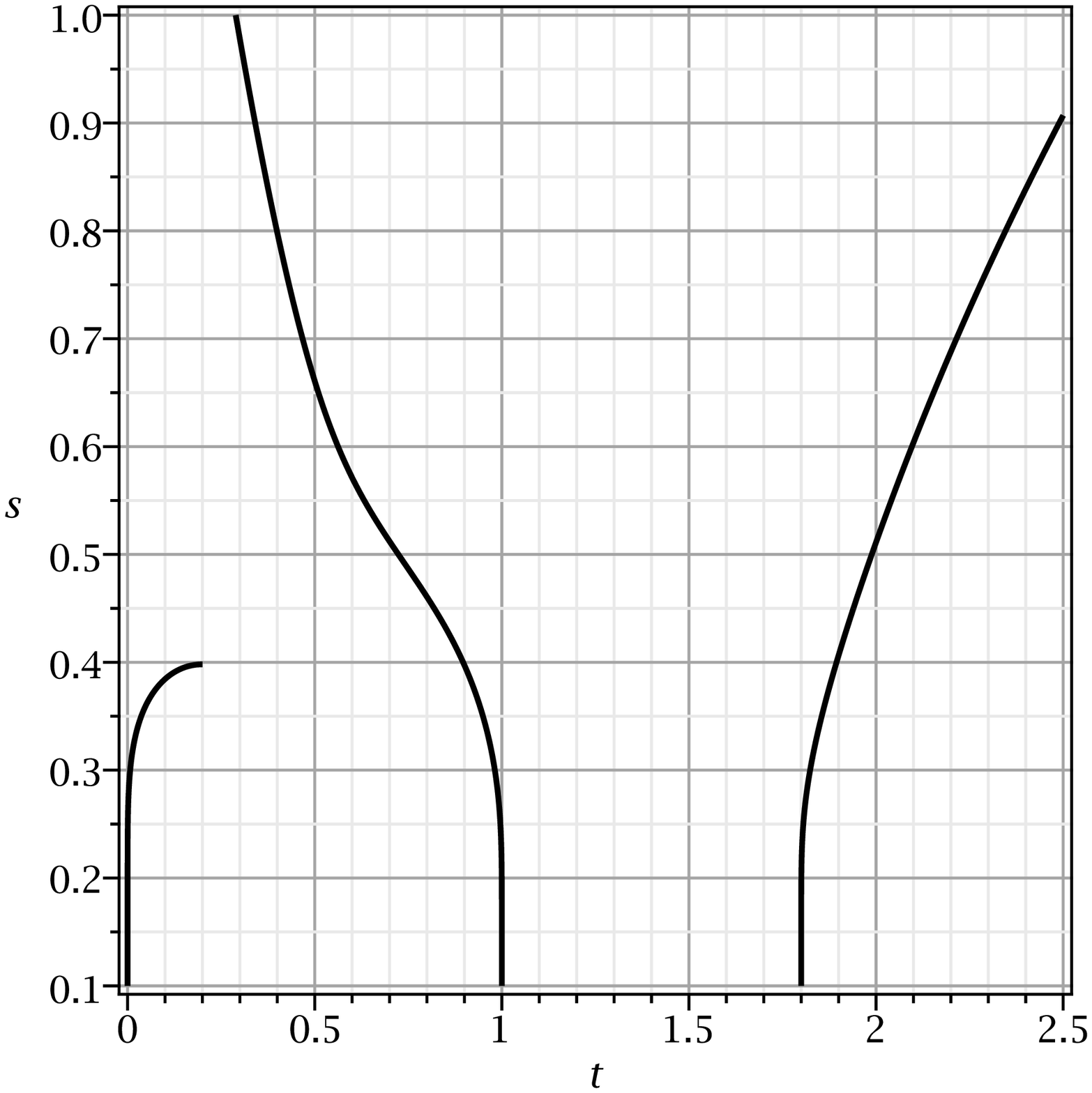}}\\
  \subfigure[Pattern 5: $\rho(t) {+} A\rho_1(t) {-}
  B\rho_{\beta}(t)$]{\includegraphics[height=1.8cm,width=6cm]
    {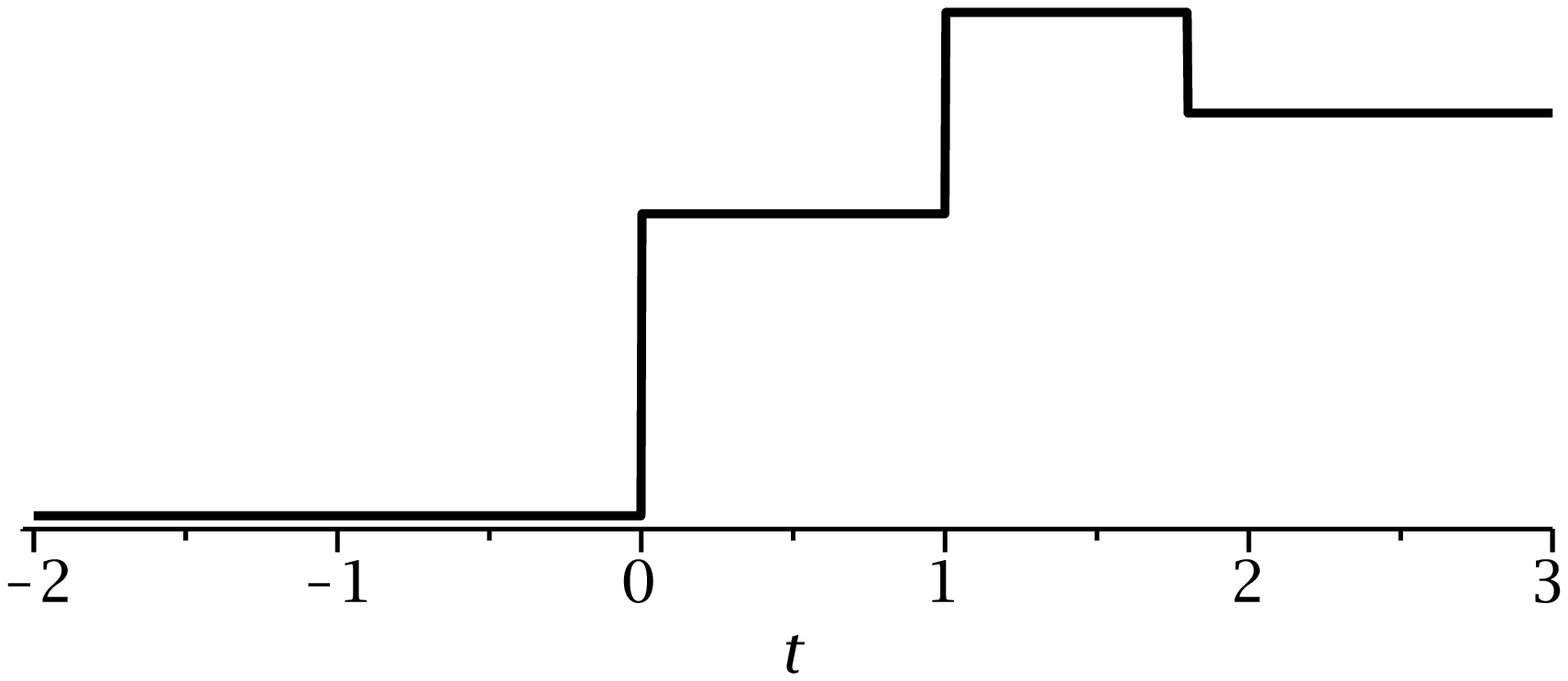}} \subfigure[Maxima-lines of pattern
  5]{\includegraphics[height=1.8cm,width=6cm] {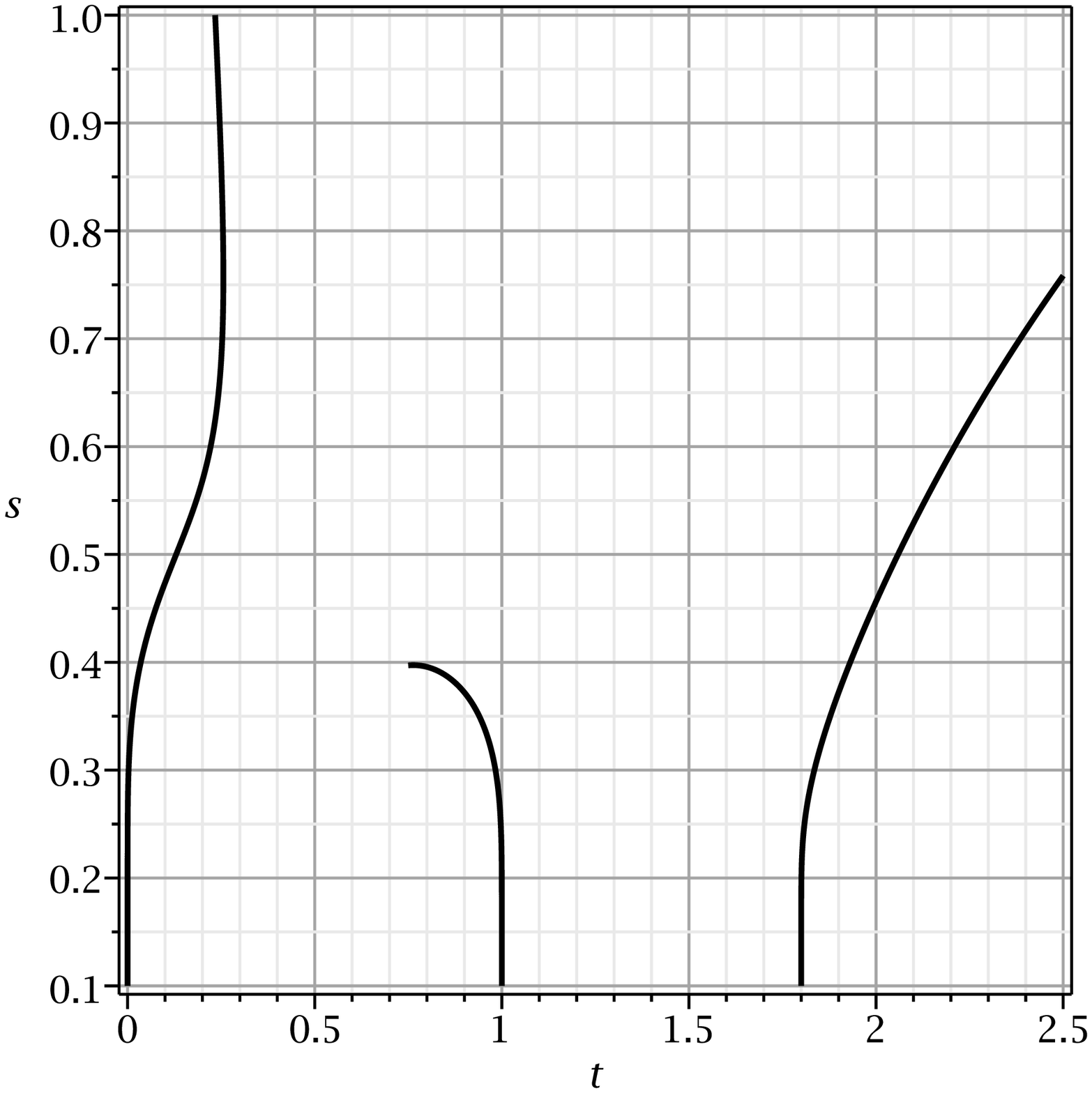}}\\
  \subfigure[Pattern 6: $\rho(t) {-} B\rho_{\beta}(t) {+}
  A\rho_1(t)$]{\includegraphics[height=1.8cm,width=6cm]
    {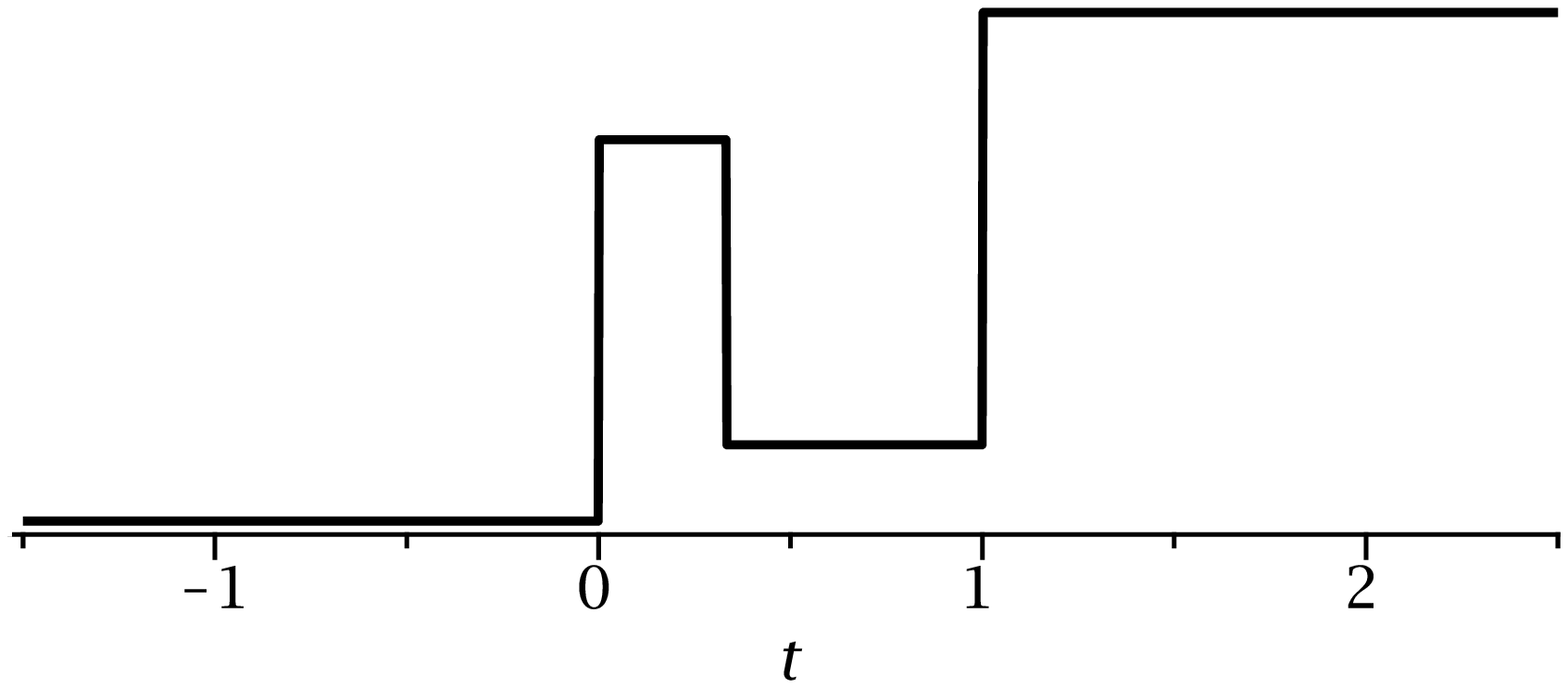}} \subfigure[Maxima-lines of pattern
  6]{\includegraphics[height=1.8cm,width=6cm] {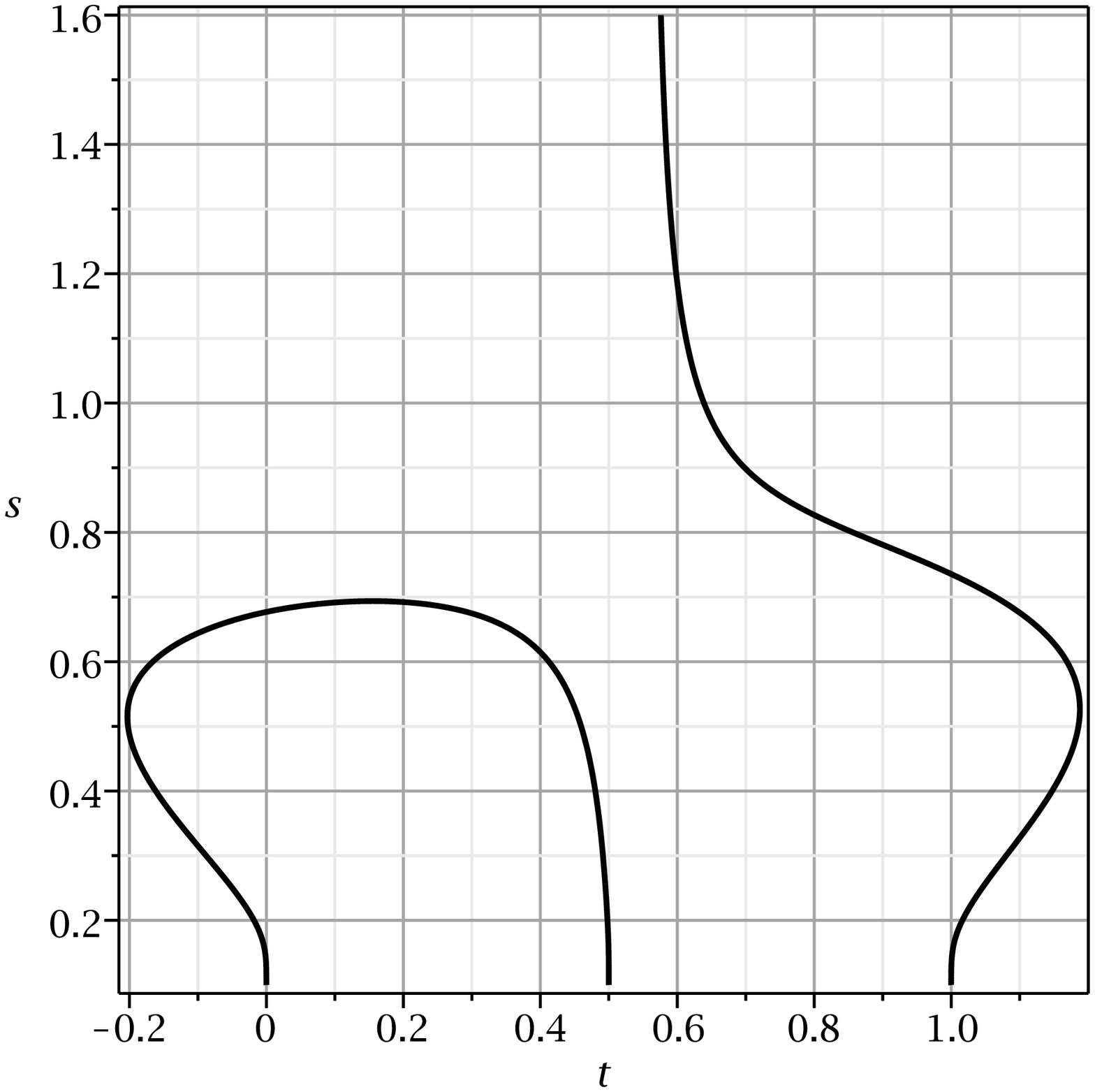}}
  \caption{The dictionary of patterns (model edges) used
    in our analysis. (a) step-edge (c) impulse-edge (e) staircase-edge
    (g)-(i)-(k) various triplet edges}\label{fig:dictionary}
\end{figure}
\indent The six patterns each model a particular type of edge. Pattern
1 models an isolated step-edge, that is, if the intensity at the
boundary of the object changes at a single jump. Pattern 2 models a
thin object or an impulse, while Pattern 3 models a boundary where the
intensity changes at two isolated jumps. Pattern 4 and pattern 5 model
an object with boundary similar to Pattern 3, but where the intensity
decreases shortly after the initial intensity-changes. This occurs for
instance if there is a 'void' within the object. If the object has a
'hole' near the boundary one can model this with Pattern 6.\\ \indent
Pattern 4 and pattern 5 differ by the time-scale behavior of the
maxima-lines, fig.\ref{fig:dictionary}(h,j). Given $A,\beta$ (or
$B,\beta$) the point of transformation between the patterns are given
by $B^*$ (or $A^*$). $B^*$ or $A^*$ can be found by solving;
\begin{equation}\label{eq:B*}
  \Bigg\{
  \begin{array}{lll}
    \frac{\partial Wf(u,s)}{\partial u} &= 0\\ \frac{\partial^2
      Wf(u,s)}{\partial u^2} &= 0\\ \frac{\partial^3
      Wf(u,s)}{\partial u^3} &= 0.
  \end{array}
\end{equation}
\indent Fig.\ref{fig:dictionary} shows typical examples of the
patterns and their respective maximal-lines. Every pattern can after a
suitable normalization and translation be assumed to have an edge at
$0$ with intensity $1$. Pattern 3-6 have (after rescaling) an
additional edge at $1$ with (positive) intensity $A$, and pattern
2,4-6 have an edge of (negative) intensity $B$ at $\beta\neq 0,1$. The
six patterns have one (pattern 1), two (pattern 2,3) and three
(pattern 4-6) maxima-lines, in what follows denoted $\ell_0$, $\ell_1$
and $\ell_{\beta}$.\\ \indent In fig.\ref{fig:dictionary} the six
patterns and characteristic examples of their maxima-lines are
displayed. As indicated in fig.\ref{fig:dictionary}h, one can expect
that distance between mod-max is a weak criterion for pattern
4. Rather surprisingly this is also the case for pattern 3. Observe
that the maxima-lines of these 'simple' patterns show a complexity
similar to those in medical ultrasound signals.\\ \indent A
common factor of pattern 3-6 is that there exists a scale where the
maxima-line $\ell_0$ (pattern 3,4,6), $\ell_1$ (pattern 5) or
$\ell_{\beta}$ (pattern 6) first appears. This scale is found by
solving the equations;
\begin{equation}\label{eq:s*}
  \Bigg\{
  \begin{array}{lll}
    \frac{\partial Wf(u,s)}{\partial u} &= 0\\ \frac{\partial^2
      Wf(u,s)}{\partial u^2} &= 0. 
  \end{array}
\end{equation}
The unique $s > 0$ for which \eqref{eq:s*} has a solution is denoted
$s^*$. As it happens, the suggested time-scale filtering procedure is often most
critical near $s^*$.
\subsection{Evaluation of the time-scale filtering
  procedure.}\label{SS:eval}
\begin{figure}[!h]
  \centering
  \subfigure[$\alpha = 1/2$]{\includegraphics[height=4cm,width=4cm]
  {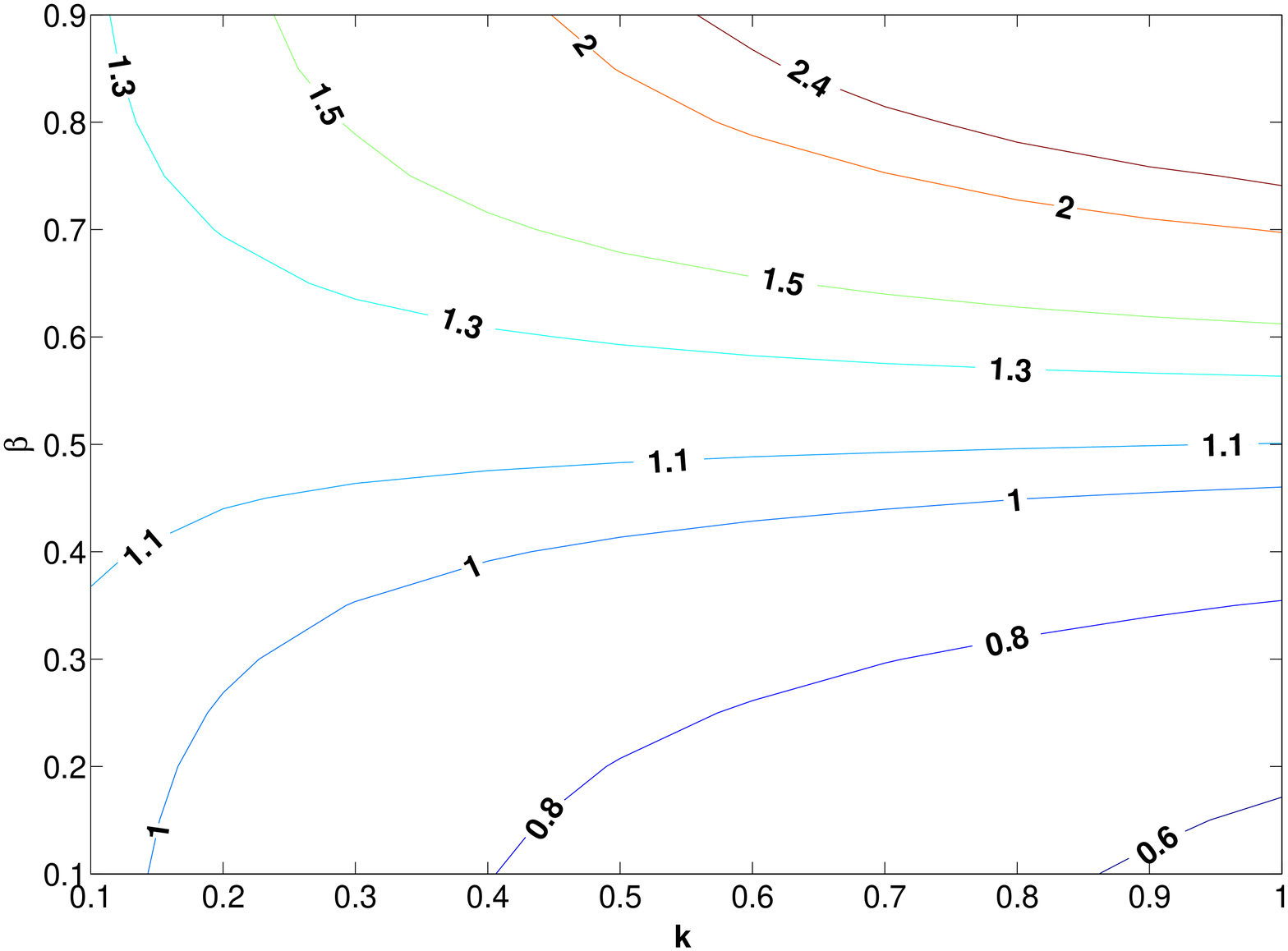}}
  \subfigure[$\alpha = -1/2$]{\includegraphics[height=4cm,width=4cm]
  {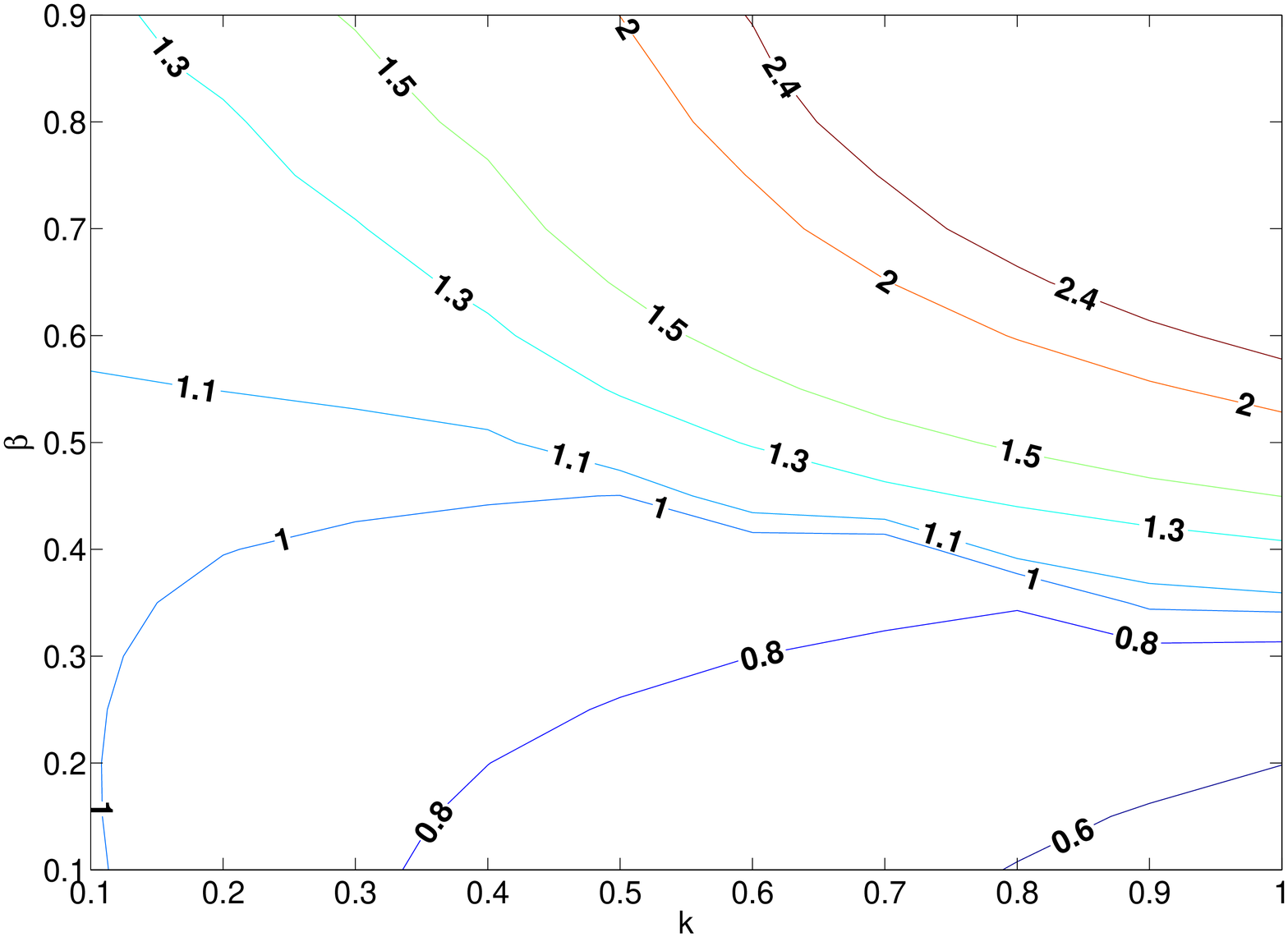}}
\caption{Numerical analysis of $Q_1$ for pattern 6. The figures
  display level curves of the surface $Q_1(A,B,\beta) = 0$ for
  $\mathrm{P}(n,m)$ given by \eqref{eq:P} with (a) $\alpha = 1/2$ and
  (b) $\alpha = -1/2$.}\label{fig:P6}
\end{figure}

\begin{figure}[!h]
  \centering \subfigure[$\alpha =
  -1/2$]{\includegraphics[height=4cm,width=4cm]
    {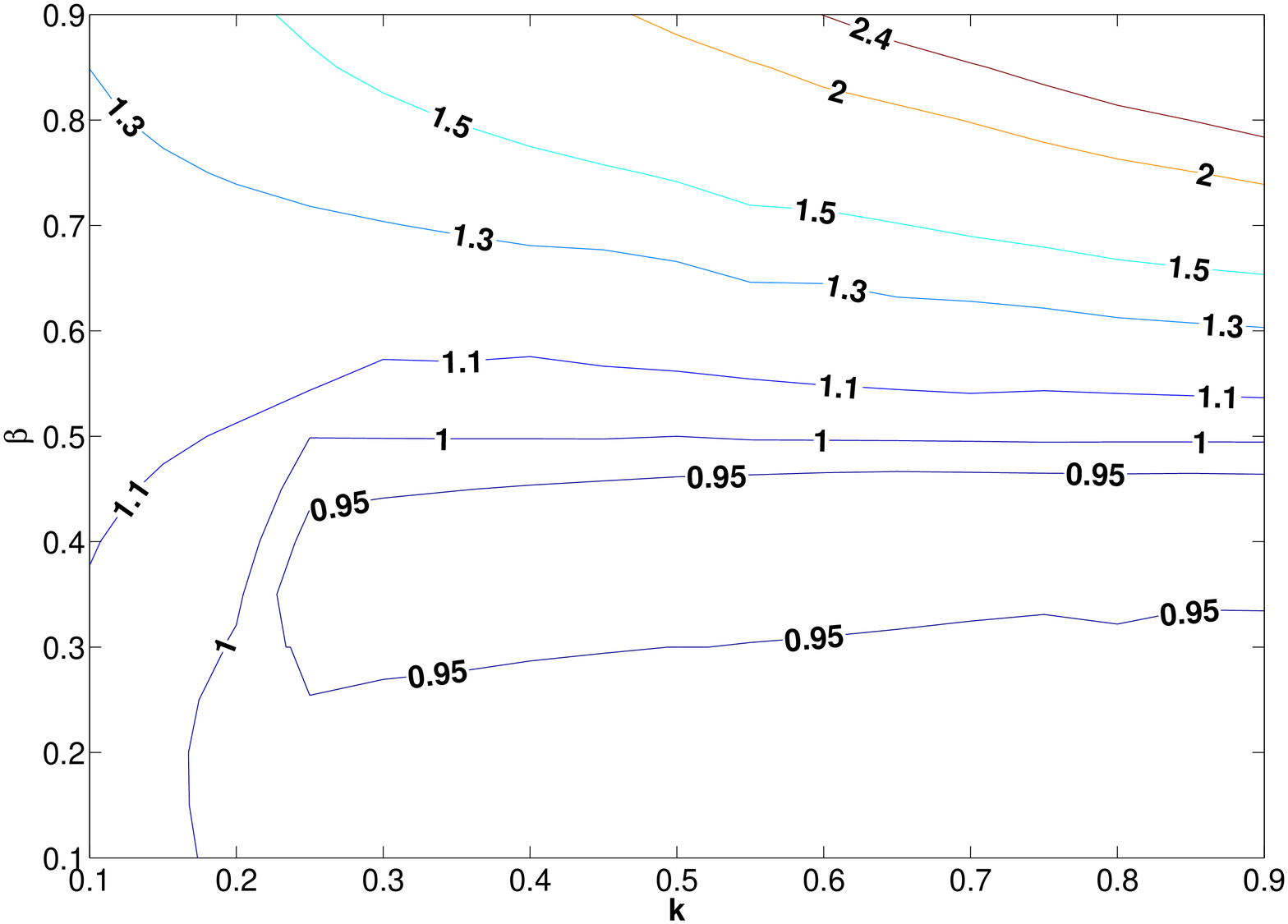}}
\caption{Numerical analysis of $Q_0$ for pattern 6. The figure display
  level curves of the surface $Q_0(A,B,\beta) = 0$ for
  $\mathrm{P}(n,m)$ given by \eqref{eq:P} with $\alpha =
  -1/2$.}\label{fig:P6fine}
\end{figure}

The purpose of the forthcoming analysis is to study for which patterns
(i.e. for which $A,B,\beta$) the time-scale filtering procedure
suggested in Sect.\ref{SS:P} works. We also study the procedure for
different control-parameters $\alpha$.\\ \indent Let as
before $\ell_0$, $\ell_1$ and $\ell_{\beta}$ denote the maxima-lines
in the patterns from Sect.\ref{SS:dictionary} and fix two scales $s_2$
and $s_1$. We chose in this study $s_1 \,{=}\, s_2/2$, i.e. the scales
are dyadic distributed. In what follows let $\mathrm{P}(\cdot,\cdot)$
be the decision-function corresponding to $s_2$ and $s_1$ defined in
Sect.\ref{SS:P}.\\ \indent It is straightforward to see that the
procedure works for the first two patterns, hence we focus our
analysis on patterns $3 {-} 6$. The decision-function
$\mathrm{P}(\cdot,\cdot)$ needs to fulfill one inequality for each
maxima-line. Namely; the decision-function connects two mod-max
$\ell_1(s_2)$ and $\ell_1(s_1)$ from the maxima-line $\ell_1$ if;
\begin{small}
  \begin{equation}
    \mathrm{P}\big(\ell_1(s_2),\ell_1(s_1)\big) -
    \mathrm{P}\big(\ell_1(s_2),\ell_j(s_1)\big) > 0\qquad 
    \mathrm{for}\;j = 0, \beta. \label{eq:P1}
  \end{equation}
\end{small}
Similarly the mod-max $\ell_0(s_2)$ and $\ell_0(s_1)$ are connected if;
\begin{small}
  \begin{equation}
    \mathrm{P}\big(\ell_0(s_2),\ell_0(s_1)\big) -
    \mathrm{P}\big(\ell_0(s_2),\ell_j(s_1)\big) > 0\qquad 
    \mathrm{for}\;j = 1, \beta. \label{eq:P0}
  \end{equation}
\end{small}
Note that for pattern 3 the inequalities \eqref{eq:P1} and
\eqref{eq:P0} have to hold only for $j \,{=}\, 0$ and $j \,{=}\, 1$
respectively since $\ell_{\beta}$ does not exist. Finally, for pattern
4-6 the suggested procedure connects two mod-max from the maxima-line
$\ell_{\beta}$ if;
\begin{small}
  \begin{equation}
    \mathrm{P}\big(\ell_{\beta}(s_2),\ell_{\beta}(s_1)\big) -
    \mathrm{P}\big(\ell_{\beta}(s_2),\ell_j(s_1)\big) > 0\qquad \mathrm{for}\;j
    = 0, 1. \label{eq:Pb}
  \end{equation}
\end{small}
Inequality \eqref{eq:Pb} always holds for pattern 4-6 since
$\mathrm{Sign}\big(\ell_{\beta}(s_2) , \ell_{j}(s_1)\big) \,{=}\, 0$
only if $j \,{\neq}\, \beta$ \big(since $Wf\big(\ell_{\beta}(s),s\big)
\,{<}\, 0$ and $Wf\big(\ell_j(s),s\big) \,{>}\, 0,\,j \,{=}\,
0,1$\big).  We will therefore focus the
analysis on the inequalities \eqref{eq:P1} and \eqref{eq:P0}.\\
\indent For pattern 3,4,6 the decision-function successfully connects
any two mod-max if \textit{both} inequalities \eqref{eq:P1}
\textit{and} \eqref{eq:P0} hold for \textit{all} scales such that
$\ell_0$ exists. Consequently \eqref{eq:P1} has to hold for every $0
{<} s_1 {<} s^*$ and \eqref{eq:P0} for all $0 {<} s_2 {<} s^*$. For
pattern 5 \textit{both} \eqref{eq:P1} and \eqref{eq:P0} have to hold
for \textit{all} scales such that $\ell_1$ is well-defined. Thus
\eqref{eq:P1} has to hold for
every $0 {<} s_2 {<} s^*$ and \eqref{eq:P0} for all $0 {<} s_1 {<} s^*$.\\
\textbf{Comment:} \textit{The preceding analysis of the
  decision-function $\mathrm{P}(\cdot,\cdot)$ focuses on patterns with
  values of $A,B,\beta$ typical for our specific application (locally
  model medical ultrasound signals). Values of $A, B$ and/or $\beta$
  near the 'transformation-limit' (value of $A,B,\beta$ where a
  pattern transforms into another) will not be studied. In practice
  one can
  often consider such 'critical' configurations by another pattern.}\\ \\
\indent \textit{Pattern 3:} \indent First we consider the
decision-function and corresponding time-scale filtering procedure
with respect to \textit{pattern 3}. Numerical analysis suggests that
the time-scale filtering procedure is most likely to fail at the
largest scale where both $\ell_1$ and $\ell_0$ exist. We can therefore
simplify the analysis by considering the left-hand-side (lhs.)  of
\eqref{eq:P1} with $s_1 \,{=}\, s^*$, $s_2 = 2s_1$ and lhs. of
\eqref{eq:P0} with $s_2 \,{=}\, s^*$, $s_1 = s_2/2$. Note that the
lhs. of \eqref{eq:P1} and \eqref{eq:P0} - denoted $Q_1(A)$ and
$Q_0(A)$ - are now functions depending \textit{only} on $A$. In
particular, if both $Q_1(A) > 0$ and $Q_0(A) > 0$ the suggested
procedure holds.\\
\begin{figure}[!t]
  \centering 
  \subfigure[]{\includegraphics[height=3cm,width=5.2cm]
  {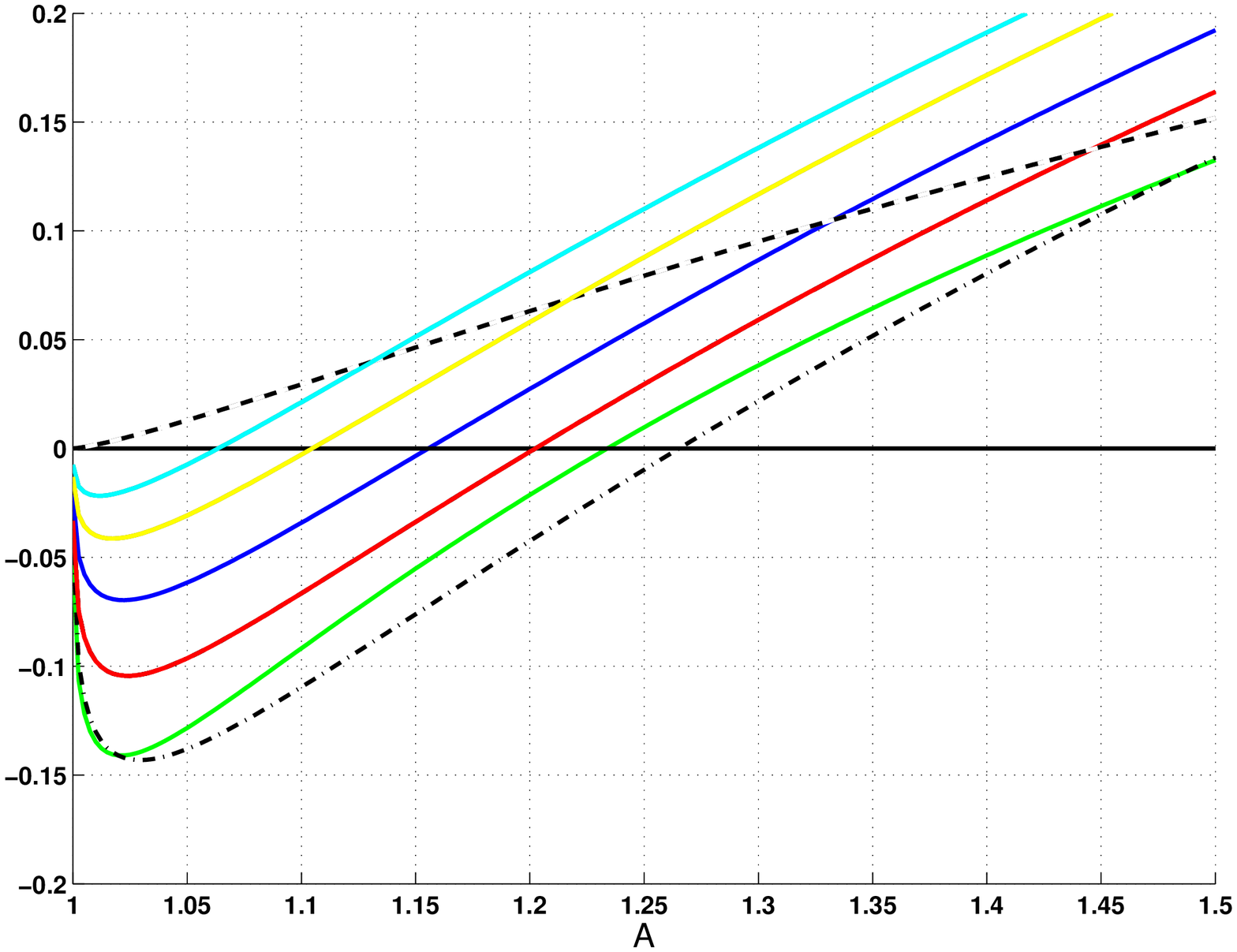}}
  \subfigure[]{\includegraphics[height=3cm,width=5.2cm]
  {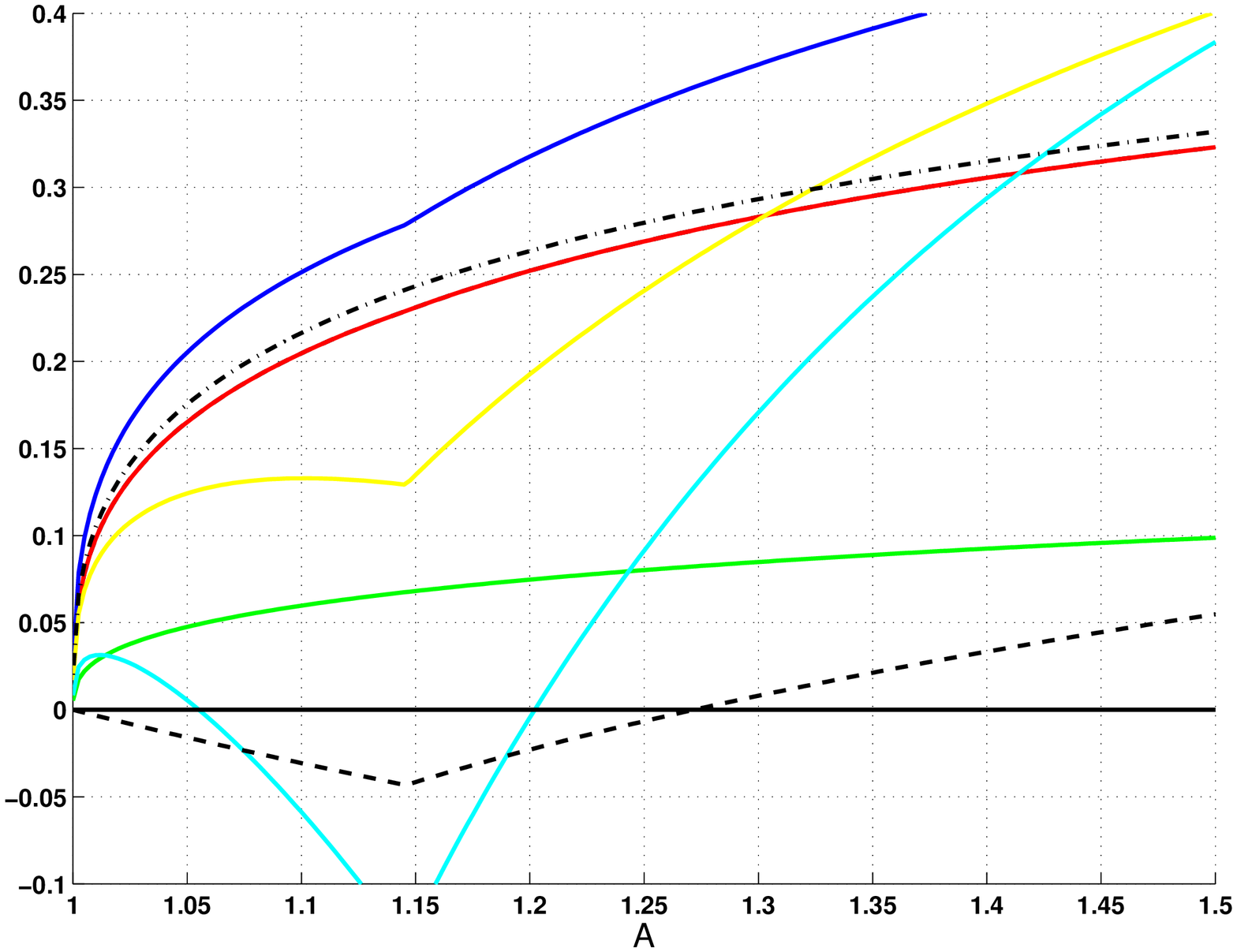}}
\caption{(a) The graph of $Q_1(A)$ and (b) the graph of $Q_0(A)$ for
  $1 < A < 1.5$. The cyan, yellow, blue, red, green curves correspond
  to $\mathrm{P}(n,m)$ defined by \eqref{eq:P} with respectively
  $\alpha = \{-1/2, 0, 1/2, 1, 3/2 \}$ in the expression of
  $\mathrm{P}$. The dashed line corresponds to $\mathrm{P} =
  \mathrm{D}$ and dash-dotted line corresponds to $\mathrm{P} =
  \Delta$.}\label{fig:P3}
\end{figure}
\indent Fig.\ref{fig:P3} displays the graphs of respectively $Q_1(A)$
and $Q_0(A)$ for $\mathrm{P}(\cdot,\cdot)$ given by \eqref{eq:P} with
control-parameters $\alpha {=} \{-1/2, 0, 1/2, 1, 3/2\}$ (cyan,
yellow, blue, red, green). The dashed and dash-dotted lines correspond
to the 'limit cases' resp. $\mathrm{P}(n,m) {=} \mathrm{D}(n,m)$ and
$\mathrm{P}(n,m) {=} \Delta(n,m)$.\\ \indent From fig.\ref{fig:P3}a
one can see that along the maxima-line $\ell_1$ the decay-criterion
(dotted line) is optimal for connecting mod-max; if $P(n,m) \,{=}\,
D(n,m)$ then $Q_1(A) \,{>}\, 0$ (i.e. \eqref{eq:P1} holds) for all
$A>1$. Along $\ell_0$ one need to have $A \,{\gtrsim}\, 1.3$,
fig.\ref{fig:P3}b. On the contrary; if one considers the
distance-criterion (dash-dotted line) it suffices to have $A \,{>}\,
1$ at fine scales, but needs to have $A \,{\gtrsim}\, 1.3$ at coarse
scales. Consequently one must have $A \,{\gtrsim}\, 1.3$ in order to
connect mod-max between any two scales $s_2 \,{=}\, 2s_1$ and $0
\,{<}\, s_1 \,{<}\, s^*$ for pattern 3 if either $\mathrm{P}(n,m)
\,{=}\, \mathrm{D}(n,m)$ or $\mathrm{P}(n,m) \,{=}\, \Delta(n,m)$.\\
\indent Combining distance and decay increase the $A$-interval for
which the time-scale filtering procedure works, fig.\ref{fig:P3}. If
e.g. $\mathrm{P}(n,m)$ is given by \eqref{eq:P} with $\alpha \,{=}\,
0$ (yellow line) then it suffices to have $A \,{\gtrsim}\, 1.1$ to
connect mod-max between (any) two scales in pattern 3.  If $\alpha
\,{=}\, 1/2$ (blue line) then both $Q_1(A)
\,{>}\, 0$ and $Q_0(A) \,{>}\, 0$ for all $A \,{\gtrsim}\, 1.16$.\\
\begin{figure}[!h]
  \centering \subfigure[$\alpha =
  1/2$]{\includegraphics[height=4cm,width=4cm] {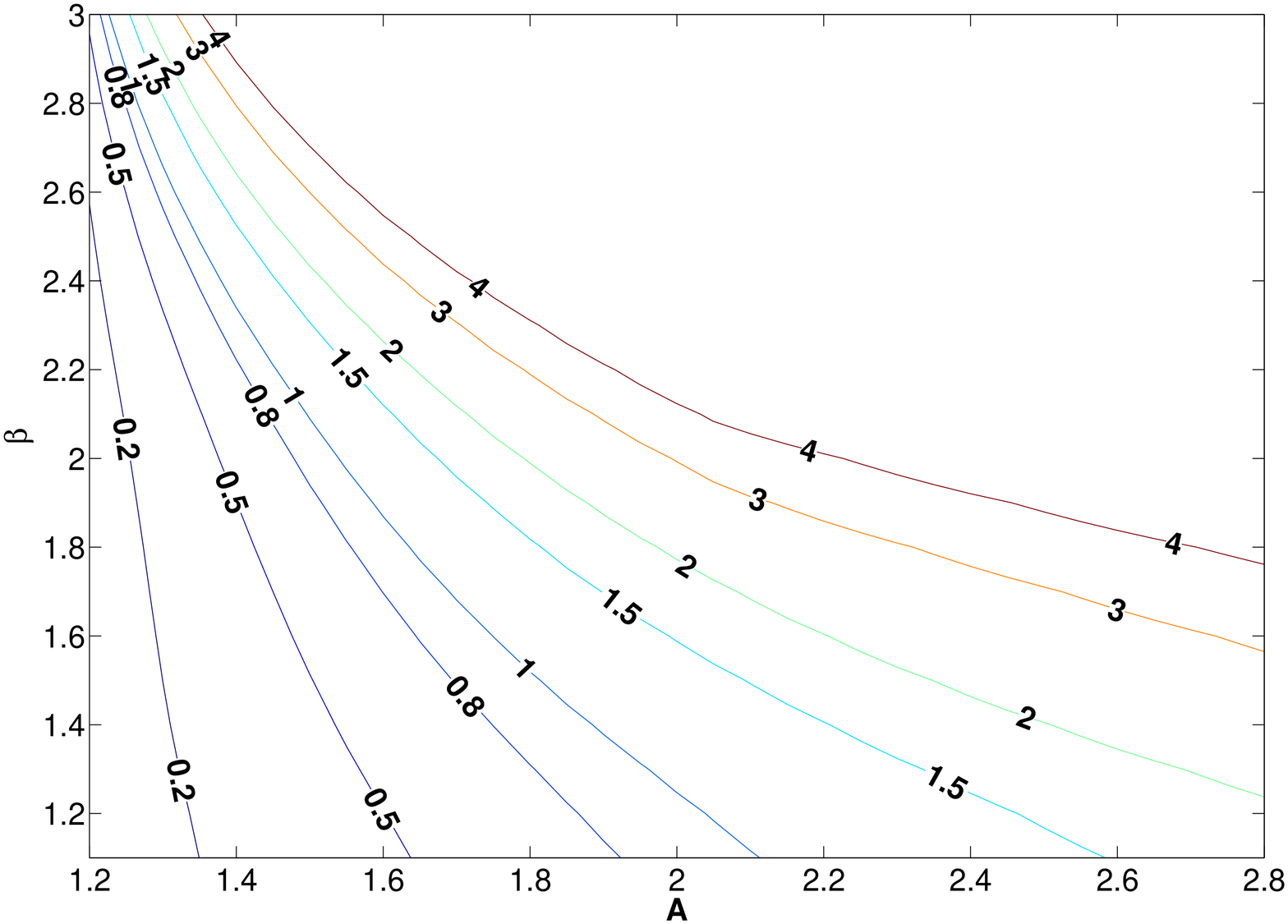}}
  \subfigure[$\alpha = -1/2$]{\includegraphics[height=4cm,width=4cm]
    {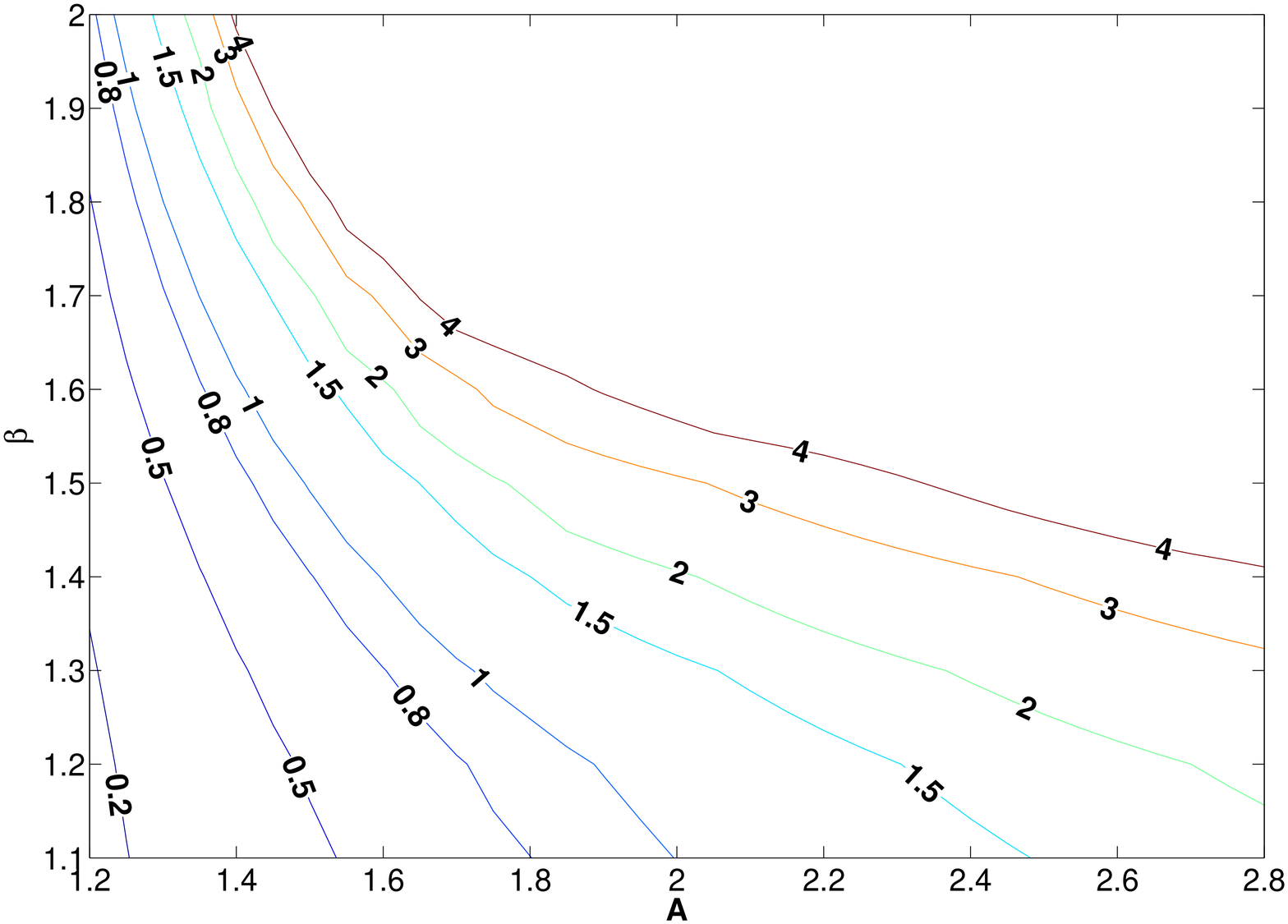}}
\caption{Numerical analysis of $Q_1$ for pattern 4. The figures
  display level curves of the surface $Q_1(A,B,\beta) = 0$ for
  $\mathrm{P}(n,m)$ given by \eqref{eq:P} with (a) $\alpha = 1/2$ and
  (b) $\alpha = -1/2$. NOTE that in (a) we consider $1.1 < \beta < 3$
  and in (b) $1.1 < \beta < 2$.}\label{fig:P4}
\end{figure}
\begin{figure}[!h]
  \centering \subfigure[$\alpha =
  -1/2$]{\includegraphics[height=4cm,width=4cm]
    {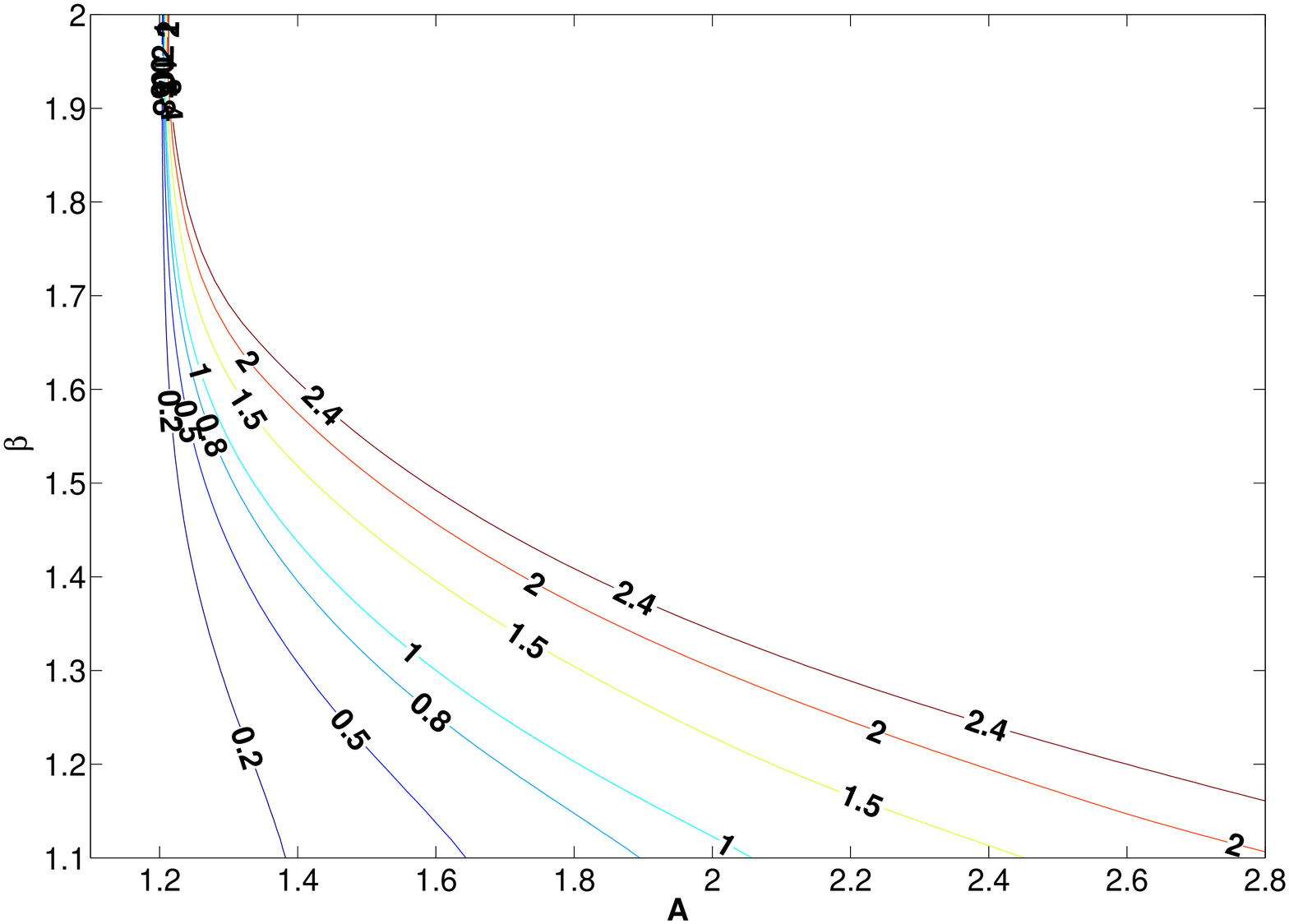}}
\caption{Numerical analysis of $Q_0$ for pattern 4. The figure
  displays some level-curves of the surface $Q_0(A,B,\beta) = 0$ for
  $\mathrm{P}(n,m)$ given by \eqref{eq:P} with $\alpha =
  -1/2$.}\label{fig:P4fine}
\end{figure}
\indent \textit{Pattern 4:} \indent For \textit{pattern 4} there are
three maxima-lines, $\ell_0$, $\ell_1$ and $\ell_{\beta}$. As
illustrated in fig.\ref{fig:dictionary}h the edge at $\beta$ 'repels'
the maxima-lines corresponding to the edges at $0$ and $1$. The result
is that the distance between $\ell_0$ and $\ell_1$ gets smaller, hence
weakening the distance-criterion for pattern 4.\\ \indent For pattern
4 the time-scale filtering procedure successfully connect mod-max
across scales if both \eqref{eq:P1} and \eqref{eq:P0} holds. For this
purpose let $Q_1(A,B,\beta)$ denote the minimum of the lhs. of
\eqref{eq:P1} with respect to $s_2 \,{=}\, 2s_1$, $0 \,{<}\, s_1
\,{<}\, s^*$, $j \,{=}\, 0$ and $0 \,{<}\, B \,{<}\, B^*$, where $B^*$
is given by \eqref{eq:B*} (recall that pattern 4 transforms into
pattern 5 for $B \,{>}\, B^*$). Similarly let $Q_0(A,B,\beta)$ denote
the minimum of the lhs. of \eqref{eq:P0} with $s_1 \,{=}\, 2s_2, 0
\,{<}\, s_2 \,{<}\, s^*$, $j\,{=}\,1$ and $0 \,{<}\, B \,{<}\,
B^*$. If \textit{both} $Q_1(A,B,\beta) \,{>}\, 0$ \textit{and}
$Q_0(A,B,\beta) \,{>}\, 0$ the procedure
successfully connects any two mod-max with respect to pattern 4.\\
\indent The additional variables $B$ and $\beta$ make the analysis of
$Q_1$ and $Q_0$ more complicated for pattern 4-6 than for pattern
3. Rather than considering the graphs of $Q_1$ and $Q_0$ we study the
(level-curves of the) surfaces $Q_i(A,B,\beta) \,{=}\, 0$,
$i=0,1$. Numerical analysis indicates that both $Q_1$ and $Q_0$
increase if: \textit{(i):} the jump at location $1$ increases,
\textit{(ii):} the jump at $\beta$ decreases, or \textit{(iii):} the
distance from $1$ to $\beta$ increases. In particular, if
$Q_1(A_0,B_0,\beta_0) \,{=}\, 0$ then $Q_1(A,B,\beta) \,{>}\, 0$ for
all $A \,{>}\, A_0, B_0 \,{>}\, B
\,{>}\, 0$ and $\beta \,{>}\, \beta_0$, and similarly for $Q_0$.\\
\indent As with pattern 3; different values of the control-parameter
$\alpha$ 'interpolates' $\mathrm{P}(\cdot,\cdot)$ given by
\eqref{eq:P} in between the 'limit' cases $\mathrm{P}(n,m) \,{=}\,
\mathrm{D}(n,m)$ and $\mathrm{P}(n,m) \,{=}\,
\Delta(n,m)$. Fig.\ref{fig:P4} displays the level-curves of the
surface $Q_1(A,B,\beta) = 0$ with respect to $\mathrm{P}(n,m)$ given
by \eqref{eq:P} with $\alpha \,{=}\,\pm1/2$ (note that the y-axis
differs in the figures). Fig.\ref{fig:P4fine} displays the
level-curves of $Q_0(A,B,\beta) = 0$ for $\alpha \,{=}\, {-}
1/2$. Numerical studies indicate that $Q_0(A,B,\beta) \,{>}\, 0$ if
$\alpha \,{=}\, 1/2$, except possibly near critical values of
$A,B,\beta$ such as e.g. $\beta \,{\approx}\, 1$ (which are
uninteresting to study for our purpose).\\ \indent As expected;
increasing the influence of the decay-criterion increases the accuracy
of the time-scale filtering procedure for pattern 4. The following
table illustrates this for a few concrete values of $A$ and $\beta$.
The table shows the maximal $B$-interval such that the time-scale
filtering procedure works for some values of $A$ and $\beta$;
\begin{small}
  \begin{center}
    \begin{tabular}{|c|c|c|c|c|}
      \hline A & $\beta$ & $\alpha = -1/2$ & $\alpha = 1/2$ &
      $\mathrm{P} = \Delta$\\  
      \hline $2$   & $1.6$     & $0 < B < 4.57$  & $0 < B < 1.52$ & $0
      < B < 0.83$\\  
      \hline $1.5$ & $1.6$     & $0 < B < 1.35$  & $0 < B < 0.54$ & $0
      < B < 0.25$\\ 
      \hline
    \end{tabular}
  \end{center}
\end{small}
As one can observe in the table, the reliability of the time-scale
filtering procedure increases with respect to pattern 4 if we increase
the influence of the decay-criterion.\\
\begin{figure}[!h]
  \centering
  \subfigure[$\alpha = 1/2$]{\includegraphics[height=4cm,width=4cm]
  {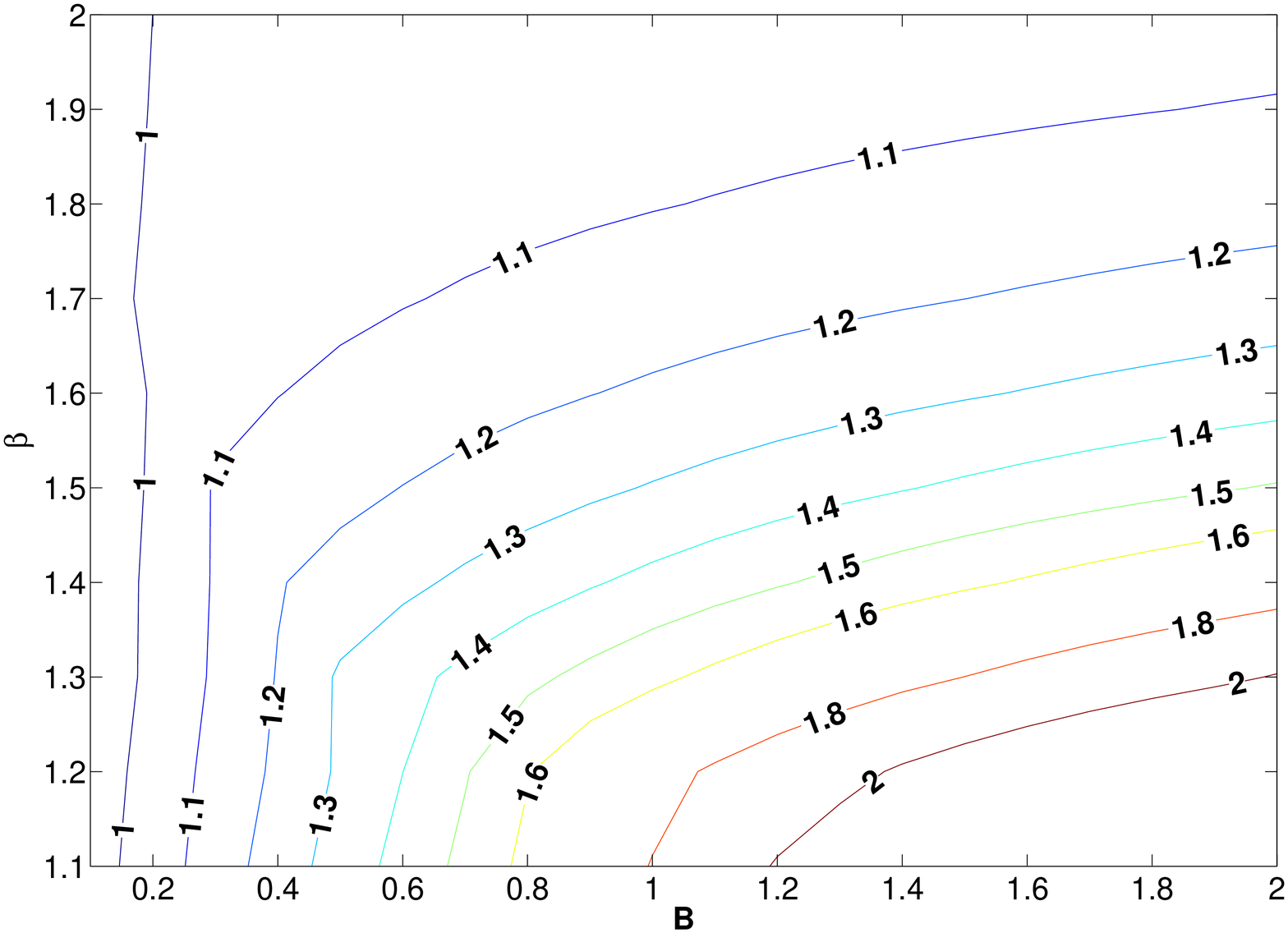}}
  \subfigure[$\alpha = -1/2$]{\includegraphics[height=4cm,width=4cm]
  {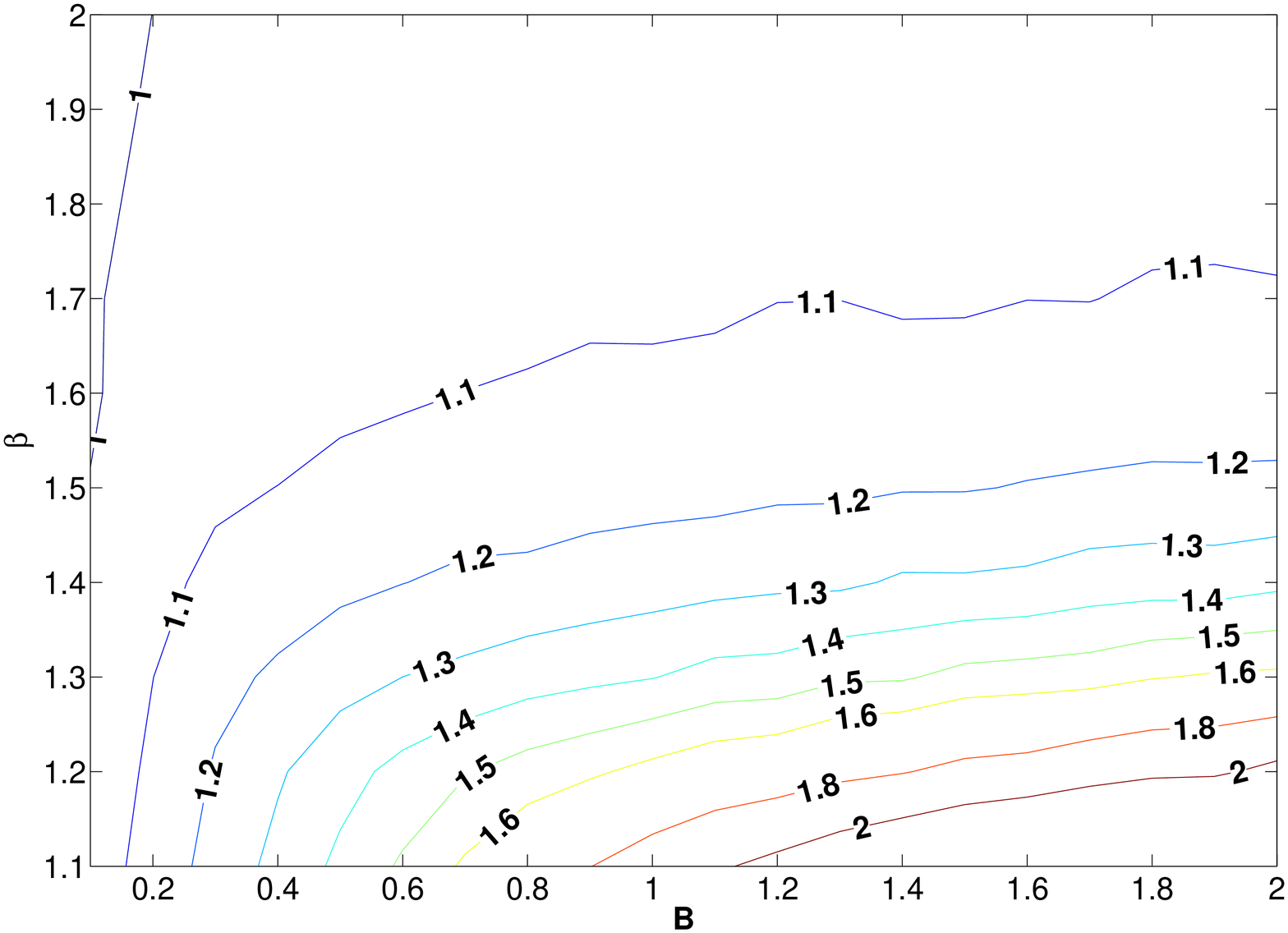}}
\caption{Numerical analysis of $Q_0$ for pattern 5. The figures
  display level curves of the surface $Q_0(A,B,\beta) = 0$ for
  $\mathrm{P}(n,m)$ given by \eqref{eq:P} with (a) $\alpha = 1/2$ and
  (b) $\alpha = -1/2$.}\label{fig:P5}
\end{figure}
\begin{figure}[!h]
  \centering 
  \subfigure[$\alpha =
  1/2$]{\includegraphics[height=4cm,width=4cm]
    {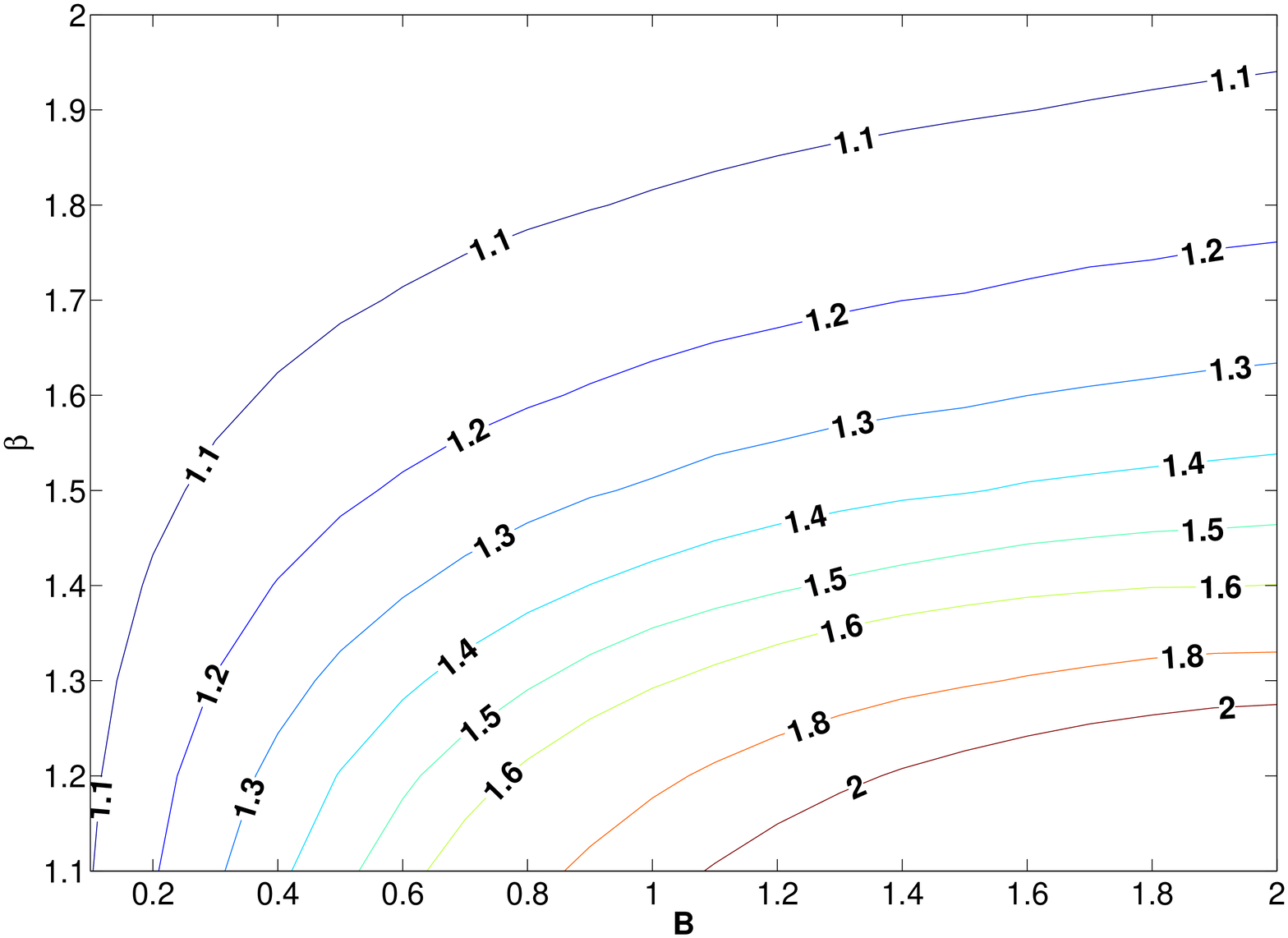}}
  \subfigure[$\alpha =
  -1/2$]{\includegraphics[height=4cm,width=4cm]
    {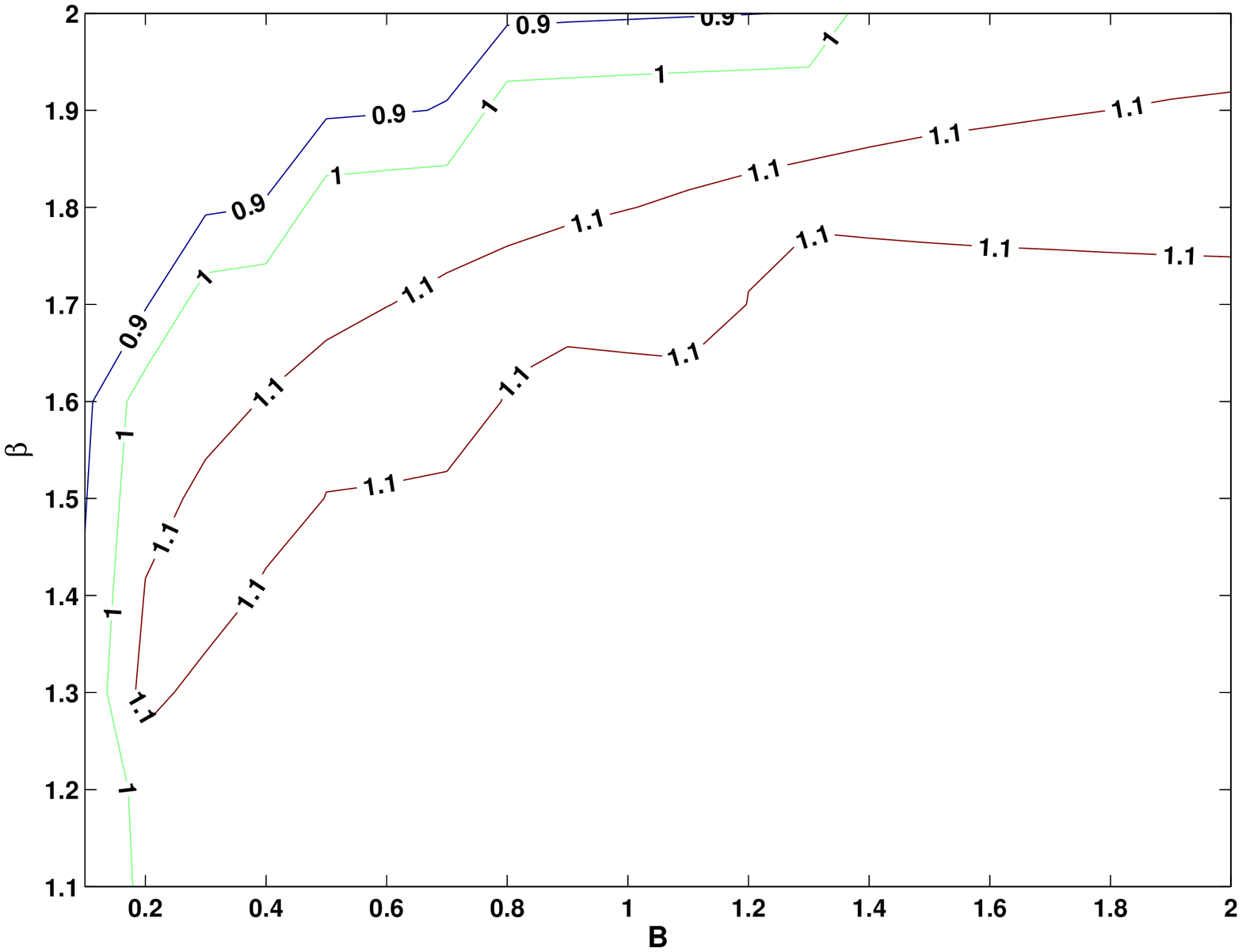}}  
  \caption{Numerical analysis of $Q_1$ for pattern 5. The figures
    display level curves of the surface $Q_1(A,B,\beta) = 0$ for
    $\mathrm{P}(n,m)$ given by \eqref{eq:P} with (a) $\alpha = 1/2$
    and (b) $\alpha = -1/2$.}\label{fig:P5fine}
\end{figure}
\indent \textit{Pattern 5:}\indent For \textit{pattern 5} the
situation is similar as with pattern 4. The edge at $\beta$ will - as
before - 'push' the other maxima-lines away. However, since the 'long'
maxima-line now approaches $0$ (and not $1$ as for pattern 4,
fig.\ref{fig:dictionary}(h,j)), we expect that the distance-criterion
will be strengthened.\\ \indent Let as before $Q_1(A,B,\beta)$ denote
the minimum of the lhs. of \eqref{eq:P1} for $0 \,{<}\, s_2 \,{<}\,
s^*,\,s_2 \,{=}\, 2s_1,\,j\,{=}\,0$ and $0 \,{<}\, A \,{<}\, A^*$.
Given $B,\beta$ one can find $A^*$ by solving \eqref{eq:B*}. Let
$Q_0(A,B,\beta)$ denote the minimum of the lhs. of \eqref{eq:P0} for
$0 \,{<}\, s_1 \,{<}\, s^*,\,s_2 \,{=}\, 2s_1,\,j \,{=}\, 1$ and $0
\,{<}\, A \,{<}\, A^*$.\\ \indent Fig.\ref{fig:P5} and
fig.\ref{fig:P5fine} show some level-curves of the surfaces
$Q_i(A,B,\beta) \,{=}\, 0,\,i=0,1$ for $\mathrm{P}(n,m)$ defined in
\eqref{eq:P} with $\alpha \,{=}\, \pm 1/2$. Given the pair $(B,\beta)$
in the $B\beta$-plane the point $(B,\beta,A)$ on the surface
$Q_i(A,B,\beta) = 0$ represents the maximal value of $A$ such that the
procedure successfully connects mod-max along the maxima-line $\ell_i$
for $i=0,1$.\\ \indent One may observe that the reliability of the
decision function $\mathrm{P}(\cdot,\cdot)$ increases if we increase
the weighting of the distance-criterion. Indeed; $Q_i(A,B,\beta)
\,{>}\, 0$, $i\,{=}\,0,1$ for 'almost' all $0 \,{<}\, A \,{<}\, A^*$ if
$\mathrm{P}(n,m) \,{=}\, \Delta(n,m)$ or if $\mathrm{P}(n,m)$ is given
by \eqref{eq:P} with $\alpha \,{=}\, 1/2$. On the other hand, if
$\alpha = -1/2$ one can roughly say that one must have $A
\,{\lesssim}\, 1$ to ensure that both $Q_1 > 0$ and $Q_0 > 0$.\\ 
\indent \textit{Pattern 6:} \indent Fig.\ref{fig:P6} and
fig.\ref{fig:P6fine} show a few examples of the analysis of the
decision-function with respect to \textit{pattern 6}. Similar to the
previous patterns let $Q_1(A,B,\beta)$ denote the minimum of the
lhs. of \eqref{eq:P1} wrt. $s_2 \,{=}\, 2s_1,\, 0 \,{<}\, s_1 \,{<}\,
s^*,\,j \,{=}\, 0$ and $A \,{>}\, A^*$. Given $(B,\beta)$ one can find
$A^*$ by solving \eqref{eq:B*} (If $A = A^*$ the maxima-line at coarse
scales trifurcates). Let $Q_0(A,B,\beta)$ denote the minimum of the
lhs. of \eqref{eq:P0} wrt. $s_2 \,{=}\, 2s_1,\, 0 \,{<}\, s_2 \,{<}\,
s^*$, $j \,{=}\, 1$ and $A \,{>}\,A^*$.\\ \indent Fig.\ref{fig:P6} and
fig.\ref{fig:P6fine} show some results of the analysis of pattern 6
with $B \,{=}\, kA$ for $0 \,{<}\, k \,{<}\, 1$ and $0.1 \,{<}\, \beta
\,{<}\, 0.9$. If $\beta$ is close to either $0$ or $1$ one can in
practice model the signal as pattern 1, pattern 2 or pattern 3,
depending on mutual relations between $k$ and $\beta$. For large
values of $\beta$ and $k \,{>}\, 1$ the maxima-lines become unstable
making numerical analysis difficult.\\ \indent The surfaces
$Q_i(A,B,\beta) {=} 0,\,i {=} 0,1$ represent the minimal
$A_0 {>} 0$ such that $Q_i(A,B,\beta) {>} 0$ for all $A {>}
0$. Fig.\ref{fig:P6} shows some level-curves of the surface
$Q_1(A,B,\beta) {=} 0$ for $\mathrm{P}(n,m)$ given by \eqref{eq:P}
with $\alpha {=} \pm 1/2$. Fig.\ref{fig:P6fine} shows the some
level-curves of the surface $Q_0(A,B,\beta) {=} 0$ for
$\mathrm{P}(n,m)$ given by \eqref{eq:P} with $\alpha \,{=} \,{-}\,
1/2$. Numerical studies suggest that $Q_0(A,B,\beta) \,{>}\, 0$ for
all $0.1 \,{<}\, \beta \,{<}\, 0.9$, $0 \,{<}\, B \,{<}\, A$ and $A^*
\,{<}\, A$ if $\alpha \,{=}\, 1/2$ or $\mathrm{P} \,{=}\, \Delta$,
except possibly for $A \,{\approx}\, A^*$.\\ \indent One may observe
in fig.\ref{fig:P6} and fig.\ref{fig:P6fine} that increasing the
influence of the distance-criterion increases the reliability of the
scale-space filtering procedure for pattern 6. This is caused by the
edge at $0 \,{<}\, \beta \,{<}\, 1$ which 'pushes' the other
maxima-lines away in different directions and hence strengthens the
distance-criterion. In addition, the edge at $\beta$ has negative
intensity $B$ which disrupts the decay-coefficient of the
wavelet transform, and hence the accuracy of the decay-criterion.
\subsection{Summary and discussion of 1-D
  time-scale filtering procedure.}\label{SS:summary} We have studied the
suggested time-scale filtering procedure from Sect.\ref{SS:P} with
respect to the six model-patterns in Sect.\ref{SS:dictionary}.\\
\indent Connecting mod-max across time-scale plane for the first two
patterns is trivial. For the other patterns the overall accuracy of
the decision-function $\mathrm{P}(\cdot,\cdot)$ in general increases
if we use a suitable weighting between the distance- and
decay-criterion. For pattern 3 and pattern 4 the accuracy of the
decision-function $\mathrm{P}(\cdot,\cdot)$ benefits from increasing
the influence of the decay-criterion. For the last two patterns we saw
that increasing the influence of the distance-criterion increases the
accuracy of the time-scale filtering procedure. It is clear that one
should take specific features of the concrete signals into account
when defining the mutual weighting between distance- and
decay-criterion. When analyzing the medical ultrasound images in
Sect.\ref{SS:expresconn} we use $\alpha {=} 0$.\\ \indent In
Sect.\ref{SS:expresreal} we will evaluate the performance of the
proposed time-scale filtering procedure with respect to medical
ultrasound signals.\\ \indent The results from the previous sections
are useful also for more complicated signals. For instance the signal
$f(t) \,{=}\, \rho(t) {+} A\rho(t{-}1) {+} B'\rho(t{-}7) {-}
A'\rho(t{-}9) {+} \rho(t{-}10)$ is locally similar to pattern 3 near
the point $0$ and pattern 4 near the point $10$. Due to fast decay of
the wavelet $\psi_s(t)$ the maxima-lines of the 'local' patterns in
$f(t)$ and their exact equivalent are practically identical for
$0\,{<}\, s \,{<}\, 3$. It makes little sense to analyze the signal at
scales $s\,{>}\,3$ for the purpose of edge-detection. Our analysis of
actual medical ultrasound signals indicates that such signals often
are composed of (distinct) patterns in this manner as illustrated in
fig.\ref{fig:atom}.\\ \indent Real medical ultrasound signals are of
course 'smooth'. The corresponding edges can be modeled by smoothed
versions of the 'ideal' patterns in Sect.\ref{SS:dictionary}. If $f$
is any of the ideal patterns say, then we can model the smoothed
patterns as $h = f\ast g_{\sigma}$ where $g_{\sigma}(t) \,{=}\,
\frac{1}{\sqrt{2\pi}\sigma} e^{-t^2/2\sigma^2}, \sigma \,{>}\,0$, is
the Gaussian smoothing kernel.\\ \indent With some natural
modifications the suggested time-scale filtering procedure applies to
the smoothed patterns as well. We need to use denser (i.e. non-dyadic)
scales; more precisely we need to choose $s_2, s_1$ such that
$\sqrt{s_2^2 + \sigma^2} = 2\sqrt{s_1^2 + \sigma^2}$. If we in
addition change $\dots -1/2$ to $\dots -1/2 - \frac{\ln
  \frac{s_2}{2s_1}}{\ln 2}$ in the expression for the decay-criterion
$\mathrm{D}(n,m)$, the results from Sect.\ref{SS:eval} apply directly
(i.e. with
the same $A,B,\beta$).\\
\subsection{Decision-function and space-scale filtering procedure for
  2-D signals.}\label{SS:P2}
We turn the attention towards a space-scale filtering procedure for
connecting mod-max across the \textit{space-scale plane} of 2-D
signals. In images mod-max are chained together to form
\textit{boundary-curves} corresponding to boundaries of objects, and
the natural space-scale plane object is the
\textit{maxima-surface}. The behavior of a maxima-surface is more
subtle than for a maxima-line. For instance one can not assume that a
maxima-surface does not branch. The target with the space-scale
filtering procedure is to decide which mod-max propagate toward each
of the boundary-curves at the finest scale.\\ \indent The suggested
decision-function for 2-D signals is - with some natural modifications
- similar to the 1-D decision-function defined in Sect.\ref{SS:P}. Let
$s_2$ and $s_1$ be two scales $s_2 {>} s_1$, and let $(n,s_2) {=}
\big((n_1,n_2),s_2\big)$ and $(m,s_1) {=} \big((m_1,m_2),s_1\big)$ be
two mod-max. We define the decision-function as;
\begin{equation*}
  \mathrm{P}(n,m)
  \,{=}\, \Delta(n,m) \mathrm{D}(n,m) \mathrm{Angle}(n,m)
\end{equation*}
where;
\begin{small}
  \begin{align*}
    &\Delta(n,m) = \mathrm{exp}\big(-|n-m|s_1^{-\alpha}\big)
    \\ &\mathrm{D}(n,m) = \mathrm{exp}\Bigg(-\Big|\ln
    \frac{|Wf(n,s_2)|}{|Wf(m,s_1)|}\ln^{-1} \frac{s_2}{s_1} -
    1\Big|s_1^{\alpha}\Bigg)\\ &\mathrm{Angle}(n,m) =
    \mathrm{exp}(-|Af(n,s_2) - Af(m,s_1)|). 
  \end{align*}
\end{small}
$Af(u,s)$ is the direction of the wavelet transform at the mod-max
$(u,s)$, and is given by
\begin{displaymath}
  Af(u,s) = \Bigg\{ \begin{array}{ll} \arctan
    \frac{Wf^y(u,s)}{Wf^x(u,s)} & Wf^x(u,s) \geq 0\\ \pi + \arctan
    \frac{Wf^y(u,s)}{Wf^x(u,s)} & Wf^x(u,s) < 0.\end{array} 
\end{displaymath}
The essence of the space-scale filtering procedure is - as in 1-D - to
connect mod-max which maximize $\mathrm{P}(n,m)$.\\ \indent In
particular, if one consider 2-D patterns obtained by a tensor product
of the 1-D patterns, then the analysis in Sect.\ref{SS:eval} extends
to the 2-D procedure.\\ \indent Given two scales $s_2 > s_1$ the
procedure requires that for each $n\in \mathcal{M}f(s_2)$ one has to
compute $P(n,m)$ for every $m\in\mathcal{M}f(s_1)$. One can reduce the
computational efforts by computing $\mathrm{P}(n,m)$ only for
$m\in\mathcal{M}f(s_1)$ within a given
distance ($4 s_1$ should suffice if we use dyadic scales) of $n$.\\
\indent In the next section we will see that for our purpose it is
sufficient to compute $P(n,m)$ for only a few mod-max at scale $s_2$
as well. By using these simplification one will effectively reduce the
computational complexity of the space-scale filtering procedure for
images.

\section{Edge-detection.}\label{S:IV}
\indent The uncertainty-principle states that in a noisy signal there
exists no single scale such that one can detect and simultaneously
localize only the significant edges, \cite{jC86}. However, by using
the maxima-lines one can turn the uncertainty-principle into our
advantage. At coarse scale one can detect the significant edges in the
signal, and at fine scales one can localize them. Since maxima-lines
connect coarse- and fine-scale information one can effectively improve
edge-detection in noisy signals.\\ \indent In this section we discuss
a multi-scale wavelet-based edge-detection scheme. The method is based
on the suggested time-scale filtering procedure and a perceptual
criteria similar to that in \cite{jL92}. This approach has shown
promising results for detecting and localizing (low-contrast) edges in
medical images.
\subsection{1-D edge-detection.}
Assume that the wavelet transform is computed at scales $s_j$ for $j =
1,...,J$, and let $\mathcal{M}f(s_1)$ denote the mod-max at the finest
scale $s_1$. For a mod-max $a\in\mathcal{M}f(s_1) $ let $\ell_a$
denote the maxima-line which contains $a$. To each $a$ assign the
number;
\begin{equation*}
  R(a) = \sum_{(u,s)\in\ell_a} |Wf(u,s)|.
\end{equation*}
If need be one can introduce a weight-function $\mu(s)$ in the
expression for $R(a)$, for instance to favor fine/coarse scale
information. The final step of the edge-detector is to decide a
threshold $T > 0$ and find the mod-max $a\in\mathcal{M}f(s_1)$ such
that $R(a) > T$. These mod-max are used to represent the significant
edges in the signal.
\subsection{2-D edge-detection}\label{SS:detector2}
Assume that the wavelet transform is computed at scales $s_j$ for $j =
1,...,J$ and let $c_j$ denote the boundary-curves at scale $s_1$.\\
\indent To each maxima-curve $c_j$ we assign the following quantity;
\begin{equation}\label{eq:detector2}
  S(c_j) = (\mathrm{length\;of}\;c_j) \cdot
  \mathrm{av}(\{R(a):a \in c_j\}),
\end{equation}
where $\mathrm{av}(\cdot)$ denotes the average-function. The final
step of the edge-detection is to identify the curves $c_j$ such that
$S(c_j) > T$, where $T \,{>}\, 0$ is a threshold. These
boundary-curves is used to represent the boundaries of significant
edges in the image.\\ \indent The quantity $\mathrm{av}(\{R(a):a \in
c_j\})$ reduces the influence of a few false connections made by the
space-scale filtering procedure along a boundary-curve $c_j$. Such
false connections can occur for some 'critical' patterns (e.g. $A
\approx 1$ in pattern 3).\\ \indent As mentioned in Sect.\ref{SS:P2}
on can save computational efforts by reducing the number of $m$'s for
which one has to compute $P(n,m)$. Further reduction can be achieved
by computing the average in \eqref{eq:detector2} from only a 'few'
randomly selected maxima-lines from each maxima-curve
$c_j$. Experiments indicate that this simplification has little
influence on the values of $S(c_j)$.

\section{Experiments and results.}\label{S:V}
We test the edge-detector on both phantom and actual
medical ultrasound signals. By studying the algorithm on
phantom-signals we compare the Figure of Merit \cite{wP91} with the
Canny edge-detector. The visual performance of the edge-detector is
analyzed with respect to medical ultrasound signals. This
analysis focuses on some well-known problem features of
medical ultrasound images; such as low-contrast edges and speckle-noise.\\
\indent Before studying the edge-detection algorithm we analyze the
reliability of the suggested time-scale filtering procedure on actual
signals. The main question is whether the results from
Sect.\ref{S:III} also apply to actual medical ultrasound signals.
\subsection{Accuracy of time-scale filtering procedure on medical
  ultrasound signals.}\label{SS:expresreal} \indent To quantitatively
evaluate the performance of the suggested time-scale filtering
procedure from Sect.\ref{SS:P} we compare the output of the procedure
with the exact maxima-lines of the wavelet transform. Assuming that
the procedure connects the mod-max $(n,s_2)$ to $(m,s_1)$ and $n {=}
\ell_a(s_2)$ say, then we will study if $m {=} \ell_a(s_1)$. The
test-signals used to evaluate the procedure are 1-D cross-sections of
medical ultrasound images.\\ \indent In the analysis we study the
probability of a false connection, and between which scales an error
is most likely to occur. For the purpose of edge-detection the
severity of a false connection increases at coarse scales. We also
study the average displacement (in pixels) of a mod-max caused by a
false connection. If this number is small it indicates that an error
has little influence on the visual output of the edge-detector.\\
\indent In the experiments we use scales $s = 2^j,\;1\leq j \leq
5$. The control-parameter is chosen as $\alpha
=\{-1/2,0,1/2\}$\footnote{Note that in the analysis of the patterns in
Sect.\ref{S:III} we had $s_1 < 1$ in the expression for
$\mathrm{P}(n,m)$ in \eqref{eq:P}. In particular $s_1^{1/2} < s_1^{0}
< s_1^{-1/2}$. Since we are using scales $s = 2^j,\;1\leq j \leq 5$ to
analyze actual medical ultrasound signals we will necessarily have
$s_1^{1/2} > s_1^{0} > s_1^{-1/2}$. This causes that a large $\alpha$
favors decay-criteria and a small favors the
distance-criteria. I.e. $\alpha {=}-1/2$ in the preceding section
'corresponds' to $\alpha = 1/2$ in this section, and visa verse for
$\alpha {=} 1/2$.}. To obtain the actual maxima-lines we use the
edge-focusing algorithm \cite{fB87}.\\ \indent The results of the
quantitative analysis of the time-scale filtering procedure is
displayed in Tab.\ref{tab:error1}{-}\ref{tab:error2}. Each column
represents the average performance of the procedure applied to every
horizontal cross-section in the corresponding image. The images are
displayed in fig.\ref{fig:visual}(a,b,c) and fig.\ref{fig:phantom}b.\\
\begin{table}[!h]
  \begin{center}
      \begin{scriptsize}
    \begin{tabular}{|l|c|c|c|c|c|c|}
      \hline \textbf{Signal:} & \parbox[b]{3em}{Tumor}
      & \parbox[b]{4em}{Blood - vessel} & \parbox[b]{3em}{Liver}
      & \parbox[b]{3em}{phantom}\\ \hline
      False connections
      $(\%)$: & 4.8 & 4.7 & 6.2 & 4.2\\ \hline
      \parbox[b]{11em}{Spatial error in pixels:} & 12.4 (2.5) & 12.9 (2.5)
      & 12.2 (2.8) & 20.3 (7.4)\\ \hline 
      \parbox[b]{13em}{False connections between\\ scale $32$ and $16$:} & 2.5 &
      4.4 & 8.5 & 4.2\\ \hline 
      \parbox[b]{13em}{False connections between\\ scale $16$ and $8$:}  & 5.4 &
      5.2 & 9.0 & 8.0\\ \hline  
      \parbox[b]{13em}{False connections between\\ scale $8$ and $4$:}   & 6.2 &
      5.7 & 8.2 & 5.0\\ \hline  
      \parbox[b]{13em}{False connections between\\ scale $4$ and $2$:}   & 4.8 &
      4.5 & 6.0 & 1.7\\ \hline  
    \end{tabular}
  \end{scriptsize}
\end{center}\caption{Analysis of the time-scale filtering procedure in
  Sect.\ref{SS:P} on medical ultrasound signals. The control-parameter
  between distance- and decay-criteria is $\alpha {=} 1/2$. The number
  is parenthesis on the 2. row is the spatial error in percent
  relative to size of the signal.}\label{tab:error1}
\end{table}
\begin{table}[!h]
  \begin{center}
      \begin{scriptsize}
    \begin{tabular}{|l|c|c|c|c|c|c|}
      \hline \textbf{Signal:} & \parbox[b]{3em}{Tumor}
      & \parbox[b]{4em}{Blood - vessel} & \parbox[b]{3em}{Liver}
      & \parbox[b]{3em}{phantom}\\ \hline
      False connections
      $(\%)$: & 4.0 & 4.2 & 4.8 & 3.8\\ \hline
      \parbox[b]{13em}{Spatial error in pixels:} & 11.5 (2.2) & 13.2 (2.6)
      & 12.6 (2.9) & 19.2 (7.0)\\ \hline 
      \parbox[b]{13em}{False connections between\\ scale $32$ and $16$:} & 1.9 &
      4.3 & 8.6 & 3.3\\ \hline 
      \parbox[b]{13em}{False connections between\\ scale $16$ and $8$:}  & 4.3 &
      5.3 & 7.9 & 8.0\\ \hline  
      \parbox[b]{13em}{False connections between\\ scale $8$ and $4$:}   & 5.2 &
      4.8 & 5.4 & 4.1\\ \hline  
      \parbox[b]{13em}{False connections between\\ scale $4$ and $2$:}   & 4.1 &
      3.9 & 4.7 & 1.6\\ \hline  
    \end{tabular}
  \end{scriptsize}
  \end{center}\caption{Analysis of the time-scale filtering procedure in
    Sect.\ref{SS:P} with control-parameter $\alpha {=}
    0$.}\label{tab:error0}
\end{table}
\begin{table}[!h]
  \begin{center}
      \begin{scriptsize}
    \begin{tabular}{|l|c|c|c|c|c|c|}
      \hline \textbf{Signal:} & \parbox[b]{3em}{Tumor}
      & \parbox[b]{4em}{Blood - vessel} & \parbox[b]{3em}{Liver}
      & \parbox[b]{3em}{phantom}\\ \hline False connections 
      $(\%)$: & 3.9 & 4.0 & 4.6 & 3.7\\ \hline
      \parbox[b]{13em}{Spatial error in pixels:} & 11.7 (2.3) & 13.5
      (2.6) & 13.3 (3.1) & 19.3 (7.0)\\ \hline  
      \parbox[b]{13em}{False connections between\\ scale $32$ and $16$:} & 1.9 &
      4.3 & 8.6 & 3.3\\ \hline 
      \parbox[b]{13em}{False connections between\\ scale $16$ and $8$:}  & 4.4 &
      5.3 & 7.8 & 7.8\\ \hline  
      \parbox[b]{13em}{False connections between\\ scale $8$ and $4$:}   & 5.0 &
      4.5 & 5.2 & 4.0\\ \hline  
      \parbox[b]{13em}{False connections between\\ scale $4$ and $2$:}   & 4.0 &
      3.7 & 4.5 & 1.6\\ \hline  
    \end{tabular}
      \end{scriptsize}
  \end{center}\caption{Analysis of the time-scale filtering procedure in
    Sect.\ref{SS:P} with control-parameter $\alpha
    {=} -1/2$.}\label{tab:error2}
\end{table}
The analysis of pattern 6 in Sect.\ref{SS:eval} showed that the
accuracy of the suggested time-scale filtering procedure increases as
$\alpha$ increases from $-1/2$ to $1/2$. Speckle noise can often
locally be modeled as pattern 6 (with $A {\approx} B {\approx} 1$ and
$\beta {\approx} \frac{1}{2}$). Hence we expect to observe a similar
increase in accuracy for actual medical ultrasound signals. In
Tab.\ref{tab:error1}-\ref{tab:error2} one can indeed observe such
increase of the accuracy of the procedure on medical ultrasound
signals (recall that $\alpha {=} 1/2$ in this analysis corresponds to
$\alpha {=}-1/2$ in the analysis in Sect.\ref{S:III}, etc). A closer
inspection indicates that false connections \textit{in between}
maxima-lines corresponding to speckle-noise is the typical problem for
the time-scale filtering procedure on ultrasound signal. Since
speckle-noise contains information which is unwanted from an
edge-detection viewpoint, such false connections will only be a minor
problem.
\subsection{Results of edge-detection on phantom ultrasound images.}
\begin{figure}[!h]
  \centering \subfigure[]{\includegraphics[height=4cm,width=4cm]
    {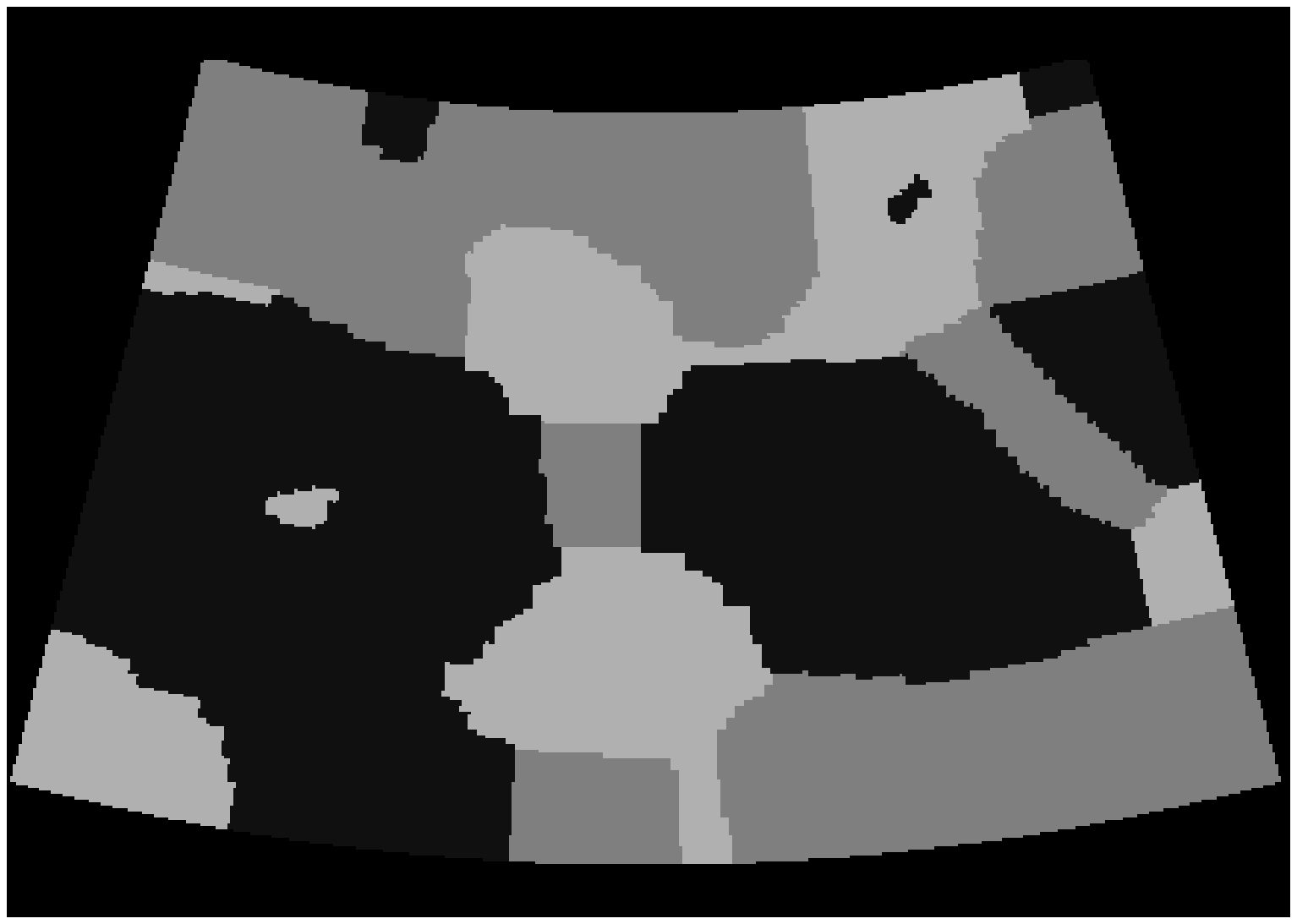}}
  \subfigure[]{\includegraphics[height=4cm,width=4cm]
    {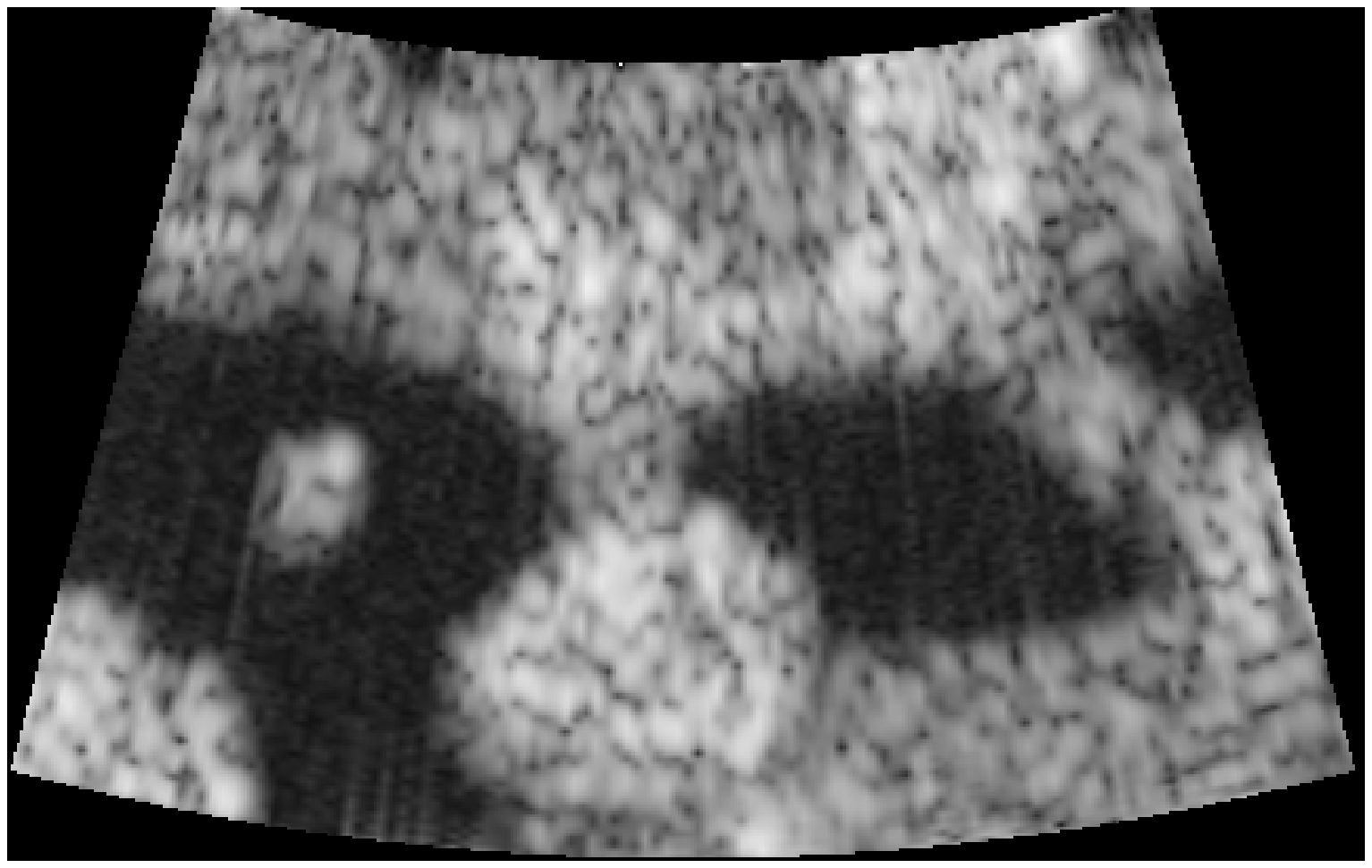}}
  \subfigure[]{\includegraphics[height=4cm,width=4cm]
    {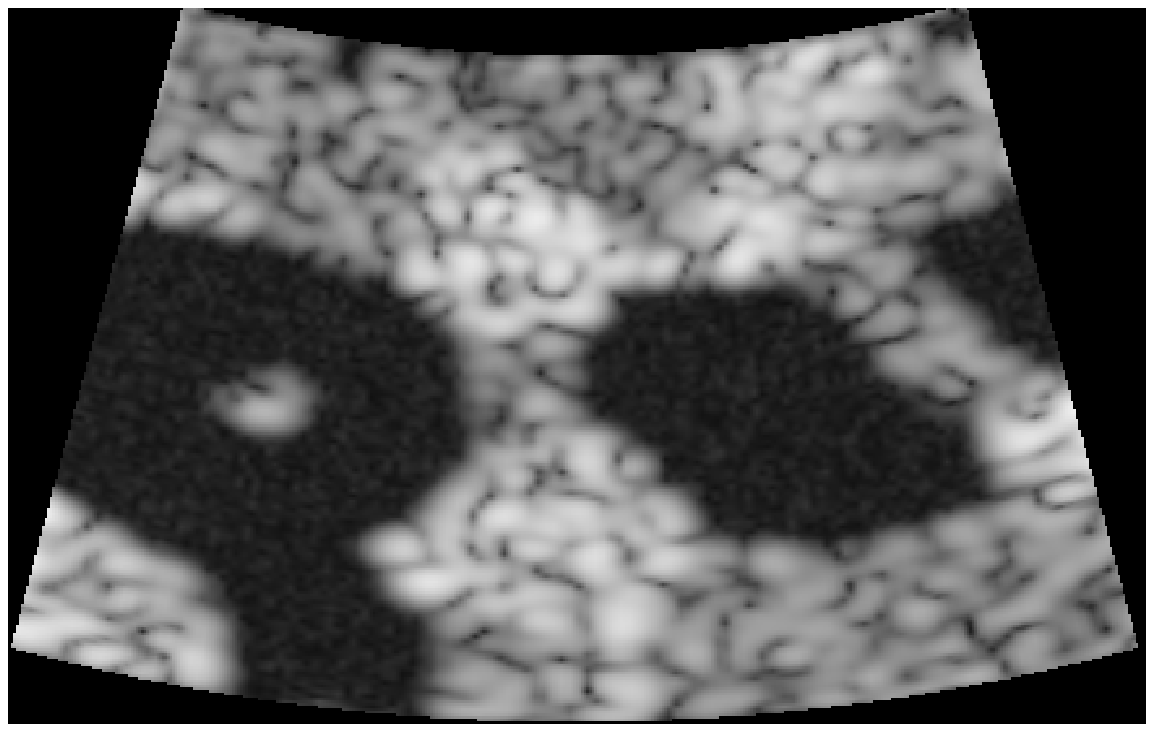}}
  \caption{Phantom medical ultrasound images. (a) Noise-free image,
    (b,c) images with speckle-noise.}\label{fig:phantom}
\end{figure}

\indent To study the performance of the edge-detection algorithm
suggested in Sect.\ref{SS:detector2} we compare it with the well-known
Canny edge-detector. We believe that comparing the Canny edge-detector
with ours demonstrates some differences between multi-scale and
single-scale wavelet-based edge-detection.\\ \indent For the Canny
edge-detector with hysteria we have used thresholds $T_{\mathrm{low}}
= 0.1 \cdot \max|Wf((\cdot,\cdot),s)|$ and $T_{\mathrm{high}} =
0.3\cdot \max|Wf((\cdot,\cdot),s)|$. The scale is chosen equal to the
finest scale used by our algorithm.\\ \indent To quantitatively
analyze the edge-detectors we study the phantom medical ultrasound
signals shown in fig.\ref{fig:phantom}.
The image in fig.\ref{fig:phantom}(a) shows the noise-free version,
and fig.\ref{fig:phantom}(b,c) shows the images with various
speckle-noise. The phantom images are generated by using a region
image as a variance field for the diffuse scattering. The images are
made by convolving the ideal image with a parametric point-spread
function and adding white Gaussian noise, and are discussed in details
in \cite{tL01,oH01}\footnote{The images are kindly provided by SINTEF
  MedTech. The author thanks T. Lang{\o} for his assistance.}.\\
\indent We use the well-known Figure of Merit (FOM)\cite{wP91} to
quantitatively evaluate the edge-detection algorithms. The FOM is the
number defined as;
\begin{equation*}
  \mathrm{FOM} = \frac{1}{\max \{n_p, n_d\}} \sum_1^{n_d}\frac{1}{1 +
    \gamma d_i^2},
\end{equation*}
where $n_p$ and $n_d$ are the number of resp. true and detected edges,
and $d_i$ is the minimal distance from a detected to a true
edge. $\gamma$ is a constant used to penalize displaced mod-max. To be
consistent with others we use $\gamma {=} 0.11$ \cite{yY06,gE09}. The
$\mathrm{FOM}$ should ideally be close to $1$.\\ \indent The values of
the Figure of Merit of the two edge-detectors are displayed in
tab.\ref{tab:FOM}\,{-}\,\ref{tab:FOM2}.
\begin{table}
  \begin{center} \begin{tabular}{|l|c|c|c|c|c|c|c|} \hline \textbf{Signal:}
    & proposed edge-detector & Canny \\ \hline \textbf{phantom 1:} & 0.56 &
    0.53\\ \hline \textbf{phantom 2:} & 0.54 &
    0.47 \\ \hline \end{tabular} \end{center}\caption{The Figure of
    Merit of the edge-detection algorithm in Sect.\ref{SS:detector2}
    and the Canny edge-detector at scale $4$.}\label{tab:FOM}
\end{table}
\begin{table}
  \begin{center}
    \begin{tabular}{|l|c|c|c|c|c|c|c|}
      \hline \textbf{Signal:}    & proposed edge-detector    & Canny \\
      \hline \textbf{phantom 1:} &  0.54 & 0.53\\ 
      \hline \textbf{phantom 2:} &  0.53 & 0.51
      \\ \hline      
    \end{tabular}
  \end{center}\caption{The Figure of Merit of the edge-detection
    algorithm in Sect.\ref{SS:detector2} and the Canny edge-detector
    at scale $6$.}\label{tab:FOM2}
\end{table}
\subsection{Result of edge-detection on real ultrasound
  images.}\label{SS:expresconn}
\indent In this section we discuss the results of the suggested
edge-detection algorithm on real signals.  For this purpose we study
the edges detected by the algorithm in several actual medical
ultrasound images. The goal is to study if the algorithm detects the
significant edges in the images without including irrelevant
information such as noise. In addition we
discuss how one can obtain a threshold as well as the computational
complexity of the
algorithm.\\
\begin{figure}[!h]
  \centering \subfigure[]{\includegraphics[height=4cm,width=4cm]
    {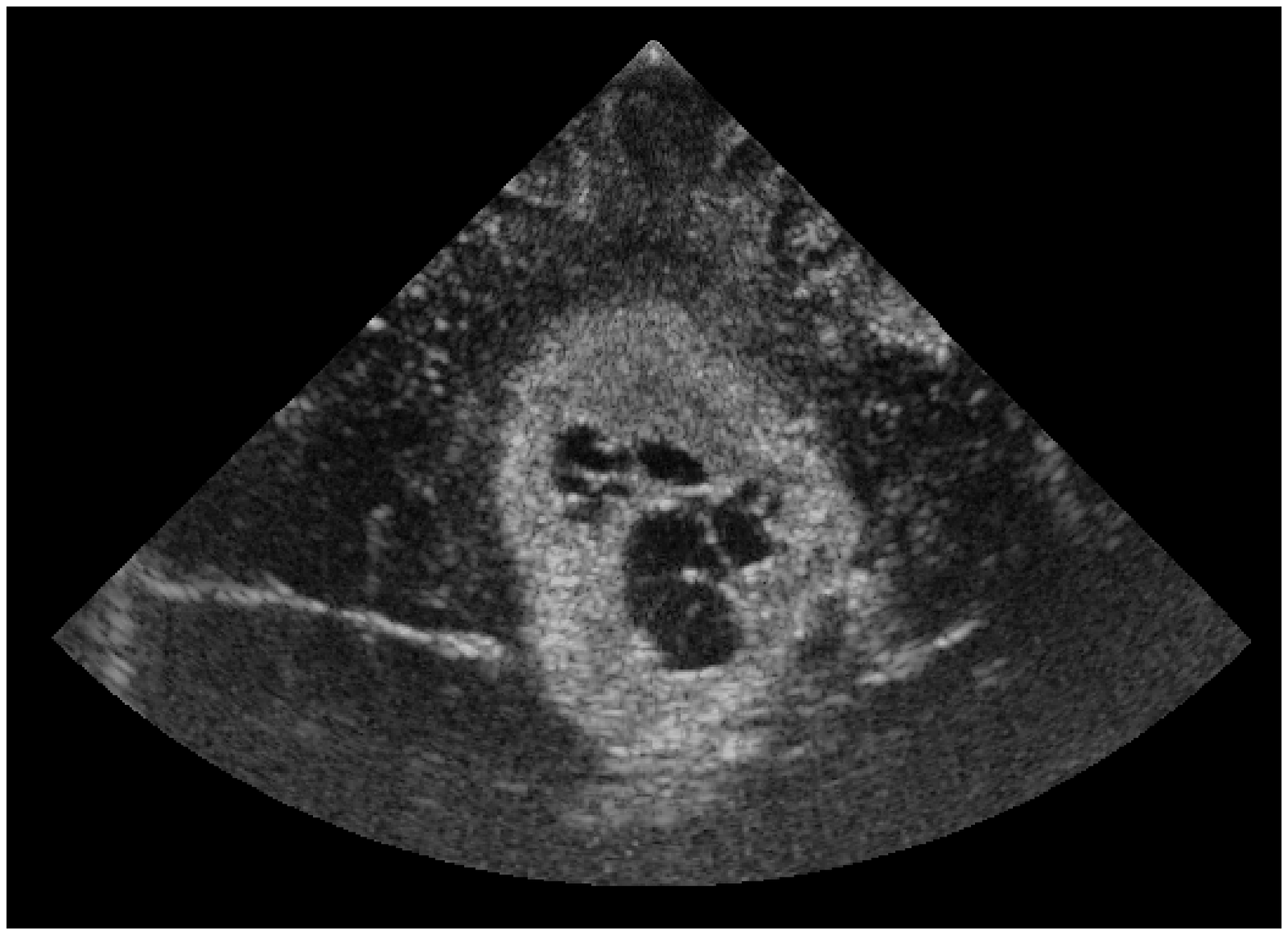}} \subfigure[]{\includegraphics[height=4cm,width=4cm]
    {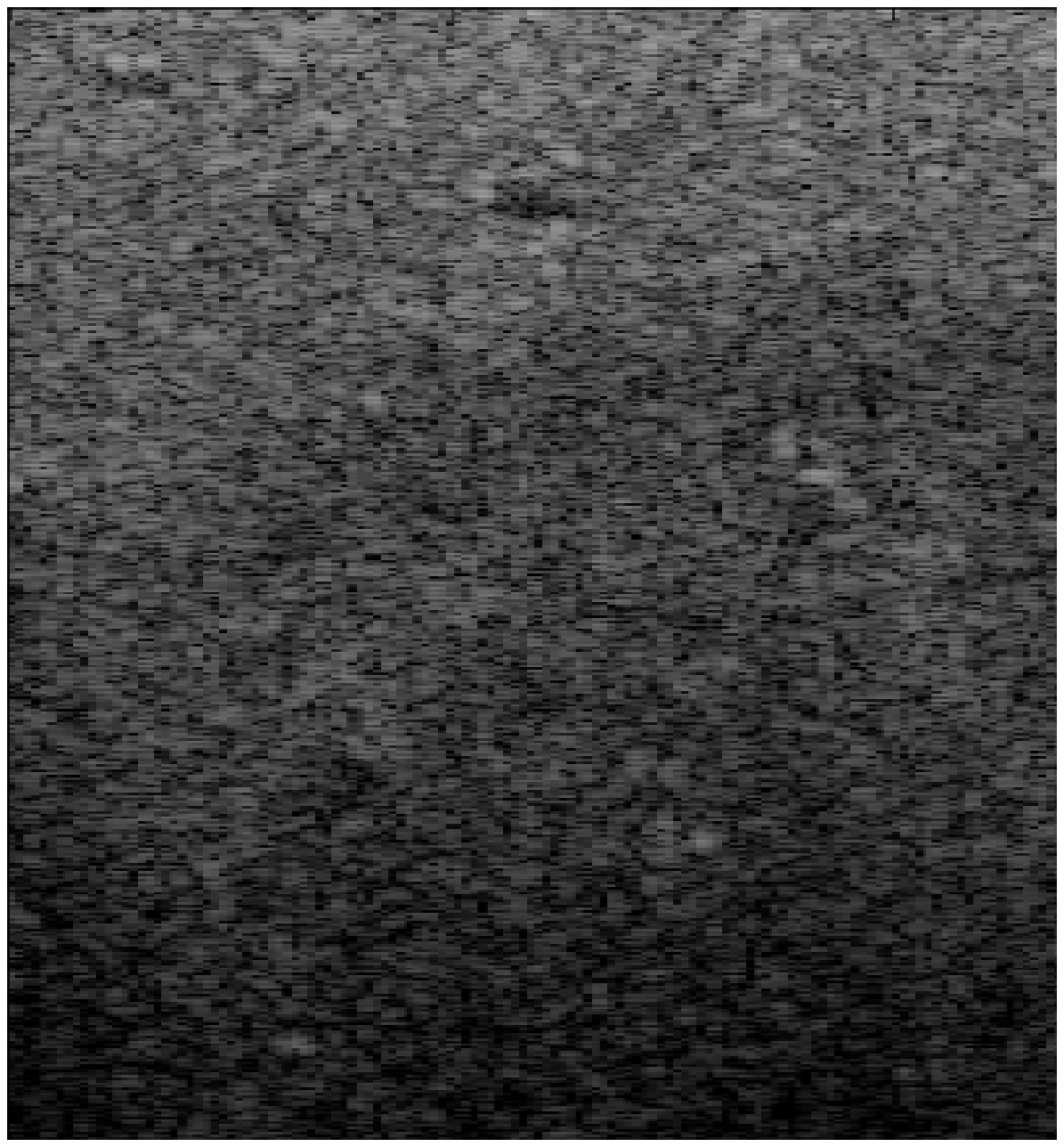}} \subfigure[]{\includegraphics[height=4cm,width=4cm]
    {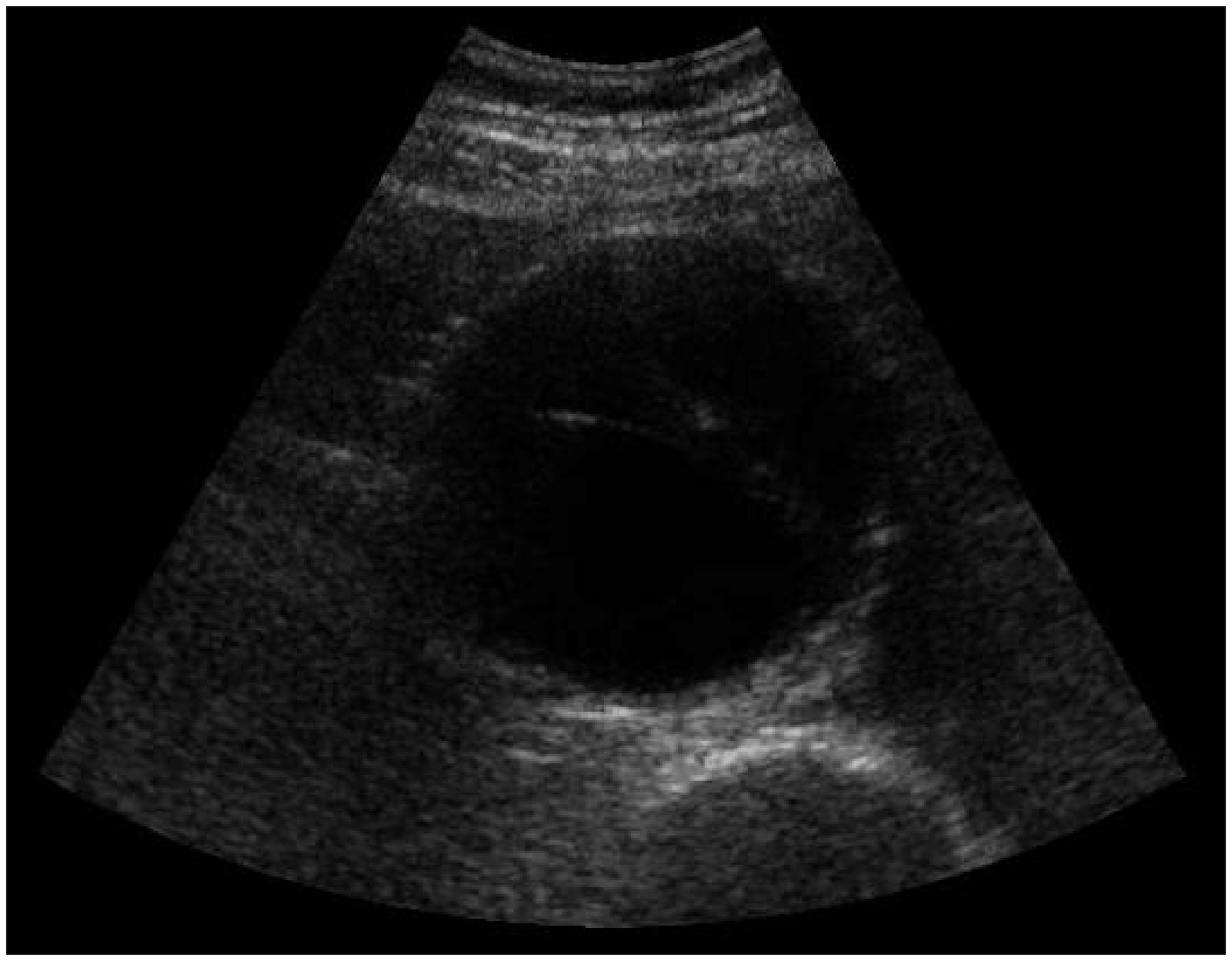}}
\caption{(a) : Tumor-image, (b) : Liver-image, (c) : Blood-vessel
  image.}\label{fig:visual}
\end{figure}
\indent Fig.\ref{fig:visual} shows the ultrasound images which are
discussed in this section. Each image contains some typical features
of medical ultrasound images which cause difficulties for
edge-detection.\\ \indent \textit{Fig.\ref{fig:visual}(a),
  Brain-tumor image:} The interesting edges are the outer boundary of
the tumor and the cortex (white 'horizontal' object to the south-west
of the tumor). The voids within the tumor are mainly interesting in
order to study how the edge-detector represents fine details.\\
\indent \textit{Fig.\ref{fig:visual}(b), Liver image:} The interesting
edge is the boundary of the liver. The boundary is the horizontal line
in the upper third in the image. This image interesting to study
because of the low CNR of the boundary.\\ \indent
\textit{Fig.\ref{fig:visual}(c), Blood-vessel image:} In this image we
want to detect the boundary of the blood-vessel. Some parts of the
boundary of the blood-vessel have low contrast making them difficult
to detect. This problem is typical for ultrasound images of
blood-vessels.\\
\begin{figure}[!h]
  \centering \subfigure[]{\includegraphics[height=4cm,width=4cm]
    {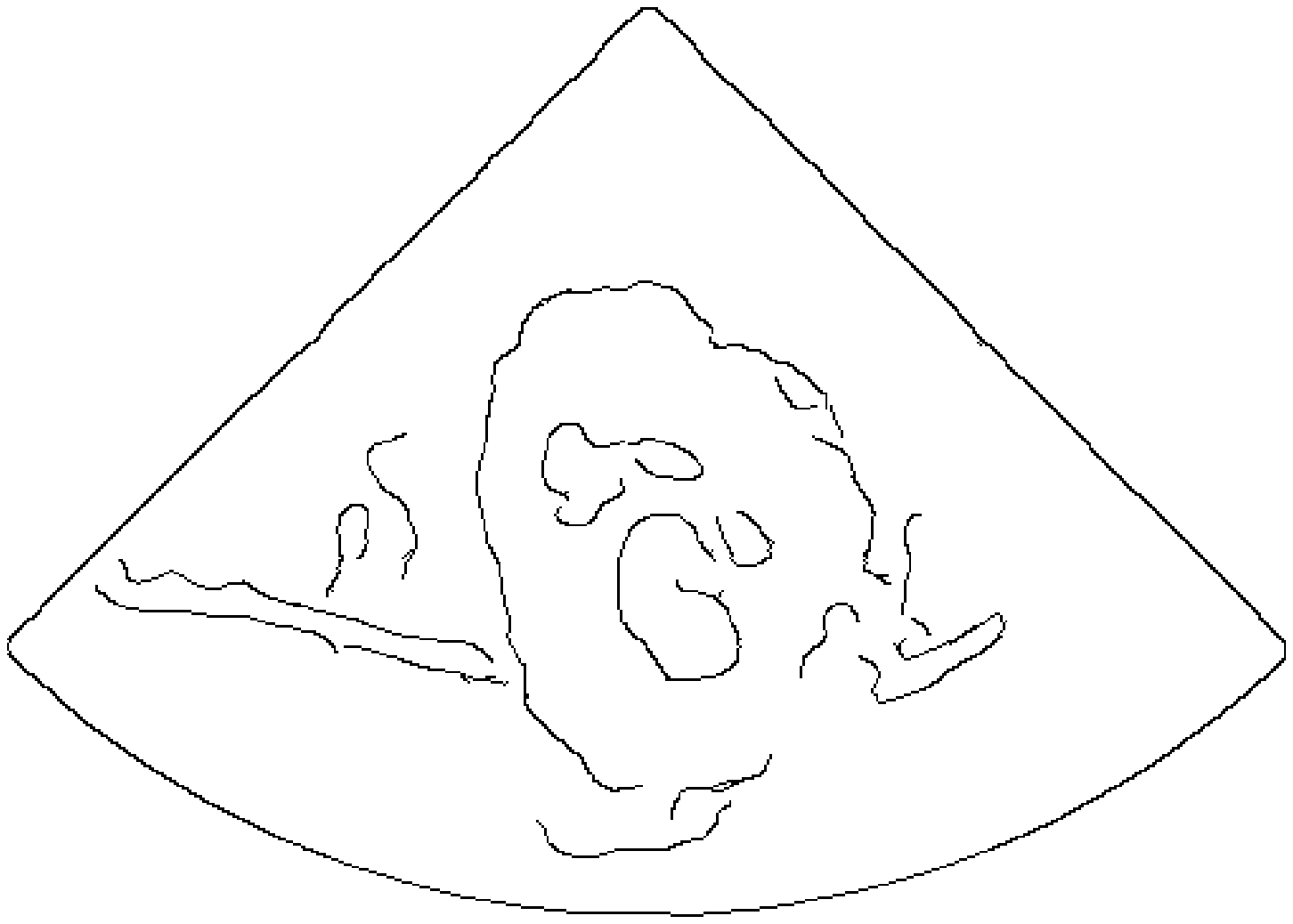}}
  \subfigure[]{\includegraphics[height=4cm,width=4cm]
    {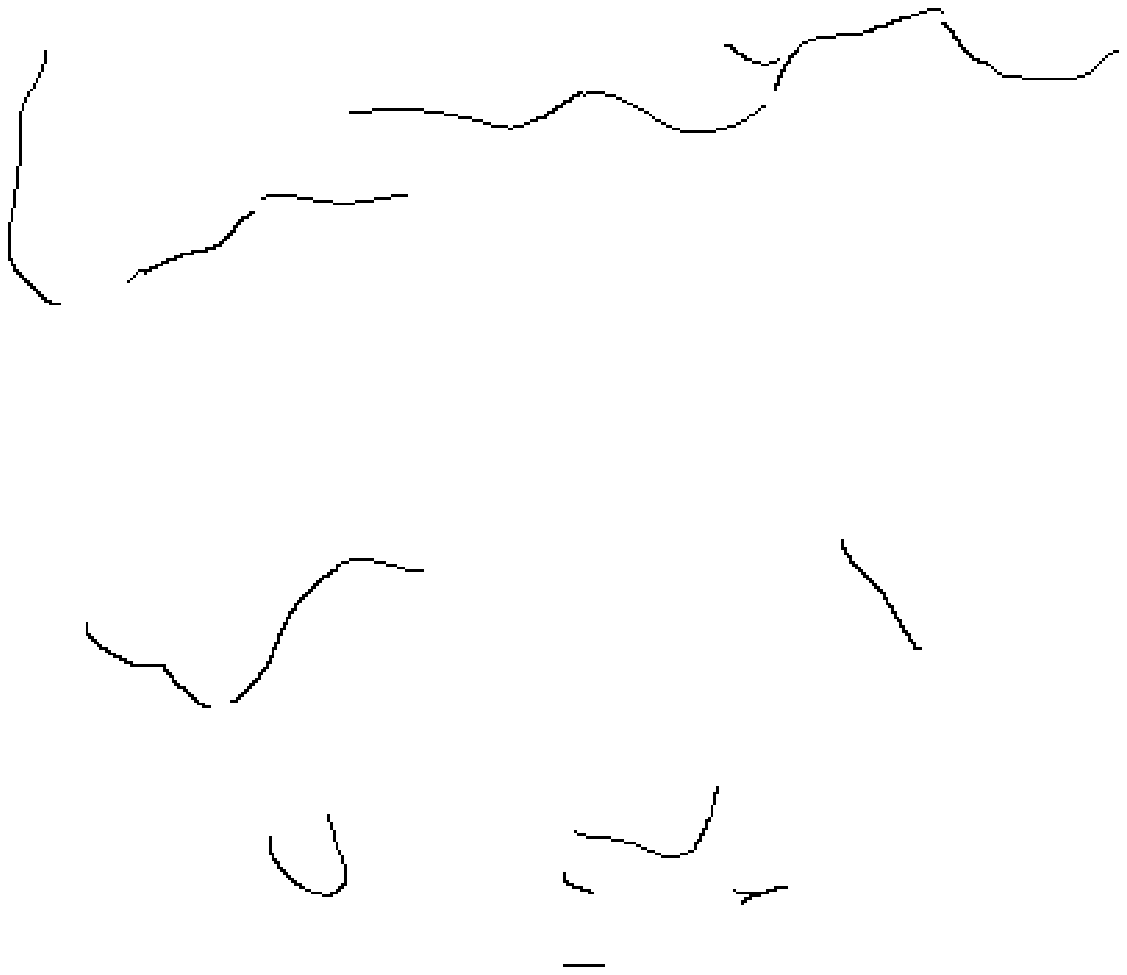}}
  \subfigure[]{\includegraphics[height=4cm,width=4cm]
    {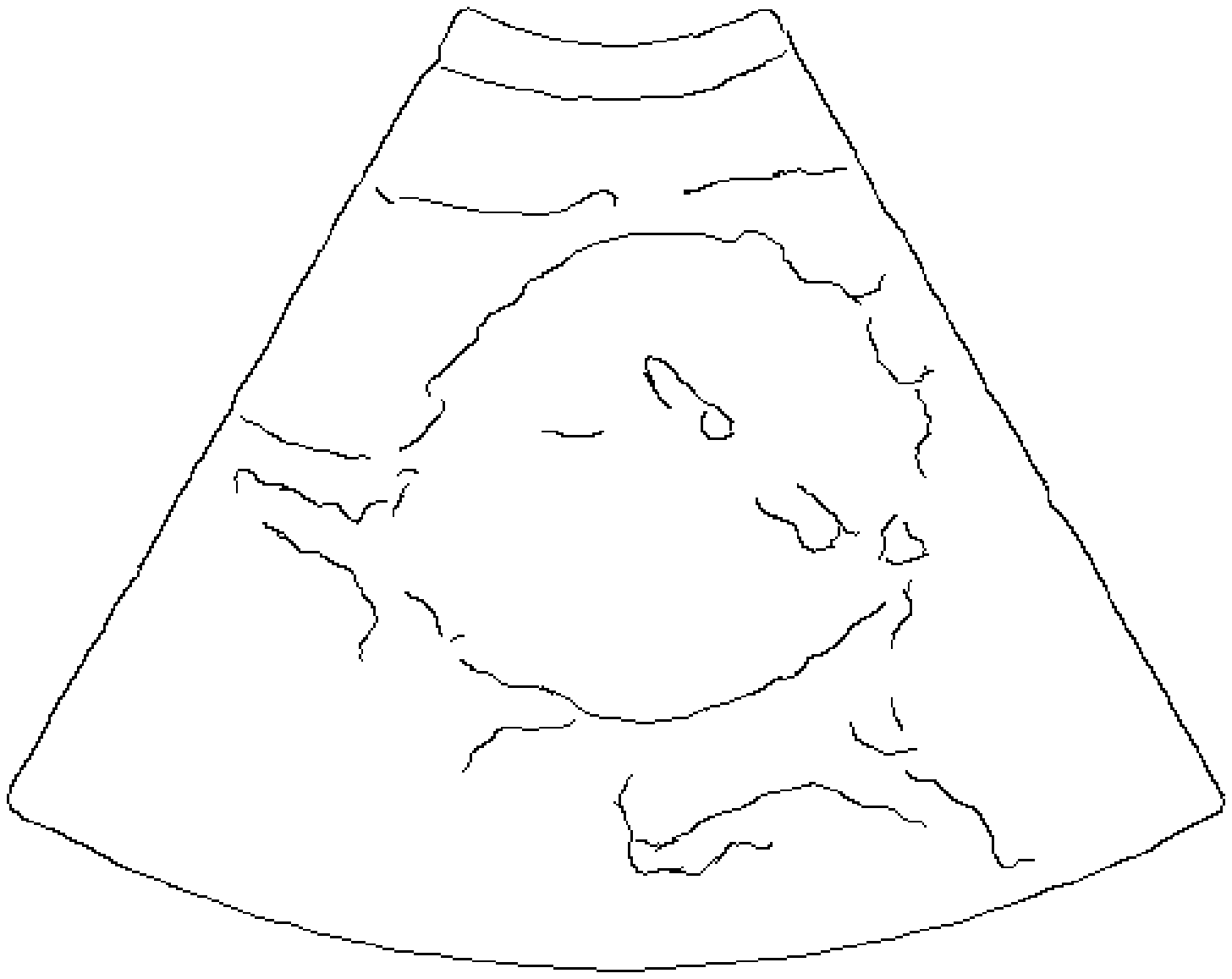}}\\
  \subfigure[]{\includegraphics[height=4cm,width=4cm]
    {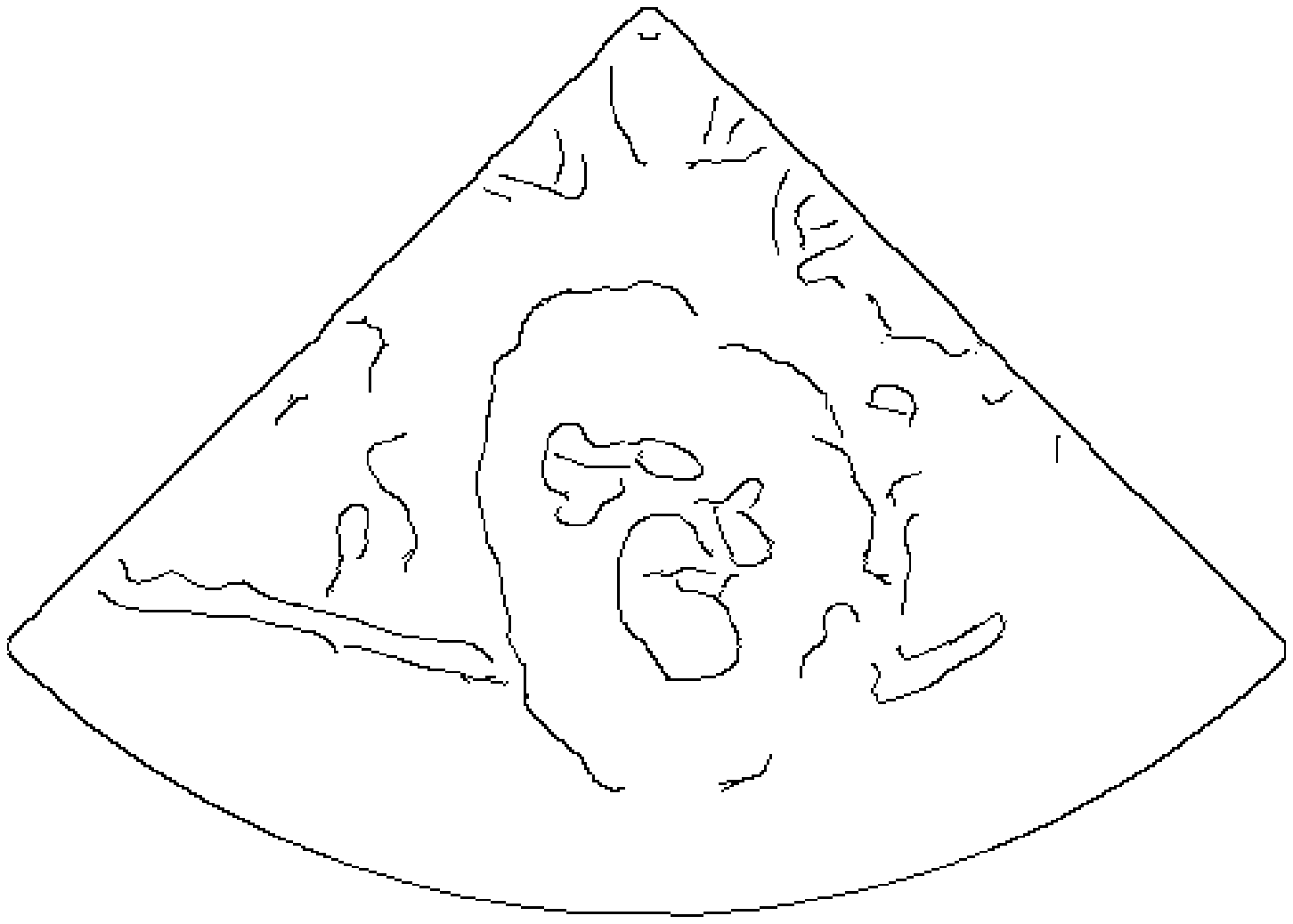}}
  \subfigure[]{\includegraphics[height=4cm,width=4cm]
    {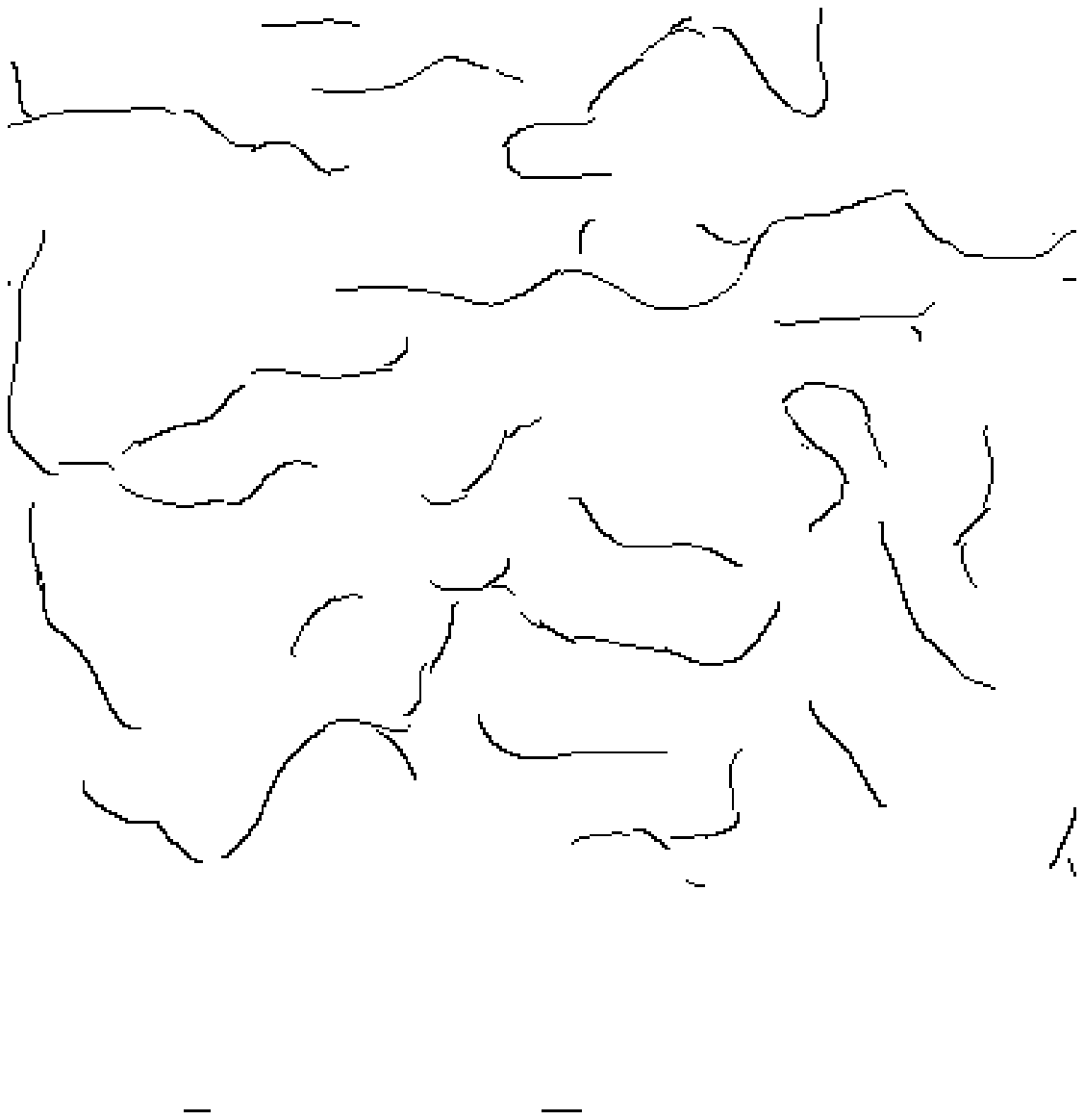}}
  \subfigure[]{\includegraphics[height=4cm,width=4cm]
    {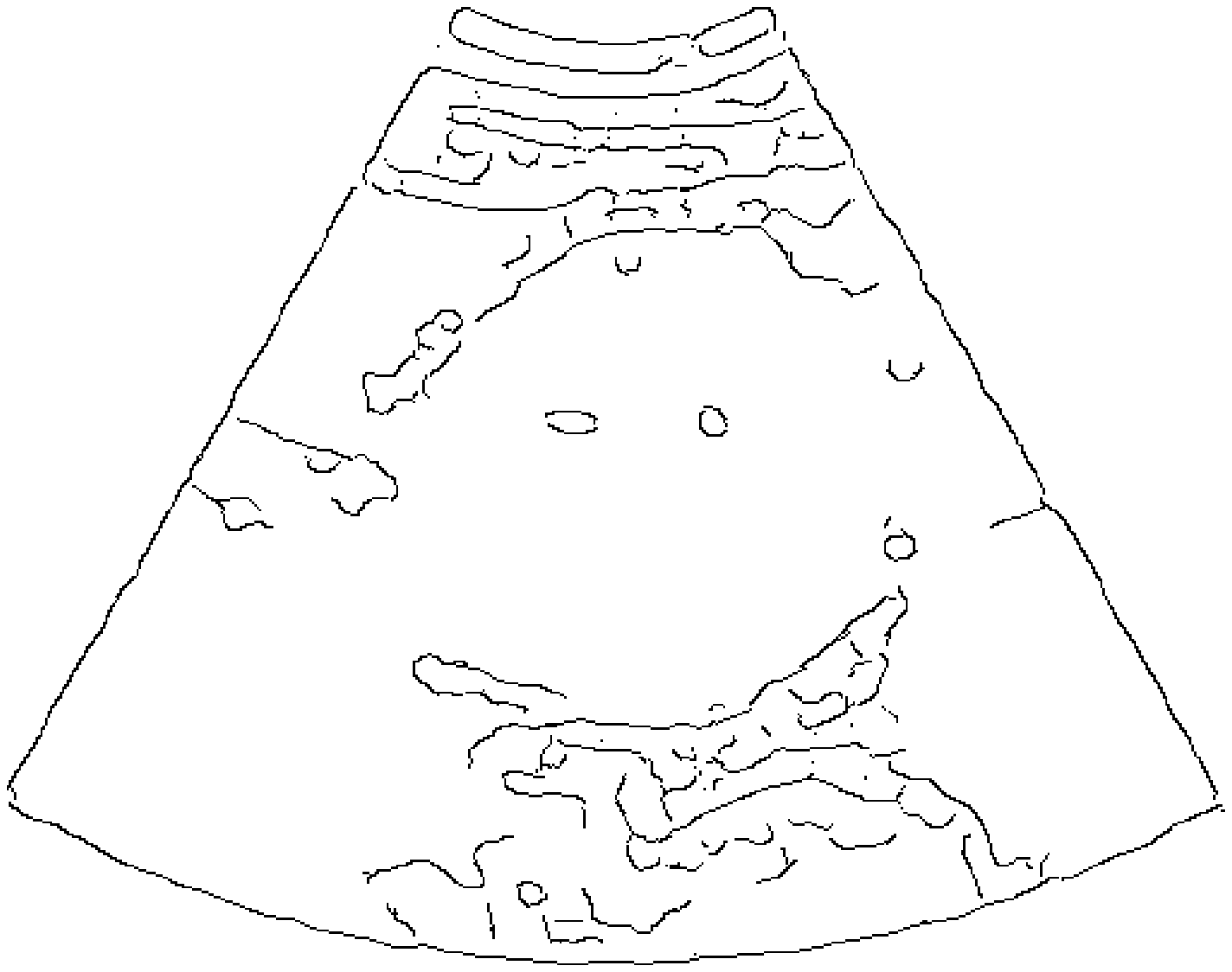}}
\caption{Results of the edge-detection algorithm of the images in
  fig.\ref{fig:visual}. In figure (a) - (c) we display the results of
  the edge-detection algorithm studied in Sect.\ref{S:IV} . In figure
  (d)-(f) we display the results obtained with a Canny
  edge-detector. (a,d) : Tumor image, (b,e) : Liver image, (c,f) :
  Blood-vessel image.}\label{fig:result}
\end{figure}
\indent \textit{Edge-detection and noise-removal:}
Fig.\ref{fig:result}(a,b,c) shows some results of our edge-detection
algorithm applied to the images in fig.\ref{fig:visual}. Most of the
significant edges which we would like to detect are present in the
representations. One may also observe, in particular in
fig.\ref{fig:result}(a,c), that the representations contain few edges
which can be ascribed to noise. The representations in
fig.\ref{fig:result} show promising results for detecting low contrast
edges and fine detail edges. However, the boundary-curves of such
edges are vulnerable to fragmenting (which make them difficult to
detect). This phenomenon occurs for instance underneath the tumor (low
contrast) and to some degree inside the tumor (fine details). The
fragmenting - in particular for low-contrast edges - appears to
increase due to speckle-noise. One can reduce this problem by
connecting fragmented boundary-curves (e.g. by using coarse scale
information as suggested in \cite{jL92}). Preliminary studies have
shown that using this as a complimentary tool improves edge-detection
of low-contrast and fine detail edges. We use no synthesizing of the
boundary-curves in this study.\\ \indent Fig.\ref{fig:result}(d,e,f)
shows the result obtained by using the Canny edge-detector (with
hysteria). The scales is chosen equal the finest scale used by our
algorithm. One can observe that the edge-detector scheme studied in
Sect.\ref{SS:detector2} in general yields better representations of
the significant edges in the images, and includes less noise.\\
\indent \textit{Threshold:} The edge-detector assigns to each
boundary-curve $c_j$ a value $S(c_j)$ representing both its length and
value of wavelet transform in both time- and scale-space. Values of
$S$ corresponding to noise will therefore be concentrated at low
values, while those corresponding to edges will be distributed
discretely at high values. The result is that there is often a
separation between values of $S$ corresponding to noise and edges. A
similar phenomenon was observed in \cite{jL92}. In addition, since
values of $S$ corresponding to significant edges are often distributed
discretely it is possible to find representations which are
perceptually optimal.\\ \indent \textit{Computational complexity:} The
computational complexity of the edge-filtering procedure in
Sect.\ref{SS:detector2} is negligible. In what follows we focus on the
computational complexity of computing the wavelet transform and the
space-scale filtering procedure from Sect.\ref{SS:P2}.\\ \indent To
reduce the computational complexity of the space-scale filtering
procedure we use the tricks discussed in Sect.\ref{SS:P2} and
Sect.\ref{SS:detector2}. In other words we find the maxima-lines of
only a few randomly selected mod-max along each boundary-curve, and
use a sliding window (of size $4\cdot\mathrm{scale}\times 4\cdot
\mathrm{scale})$ to decide candidate connections at the next finer
scale. We use the fast Fourier-transform to compute the wavelet
transform.\\ \indent The space-scale filtering procedure has been
implemented in MATLAB version 7.9.0 (R2009b) on a laptop with a
Pentium 3 (2 GHz) processor. Tested on a $362\times 512$ image the
total processing time is $4.6$ seconds if we find the maxima-lines of
$50\%$ of the mod-max.  If we find only $10\%$ randomly selected
maxima-lines the total processing time is $1.9$ sec.\\ \indent
Computationally both computing the wavelet transform at different
scales and connecting mod-max across scales is well-suited for
parallel-computing. It is expected that this should allow us to
further reduce the total processing-time of the scale-space filtering
procedure.

\section{Conclusion}
We suggest a fast multi-scale edge-detection algorithm for medical
ultrasound signals. The edge-detector is based on the continuous
wavelet transform, and takes advantage of the behavior of maxima-lines
in the space-scale plane. Our algorithm shows promising results for
detecting the significant edges in medical ultrasound signals.\\
\indent The key-point of the edge-detection scheme is a fast
space-scale filtering procedure for medical ultrasound signals. The
procedure is designed to accommodate a set of patterns modeling
intensity-changes which are typical in such signals. This allow us to
manage and reduce the ambiguity introduced when using the multi-scale
wavelet transform to detect edges in a medical ultrasound signal. The
space-scale filtering procedure connects mod-max which belong to the
same maxima-line by computing the wavelet transform at only a sparse
set of scales. For our medical ultrasound signals we typically need to
compute the wavelet transform at the four scales $s =
\{4,8,16,32\}$. In the popular Edge-focusing algorithm \cite{fB87} one
computes the wavelet transform at a linear set of scales ($\Delta s =
1/2$) to trace mod-max in the space-scale plane. Our proposed
space-scale filtering procedure will therefore considerably reduce the
computational efforts needed to decide which mod-max belong to the
same maxima-line.\\ \indent Based on the space-scale filtering
procedure we study a novel edge-detection algorithm for medical
ultrasound signals. The edge-detector assigns to each candidate edge a
value representing the behavior of its corresponding maxima-line in
the space-scale plane. Experimental results indicate that the
edge-detector finds both high- and low-contrast edges in medical
ultrasound images and yet disregards noise. Compared to the Canny
edge-detector we observe that our algorithm finds the significant
edges, and includes less redundant information such as noise. Our
study has showed that one can construct multi-scale edge-detection
algorithm for medical ultrasound signals which meets the demand of
speed, but not on the expense of reliability.\\ \indent We also expect
that using multi-scale information opens new possibilities for further
improvements of edge-detection in medical ultrasound images. One
possibility is to dynamically choose the finest scale based on local
automatic scale selection as described in e.g. \cite{tL98}.

\section*{Acknowledgment}
The author would first of all like to thank his supervisor Prof. Yurii
Lyubarskii for the support and guidance during the work.\\ \indent The
author would also like to thank Dr. Jianhua Yao (National Institue of
Health) for useful discussions.\\ \indent The author would like to
thank SINTEF Medical Technology for providing medical ultrasound
images obtained through clinical studies at St. Olavs Hospital,
Trondheim, Norway.


\bibliographystyle{elsarticle-num}
\bibliography{bibwavelet,bibultrasound,bibimgprocess,bibshearlet}
\end{document}